\documentclass[10pt,twocolumn,letterpaper]{article}

\usepackage{cvpr}      %
\usepackage{tabularx}
\usepackage{setspace}

\usepackage[dvipsnames]{xcolor}

\usepackage{array}

\definecolor{cvprblue}{rgb}{0.21,0.49,0.74}
\usepackage[pagebackref,breaklinks,colorlinks,citecolor=cvprblue]{hyperref}

\def\paperID{296} %
\def\confName{3DV\xspace}
\def\confYear{2025\xspace}

\title{Gen3DSR: Generalizable 3D Scene Reconstruction via Divide and Conquer \\ from a Single View}

\author{Andreea Ardelean \quad \quad \quad
Mert Özer \quad \quad \quad Bernhard Egger\\
Friedrich-Alexander-Universität Erlangen-Nürnberg \\
{\tt\small \{andreea.dogaru, mert.oezer, bernhard.egger\}@fau.de}
}

\newlength{\wid}
\newlength{\mrg}
\newlength{\mrgv}

\usepackage{arydshln}
\usepackage{amssymb}%
\usepackage{pifont}%
\newcommand{\cmark}{\ding{51}}%
\newcommand{\xmark}{\ding{55}}%

\newcommand{\vcentered}[1]{\begin{tabular}{@{}c@{}}#1\end{tabular}}
\usepackage{multirow}
\usepackage{microtype}

\begin{document}

\twocolumn[{%
\renewcommand\twocolumn[1][]{#1}%
\maketitle
\begin{center}  
    \vspace{-0.2cm}
    \setlength{\wid}{0.23\textwidth}
    \setlength{\mrg}{-0.1cm}
    \setlength{\mrgv}{-0.25cm}
    \begin{tabular}{cccc}
        \includegraphics[width=\wid]{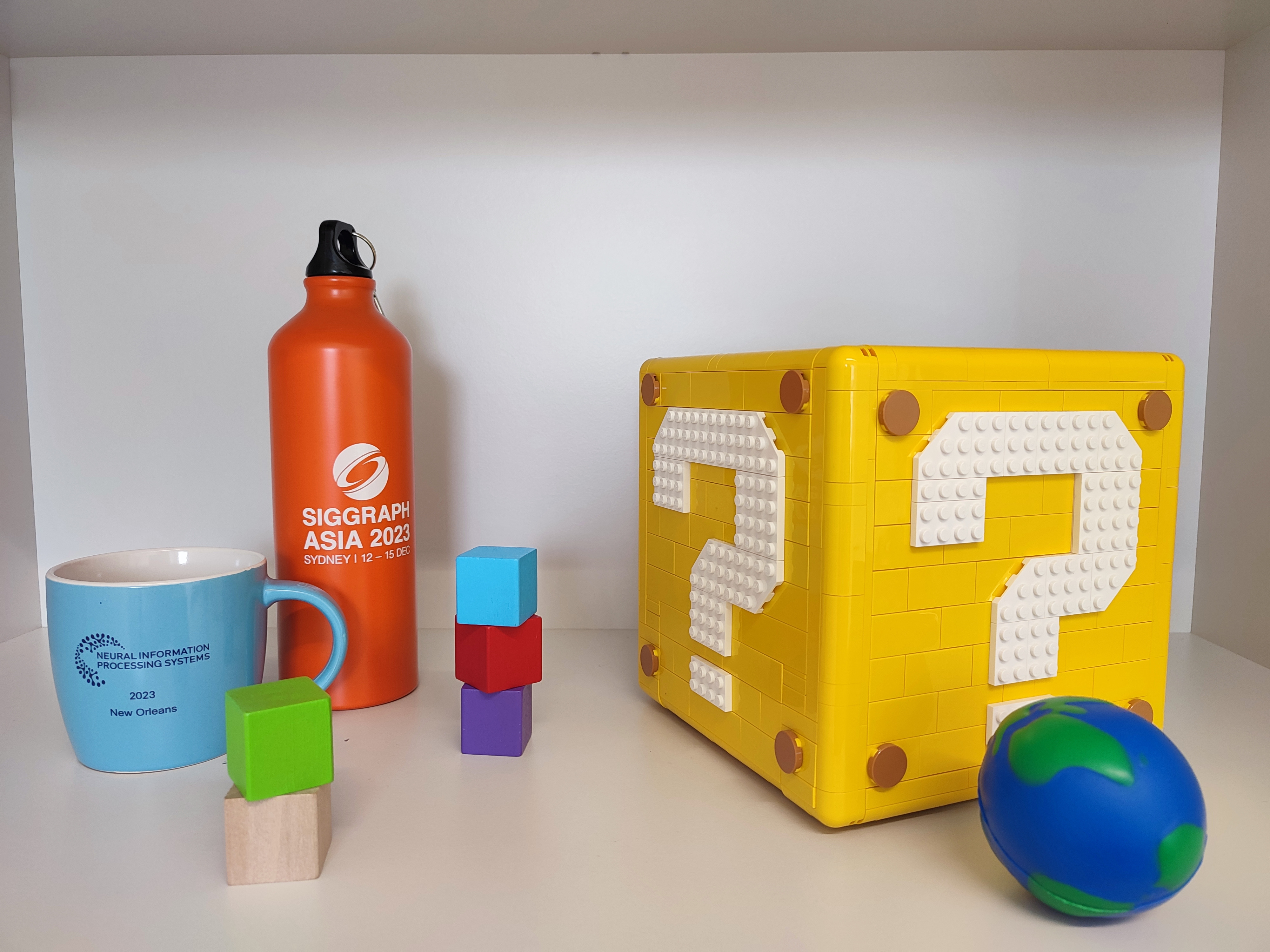} &
        \hspace{\mrg}
        \includegraphics[width=\wid]{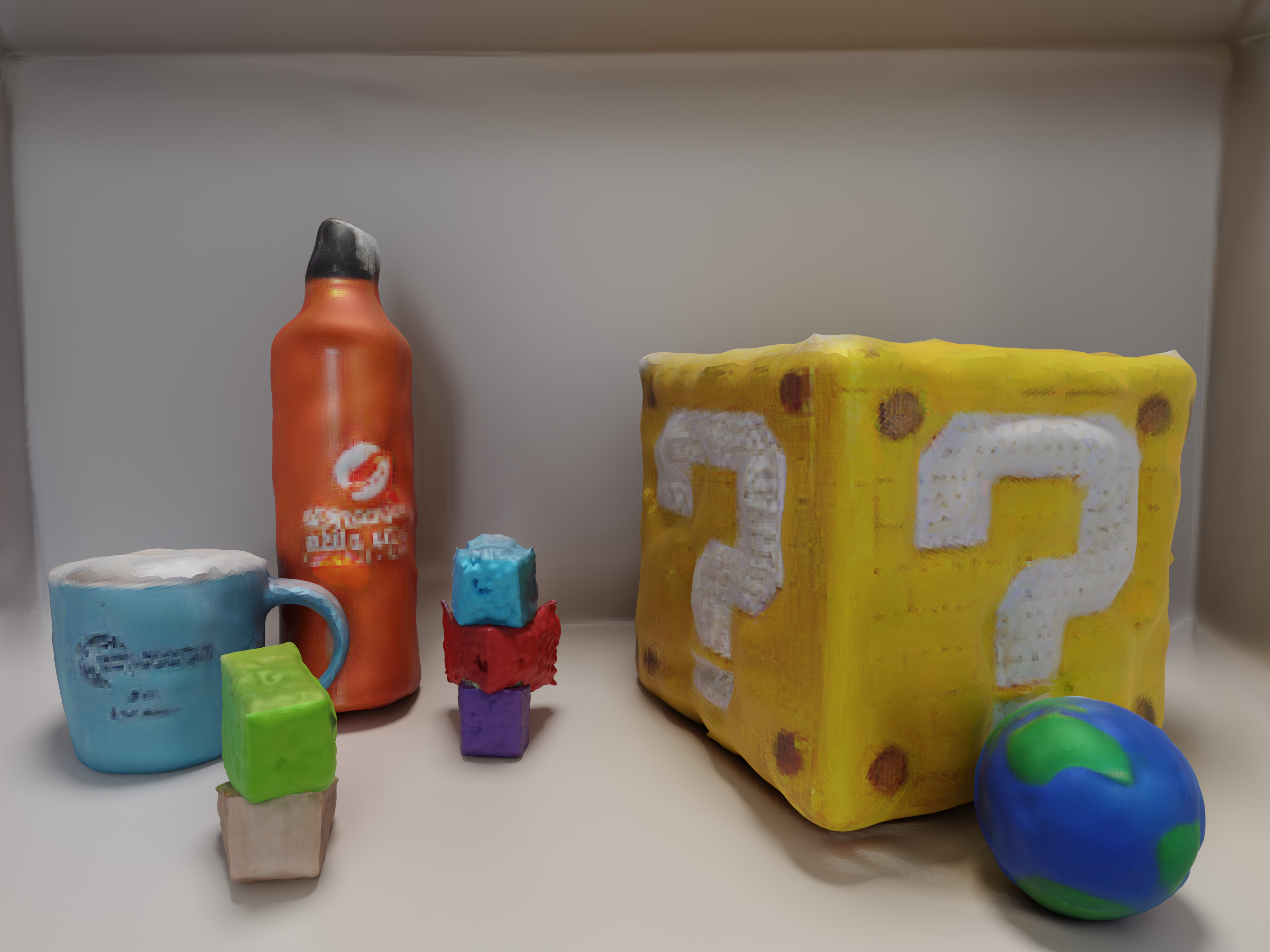} &
        \hspace{\mrg}
        \includegraphics[width=\wid]{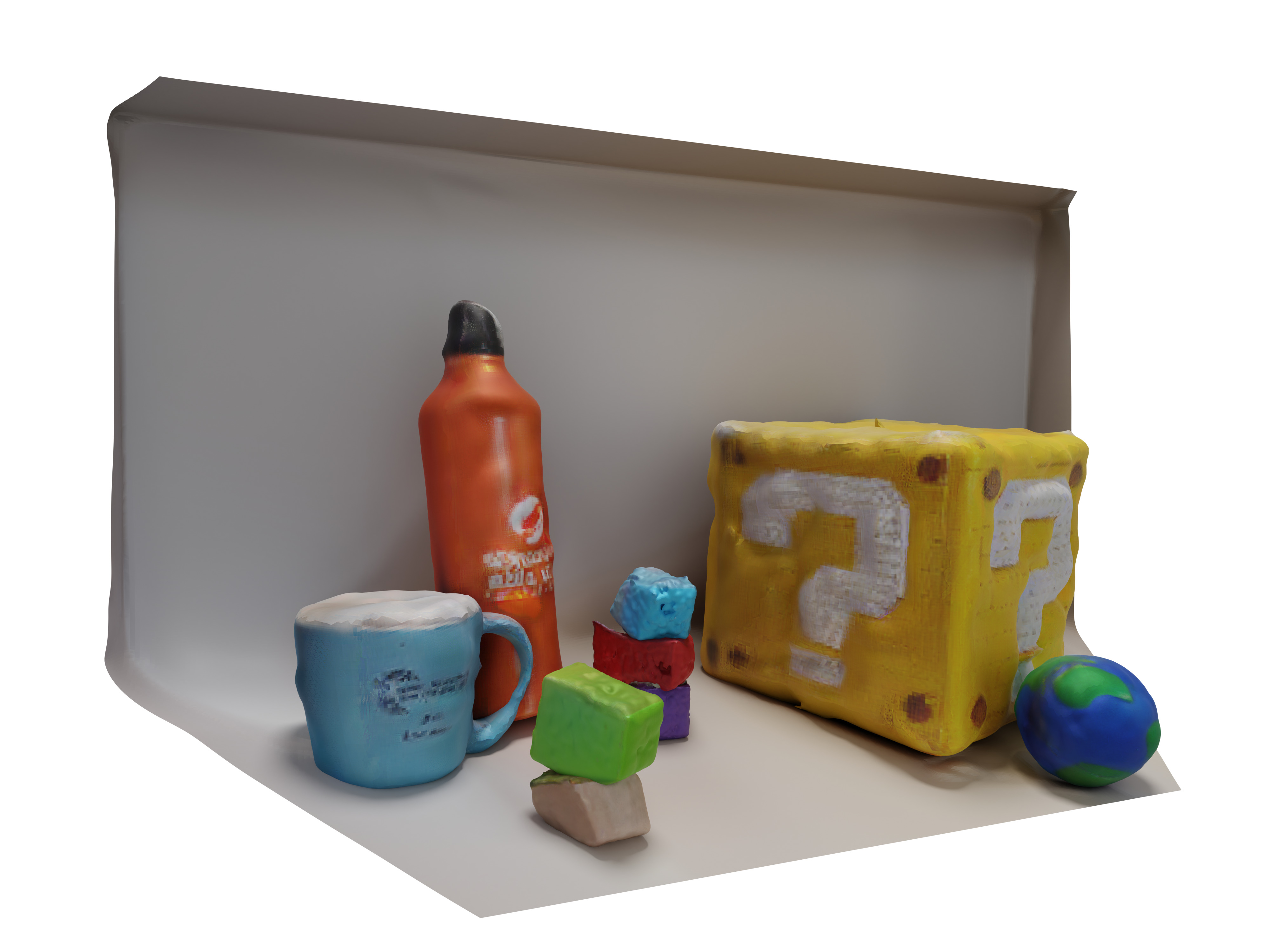} & 
        \hspace{\mrg}
        \includegraphics[width=\wid]{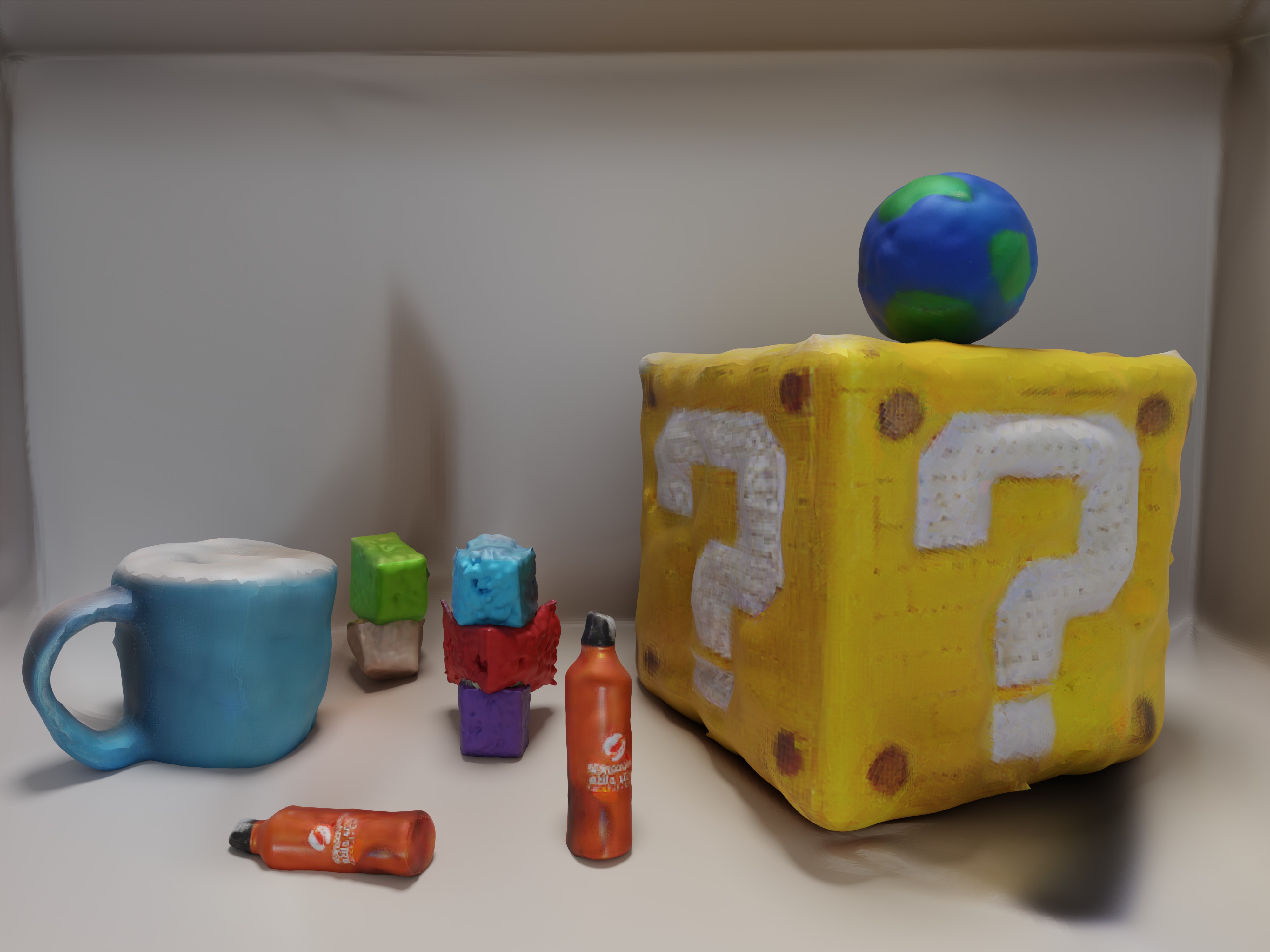} 
        \\ 
       Input image & 
        \multicolumn{2}{c}{3D Scene reconstruction}
         & \hspace{\mrg}
        3D Scene editing
        \\ 
        \multicolumn{4}{c}{\includegraphics[width=\textwidth]{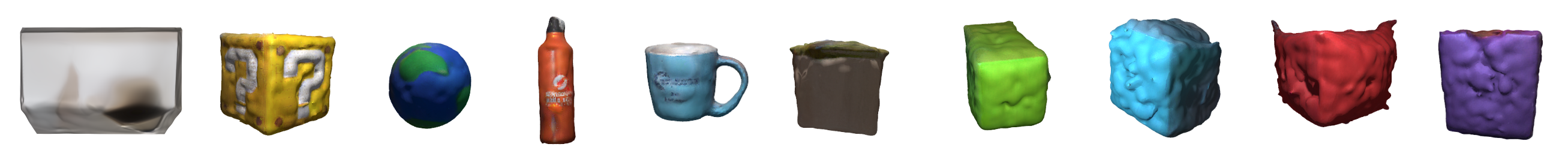} 
        }
        \\ 
        \multicolumn{4}{c}{
        Scene components
        }
    \end{tabular}
    \vspace{-0.3cm}
    \captionof{figure}{Our method can reconstruct a 3D scene from a single view. We 
    identify distinct objects, 
    address their
    occlusions through amodal completion, and reconstruct them individually. The resulting 3D objects
    are composed into the scene using monocular depth guides. Each component is reconstructed as a triangle mesh, enabling downstream applications such as scene manipulation and editing.} 
    \label{fig:teaser}
\end{center}
}]

\begin{abstract}
\vspace{-0.3cm}
Single-view 3D reconstruction is currently approached from two dominant perspectives: reconstruction of scenes with limited diversity using 3D data supervision or reconstruction of diverse singular objects using large image priors. However, real-world scenarios are far more complex and exceed the capabilities of these methods. We therefore propose a hybrid method following a divide-and-conquer strategy. We first process the scene holistically, extracting depth and semantic information, and then leverage an object-level method for the detailed reconstruction of individual components. By splitting the problem into simpler tasks, our system is able to generalize to various types of scenes without retraining or fine-tuning. We purposely design our pipeline to be highly modular with independent, self-contained modules, to avoid the need for end-to-end training of the whole system. This enables the pipeline to naturally improve as future methods can replace the individual modules. We demonstrate the reconstruction performance of our approach on both synthetic and real-world scenes, comparing favorable against prior works. Project page: \small{\url{https://andreeadogaru.github.io/Gen3DSR}
}
\end{abstract}
    
\vspace{-0.9cm}
\section{Introduction}
\label{sec:intro}
\vspace{0.04cm}

Single-view 3D scene reconstruction refers to the problem of understanding and explaining all the visible components that assembled together create a 3D scene which closely reproduces the original 2D observation. 
The computer vision and graphics communities have long been interested in automating this task, yet its complexity still leaves room for many improvements \cite{tretschk2023state, khan2022three}. 
Successful single-view applications have been developed for specific purposes such as face reconstruction \cite{egger20203d, Khakhulin2022ROME} and hair modeling \cite{wu2022neuralhdhair}. However, 3D understanding from a single image is far from solved in the case of larger scale problems such as indoor/outdoor scene reconstruction with multiple objects \cite{samavati2023deep}.

In general, even reconstructing one 3D object from a single image is a severely ill-posed problem, \eg it is impossible to tell precisely how the back side of an object looks like if the input image only observes the front. Nonetheless, if the distribution of objects that are naturally present in our day-to-day lives is known, one can plausibly predict the shape and appearance of a 3D object from very limited information. Accordingly, various priors have been used in the context of particular object classes (such as simple shapes \cite{alwala2022pre}, or human faces \cite{egger20203d}). However, modeling entire scenes is a significantly more challenging problem.

Given the complexity of real-world scenes, reversing the process of image capturing
in an end-to-end fashion would require a huge amount of data covering the variability of realistic environments. Therefore, many works solve a simplified version of the task by focusing on single objects or indoor rooms with a limited number of object classes.
Under these assumptions, most of the existent solutions rely on 3D scene geometry supervision from synthetic datasets. This class of methods usually struggles when applied to real-world images due to the domain gap and limited diversity in existing datasets.
In contrast, we propose to tackle the single-view 3D scene reconstruction problem in a divide-and-conquer approach while building on the advances in related, simpler tasks.
In Figure \ref{fig:teaser} we show that, following this approach, our pipeline is able to reconstruct multi-object scenes with unprecedented quality.

In the past few years, the field of computer vision has seen tremendous progress in solving particular tasks such as depth estimation and single-image 3D object reconstruction. It is the right time for these components to be assembled to solve the challenging task of full 3D scene reconstruction. 
We have identified the following sub-problems that together comprehensively explain a 3D scene and enable its reconstruction from a single input image: estimating the camera calibration, predicting the (metric) depth map, segmenting entities, detecting foreground instances, reconstructing the background, recovering the occluded parts of the individual objects (amodal completion), and reconstructing them. 
Our disentangled framework
is open for incremental improvements and future enhanced modules can be easily plugged in to boost the reconstruction performance of the entire system.
The main contributions of this paper are as follows:

\begin{itemize}
    \item We design a compositional framework and the corresponding abstractions, enabling scene-level 3D reconstruction without
    end-to-end training.
    \item We build a model for amodal completion and show how it can be used towards achieving full scene reconstruction. 
    \item We develop the connecting links for integrating individually reconstructed 3D objects into the scene layout by exploiting single-view depth estimation. 
    \item We achieve an unmatched level of generalizability for real-world single-view 3D scene reconstruction, which we demonstrate through extensive  evaluations.
\end{itemize}

\section{Related Work}
Starting with the pioneering work of Lawrence Roberts \cite{roberts1963machine}, the task of recovering 3D scene properties such as geometry, texture, and layout from a single 2D image has been extensively studied. 
Learning-based advancements led to substantial progress in this challenging task.
However, existing methods still have limited understanding and struggle to faithfully reconstruct complex realistic scenes. In this section, we review the major lines of work and highlight the advantages of our approach that enable state-of-the-art results for 3D scene reconstruction from a single view. 

\textbf{Feed-forward scene reconstruction.} The direct regression conditioned on the input image used in many computer vision tasks can also be employed for 3D scene reconstruction. Provided the availability of large scene collections, such systems are trained end-to-end using 3D supervision. Most such works \cite{dahnert2021panoptic, zhang2023uni, chu2023buol} rely on an encoder-decoder architecture that takes as input the image and predicts voxel grids containing scene properties such as geometry, semantics and instance labels. These methods have the advantage that by predicting the scene layout jointly with the containing objects, the object poses are intrinsically correct. 
However, these solutions suffer from resolution limitations, require large 3D data collections for training \cite{front20213d, future20213d} and usually do not generalize well to real-world scenes. In contrast, we use a modular framework that does not require end-to-end training or 3D supervision. 

\textbf{Factorized scene reconstruction.} Following the formulation of Tulsiani \etal \cite{tulsiani2018factoring}, the scene can be considered as composed of different factors, including layout, object shape and pose, that together make up a 3D scene representation. As this approach reconstructs the scene components individually, it is beneficial for downstream applications and the reconstruction system can be designed with specialized components for each type of factor. Methods following this paradigm initially reconstructed the objects in the scene as bounding boxes \cite{huang2018cooperative, schwing2013box, du2018learning} or voxels \cite{tulsiani2018factoring, li2020geometry, kulkarni20193d}, while later works use meshes \cite{gkioxari2022learning, nie2020total3dunderstanding, huang2018holistic, kuo2020mask2cad} or neural fields \cite{ liu2022towards, Zhang_2021_Holistic, zakharov2021single, stelzner2021decomposing}. Apart from the elementary factors of \cite{tulsiani2018factoring}, more recent methods also include camera attributes \cite{nie2020total3dunderstanding, huang2018holistic, liu2022towards, huang2018cooperative}, textures \cite{yan2023psdr, chen2023ssr, Yeh_2022_CVPR}, lightning \cite{yan2023psdr, Yeh_2022_CVPR}, or even material properties \cite{yan2023psdr, Yeh_2022_CVPR}. Depending on the focus of the method, some works rely on a retrieval algorithm for identifying the object candidates \cite{yan2023psdr, huang2018holistic, izadinia2017im2cad, bansal2016marr, kuo2020mask2cad} or regress their geometry \cite{gkioxari2022learning, liu2022towards, Zhang_2021_Holistic, stelzner2021decomposing}. The initially predicted scene components are sometimes further fine-tuned using derivative-free optimization \cite{izadinia2017im2cad, huang2018holistic, chen2019holistic} or differentiable rendering \cite{gkioxari2022learning, Yeh_2022_CVPR, yan2023psdr, zakharov2021single}. 
Our method also uses a compositional scene reconstruction framework; in our pipeline, each module is trained individually on specific datasets. This strategy allows our proposed solution to transcend predefined (limited) classes of objects and reconstruct diverse scenes effectively.

\begin{figure*}
    \centering
    \includegraphics[width=0.99\textwidth]{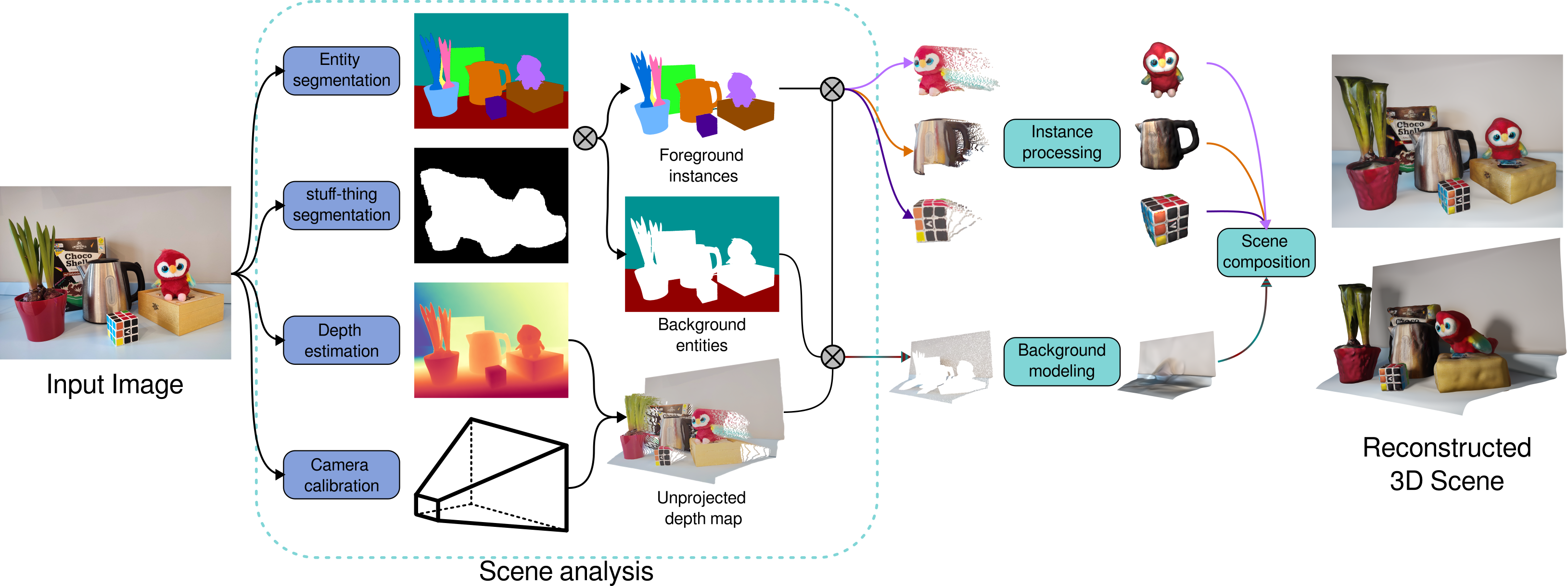}
     \vspace{-0.2cm}
    \caption{Method Overview: the input image is first analyzed collectively by an ensemble of state-of-the-art monocular models. Subsequently, the identified instances are individually processed, as elaborated in Figure~\ref{fig:obj_rec}. The reconstructed objects, along with the modeled background, are composed into the final scene, which can then be used in various applications. }
    \label{fig:pipeline}
    \vspace{-0.5cm}
\end{figure*}

\textbf{Single-view scene understanding.}
The reconstruction of simple scenes composed of a single isolated object has seen numerous approaches along the years \cite{Han2019ImageBased3O}. Nonetheless, enabled by the versatile diffusion-based models capable of generating realistic images \cite{Rombach2021HighResolutionIS, podell2023sdxl} and large collections of 3D objects \cite{deitke2023objaverse, deitke2023objaversexl}, there has recently been a surge of single-view 3D reconstruction methods \cite{liu2023one, liu2023one2345p, liu2023zero, tang2023dreamgaussian} capable of creating digital objects with unprecedented quality. 
Closely related to 3D scene reconstruction is monocular depth estimation \cite{mertan2022depthreview}, which predicts from a single image a 2.5D representation of the visible content. Recent methods focus on improving the generalizability of the estimator by following a training protocol employing large datasets \cite{bhat2023zoedepth, depthanything} or by fine-tuning a large image prior such as Stable Diffusion \cite{Rombach2021HighResolutionIS} for the depth estimation task \cite{ke2023repurposing}. 
A scene is also described by the camera used to capture the image \cite{jin2023perspective, Lee2021CTRL, HoldGeoffroy2017APM} and by the separation between different instances \cite{Kirillov_2023_segmenta, Qi_2023_ICCV, yuan2024ovsam}. Though the current state-of-the-art methods for each individual sub-task achieve robust results for real images, they only offer an incomplete explanation of the scene. Therefore, we compose them in an open framework capable of fully reconstructing complex real-world scenes from a single image.

\section{Method}
\label{sec:method}

\begin{figure*}
    \centering    \includegraphics[width=0.99\textwidth]{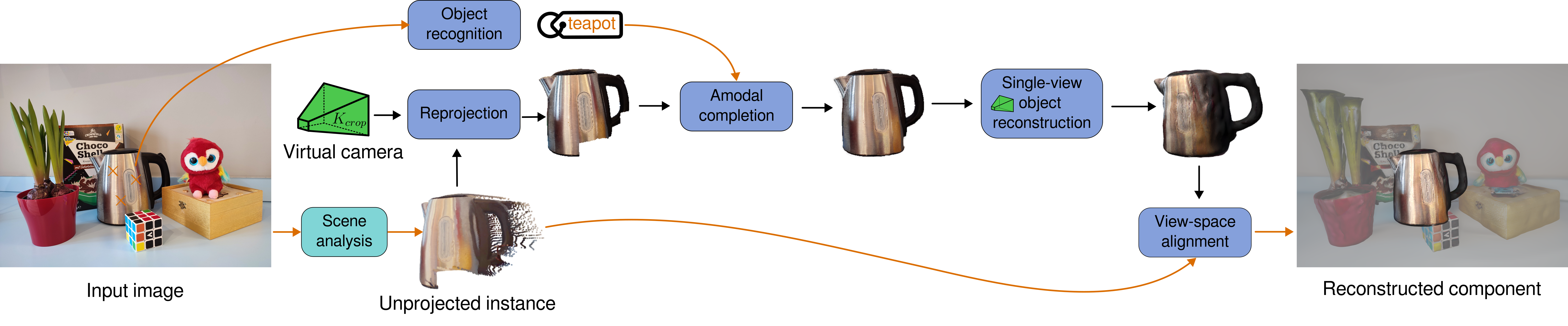}
    \vspace{-0.3cm}
    \caption{Detailed Overview over the processing steps of each instance. An image is processed through the scene analysis part of our framework as described in Figure~\ref{fig:pipeline}. Then, we add an object recognition information for diffusion guided completion to restore partially occluded objects. Lastly, we perform reconstruction and align the result back to the input view space.}
    \label{fig:obj_rec}
    \vspace{-0.5cm}
\end{figure*}

Our method, as illustrated in Figure~\ref{fig:pipeline}, takes as input a single RGB image $I$ and predicts the full 3D scene reconstruction $R(I)$ represented as a collection of triangle meshes. 
The proposed solution does not require end-to-end training and instead relies on off-the-shelf models carefully integrated into a seamless framework. 
First, we parse the image of the scene by finding the composing entities, and estimating the depth and camera parameters.
Then, we separate the identified entities in \textit{stuff} (amorphus shapes) and \textit{things} (characteristic shapes) \cite{caesar2018coco}. 
To recover the full view of each object, we perform amodal completion on the masked crops of the instances.
Each object is reconstructed individually in a normalized space and aligned to the view space using the scene layout guides from the depth map. Importantly, we address the differences in focal length, principal point, and camera-to-object distance between the two spaces through reprojection. Finally, we model the background as the surface that approximates the stuff entities collectively.
\subsection{Scene analysis}
Our framework decouples the object-agnostic from the object-specific processing, pursuing a balance between the representational power and the generalization capabilities of the integrated modules: 
That is, the perspective properties, semantic labels, and depth information are best retrieved by perceiving the scene as a whole

The geometry of a scene is characterized by its layout and the amodal shape of the contained objects. The layout of a scene refers to
the surfaces that enclose the space (\eg, walls) and the 3D locations of the objects.
To model the layout, we unproject a \textbf{monocular depth estimation} $D$, of the input image using predicted \textbf{camera calibration} parameters, $K_{img}$, as a point cloud in the 3D view space, $P^{view} \in \mathbb{R} ^ 3$, and adopt it as our guide for positioning the scene components. 
A 2.5D representation is not sufficient to fully describe the layout of a scene as it only provides information for the visible parts. 
Still, it can be used to integrate individually reconstructed 3D objects into the scene (Section~\ref{ss:object_rec}) and for background estimation (Section \ref{ss:bg_rec}).

To parse the image, we opt for an \textbf{entity segmentation} \cite{Qi_2023_ICCV, qi2022open} approach in contrast to conventional 2D object detection, which enables us to segment all semantically-meaningful entities without being constrained to a predefined set of classes. 
This step partitions the image $I$ into instances $\{M_i\}$ that can be individually reconstructed to compose the whole scene. Furthermore, we consider the natural separation of the instances in \textit{thing} and \textit{stuff} using a universal image segmentation model. This facilitates our method to tailor the reconstruction process for each group, leveraging their unique properties (objects vs background).
In this stage, we also label the identified entities, which can provide more context for instance processing.

\subsection{Instance processing}
\label{ss:object_rec}

Let an object $O_i$ be a well-defined shape categorised as \textit{thing},
identified by its entity mask $M_i$, with corresponding RGB-D region $I_i = I[M_i]$, $D_i = D[M_i]$, and a label $L_i$. The processing steps are illustrated in Figure \ref{fig:obj_rec}.
We reconstruct each instance individually to fully benefit from a view-conditioned 2D diffusion model $\mathcal{Z}$ \cite{liu2023zero, deitke2023objaversexl}, trained using multi-view images of mostly single objects from large scale collections \cite{deitke2023objaverse, deitke2023objaversexl}. As the images used to train these models are rendered with a fixed predefined camera configuration $K_{crop}$, they generalize poorly to in-the-wild object crops. Therefore, we propose to address the domain-shift via \textbf{reprojection} of the object-associated pixels. To this end, we identify 
the virtual camera that together with the desired intrinsics $K_{crop}$ closely matches the observed image. Then, we use the transformation to project the unprojected pixels $P_i^{view}$ to a crop $C_i$ that resembles the training domain of $\mathcal{Z}$. 

For simple scenes in which there is no occlusion between instances, the crop $C_i$ represents the full view of the object $O_i$. However, this is not the case for most real-world scenes, and directly feeding $C_i$ to the object reconstruction method $\mathcal{R}$ would result in an incomplete object.
Therefore, we propose to recover the missing parts of $C_i$ by leveraging the image prior embodied by pre-trained large-scale diffusion models like Stable Diffusion. We approach the task named \textbf{amodal completion} which deals with recovering the shape and appearance of partially visible instances as an image-to-image translation problem. Specifically, we train a model to predict the view $\hat{C}_i$ of the full object $O_i$ conditioned on the object parts depicted in $C_i$ and the label $L_i$. Given the difficulty of collecting well-segmented training images guaranteed to contain complete objects, we generate a synthetic dataset specifically for this task. We render synthetic objects from a large collection and then obtain the conditioning images by masking out parts of the object with a randomly overlaying silhouette of another object. 
For more details about the dataset generation and differences between amodal completion and inpainting please see the supplementary material. 
We use this dataset to fine-tune Stable Diffusion, following the methodology of InstructPix2Pix \cite{brooks2022instructpix2pix}, and concatenate the encoded conditioning image to the noisy latent. The weights added to the base network are initialized with zero, while the rest are taken from the pre-trained model to benefit from the learned image prior.

The complete crop $\hat{C}_i$ can now be used as input for the \textbf{single-image 3D object reconstruction} method $\mathcal{R}$. 
Using a view-conditioned diffusion prior enables the model to reconstruct a wide range of objects from a single view without the need for 3D training supervision. The object is reconstructed using a differentiable 3D representation (\eg, neural fields  \cite{mildenhall2020nerf, mueller2022instant} or 3D Gaussians \cite{kerbl3Dgaussians}) that is either directly fitted to multi-view images generated by the diffusion prior, or by optimizing a Score Distillation Sampling-based loss \cite{Poole2022DreamFusionTU} against the diffusion prior.
Then, a polygonal mesh $R_i^{obj}$ aligned with the input crop $\hat{C}_i$ is extracted from the 3D representation using Marching Cubes~\cite{lorensen1987mc, Lewiner_mc}. The obtained mesh can optionally be further fine-tuned to refine the texture of the reconstructed object.

We transform the reconstructed instance from the object space (DreamGaussian) $R_i^{obj}$ to the view space (our scene) with the inverse transformation determined by the virtual camera used for projection. 
The obtained mesh $R_i^{view}$ is aligned to the object points $P_i$ up to an unknown scale factor $s_i$; this is because $\mathcal{R}$ reconstructs objects at an arbitrary scale.
We estimate $s_i$ as the scale factor that minimizes the distance between the visible points in $R_i^{view}$ and $P_i^{view}$.

\subsection{Background modeling}
\label{ss:bg_rec}
Compared to the objects contained in a scene, the entities categorised as \textit{stuff}, \eg walls or ceiling,  usually have a simpler geometry, which can be partially approximated by the corresponding regions in the depth map. Nonetheless, large areas are occluded by the foreground objects.
We consider all the background instances as one described by their mask union $M_{bg}$ and the corresponding image $I_{bg}$ and depth data $D_{bg}$. Then, we fit one small Multi-layer Perceptron $f:\mathbb{R}^3 \to \mathbb{R}$ to represent the signed distance function (SDF) to the unprojected background points $P_{bg}^{view}$, and another one $c:\mathbb{R}^3 \to \mathbb{R}^3$ to model the color associated to them. 
Lastly, we sample a dense grid of points in the camera frustum and evaluate their SDF values. The background surface is defined as 
$\mathcal{S} = \{x \in \mathbb{R}^3 | f(x) = 0\}$
and can be extracted from the grid using marching cubes \cite{lorensen1987mc, Lewiner_mc}.

\section{Experiments}
\begin{figure*}
    \centering    
    \setlength{\wid}{0.195\textwidth}
    \setlength{\mrg}{-0.35cm}
    \setlength{\mrgv}{-0.1cm}
    \begin{tabular}{ccccc}
        \includegraphics[width=\wid]{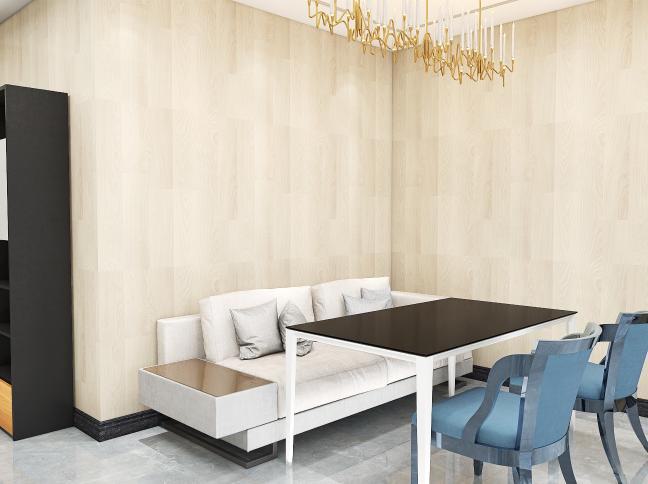} &
        \hspace{\mrg}
        \includegraphics[width=\wid]{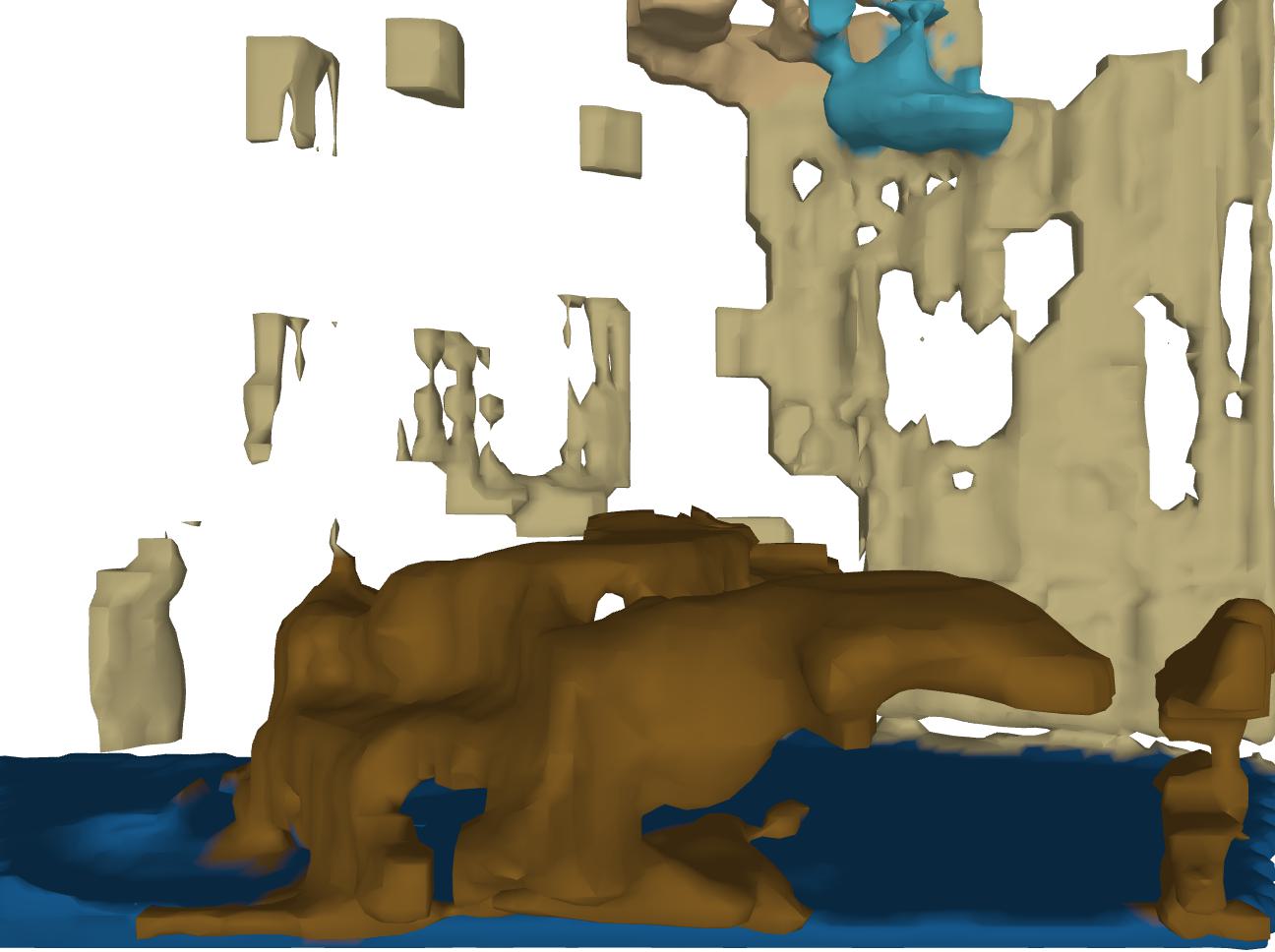} &
        \hspace{\mrg}
        \includegraphics[width=\wid]{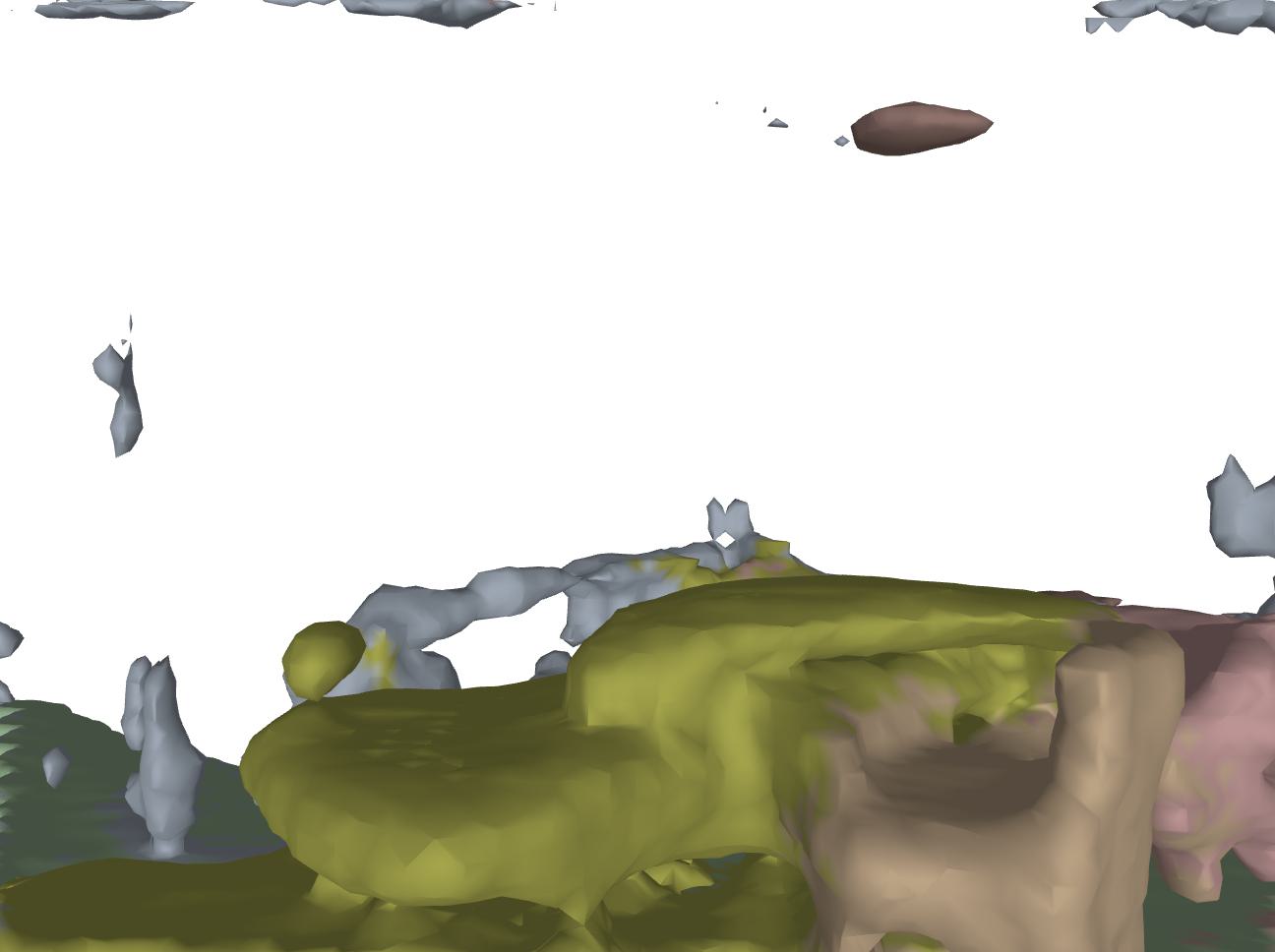} & 
        \hspace{\mrg}
        \includegraphics[width=\wid]{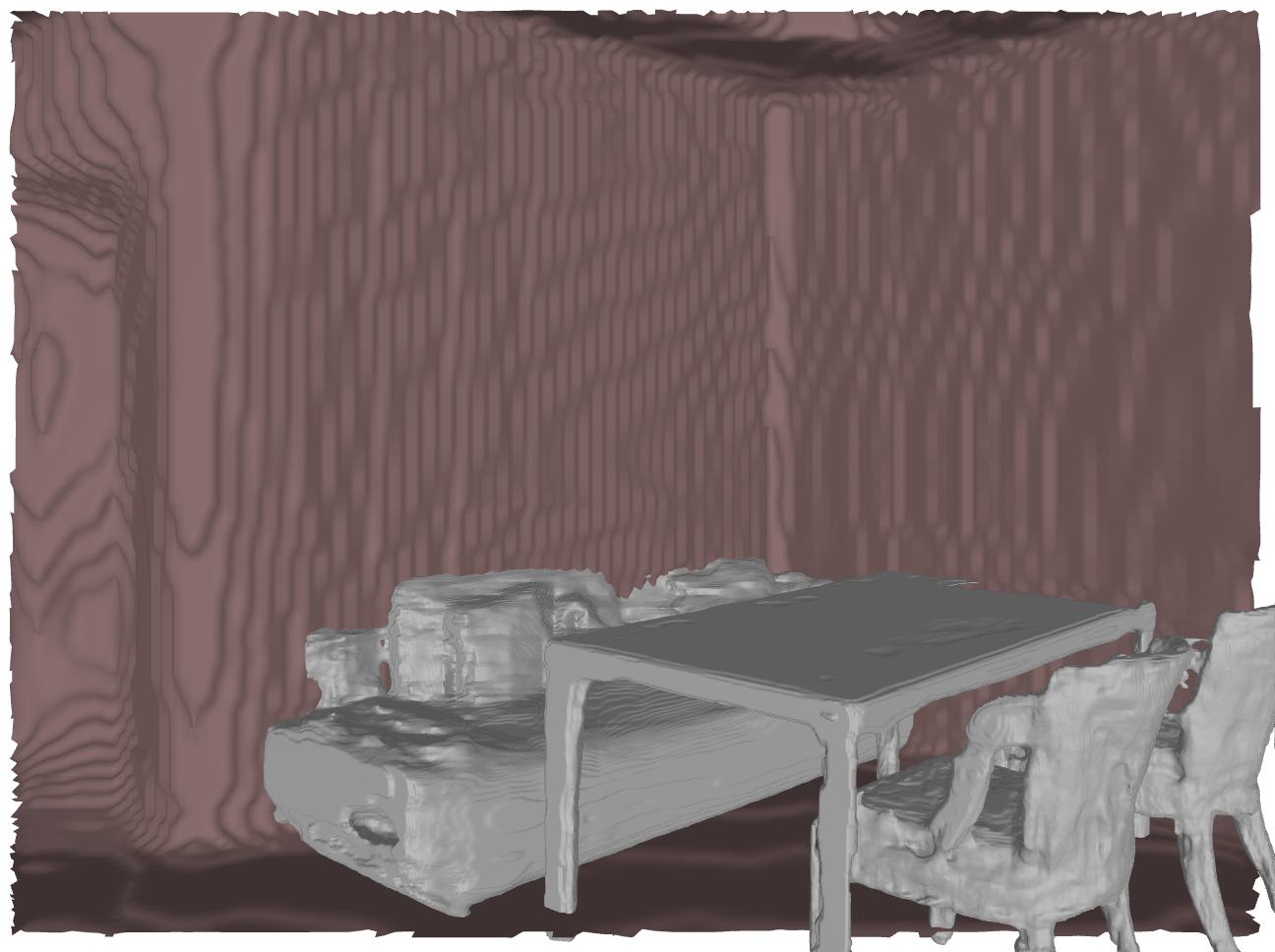} & 
        \hspace{\mrg}
        \includegraphics[width=\wid]{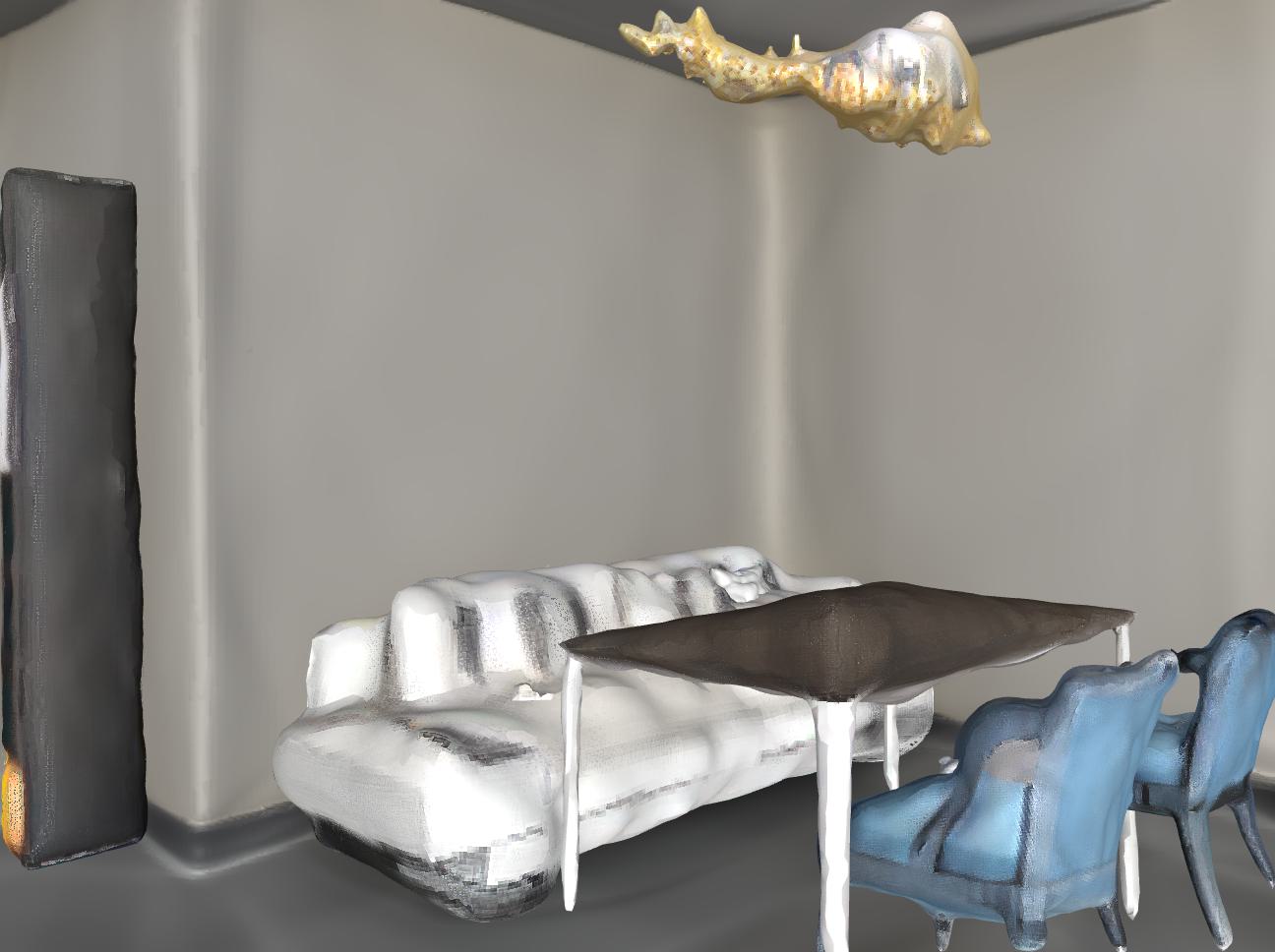} 
        \\ 
        \includegraphics[width=\wid]{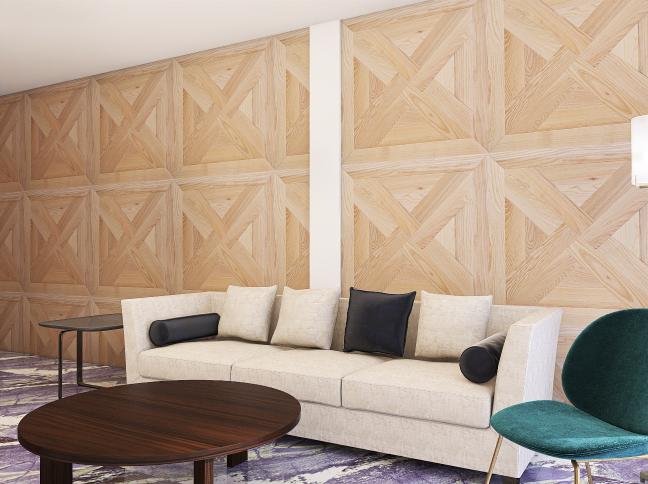} &
        \hspace{\mrg}
        \includegraphics[width=\wid]{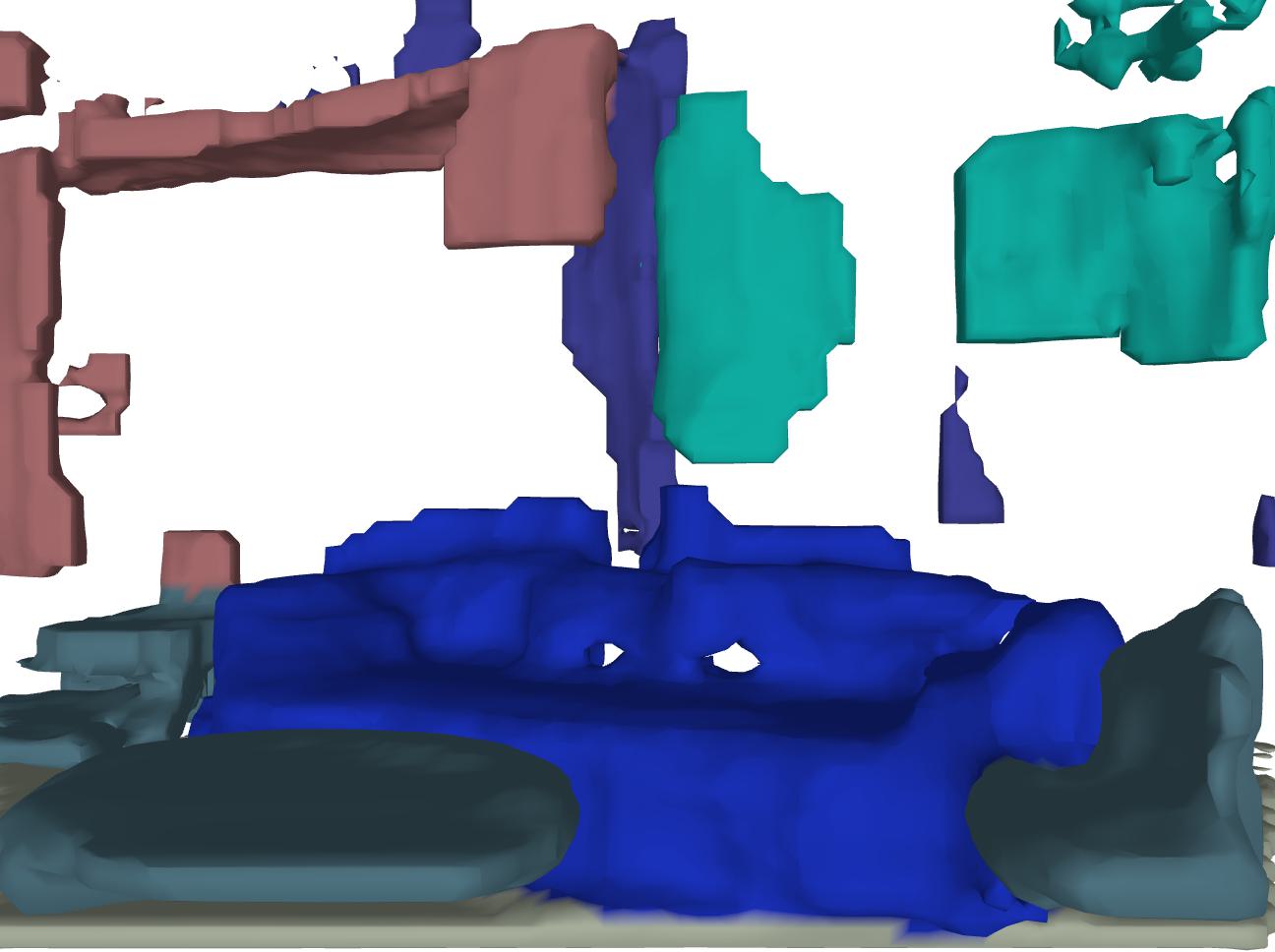} &
        \hspace{\mrg}
        \includegraphics[width=\wid]{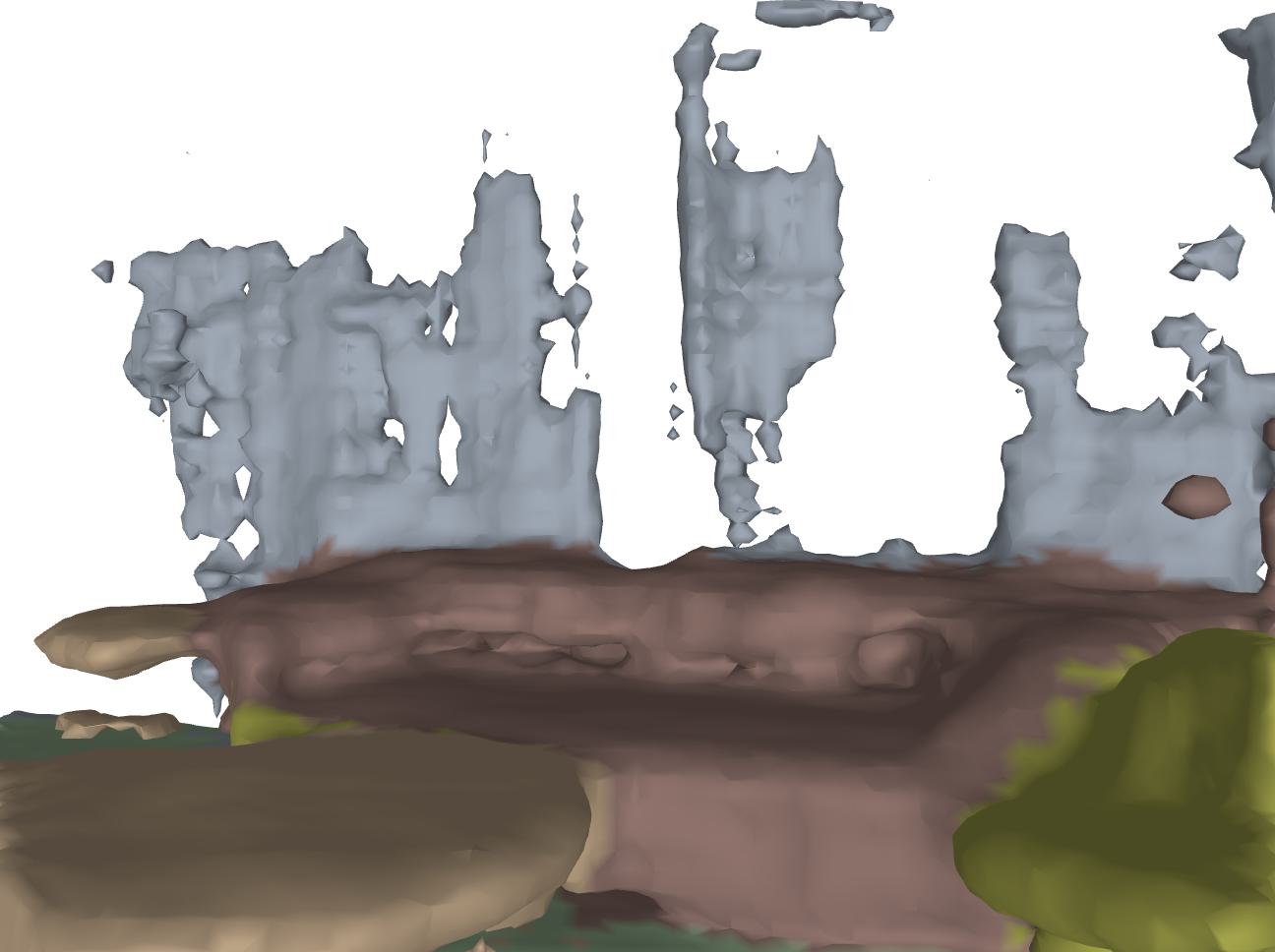} & 
        \hspace{\mrg}
        \includegraphics[width=\wid]{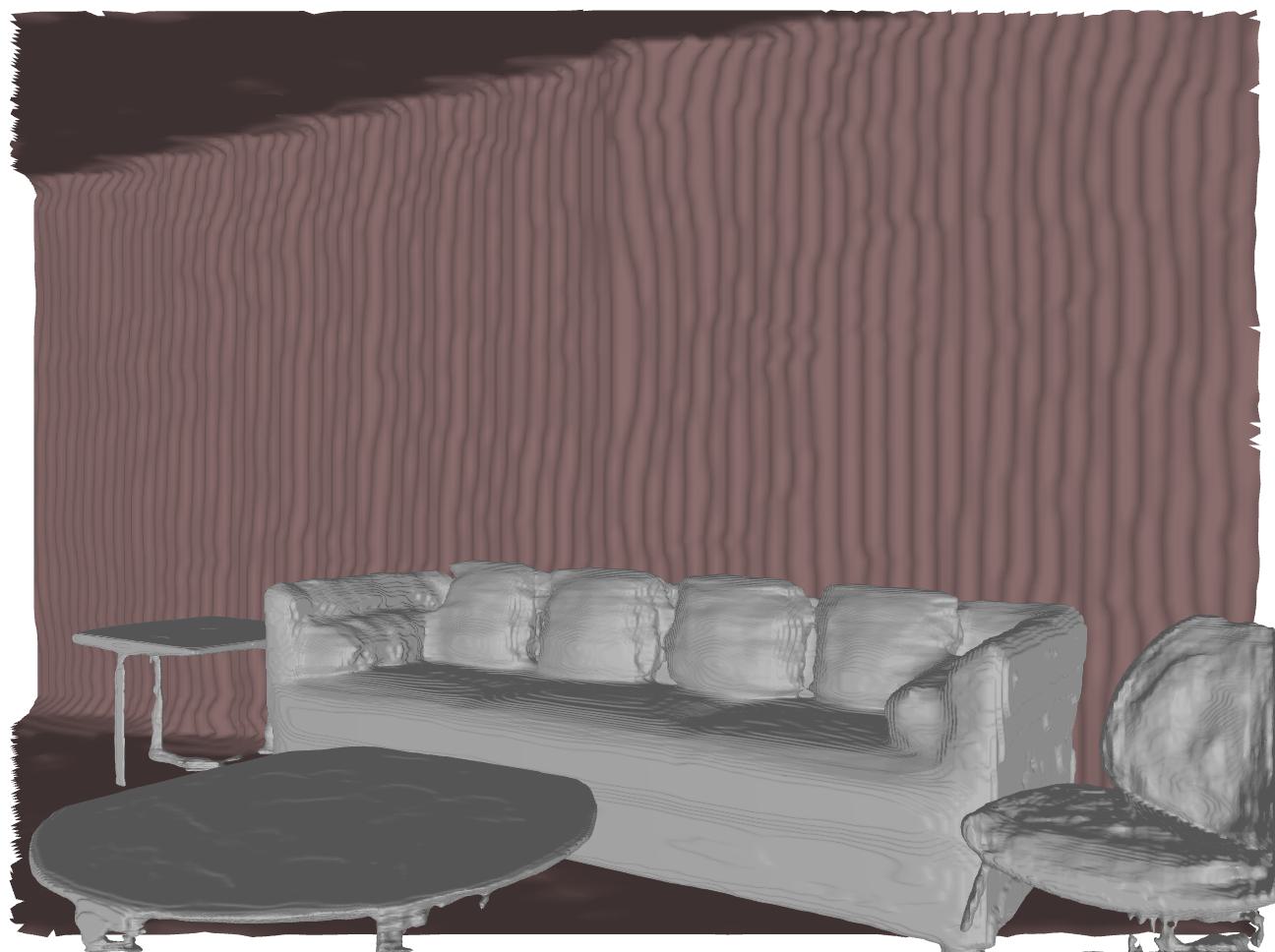} & 
        \hspace{\mrg}
        \includegraphics[width=\wid]{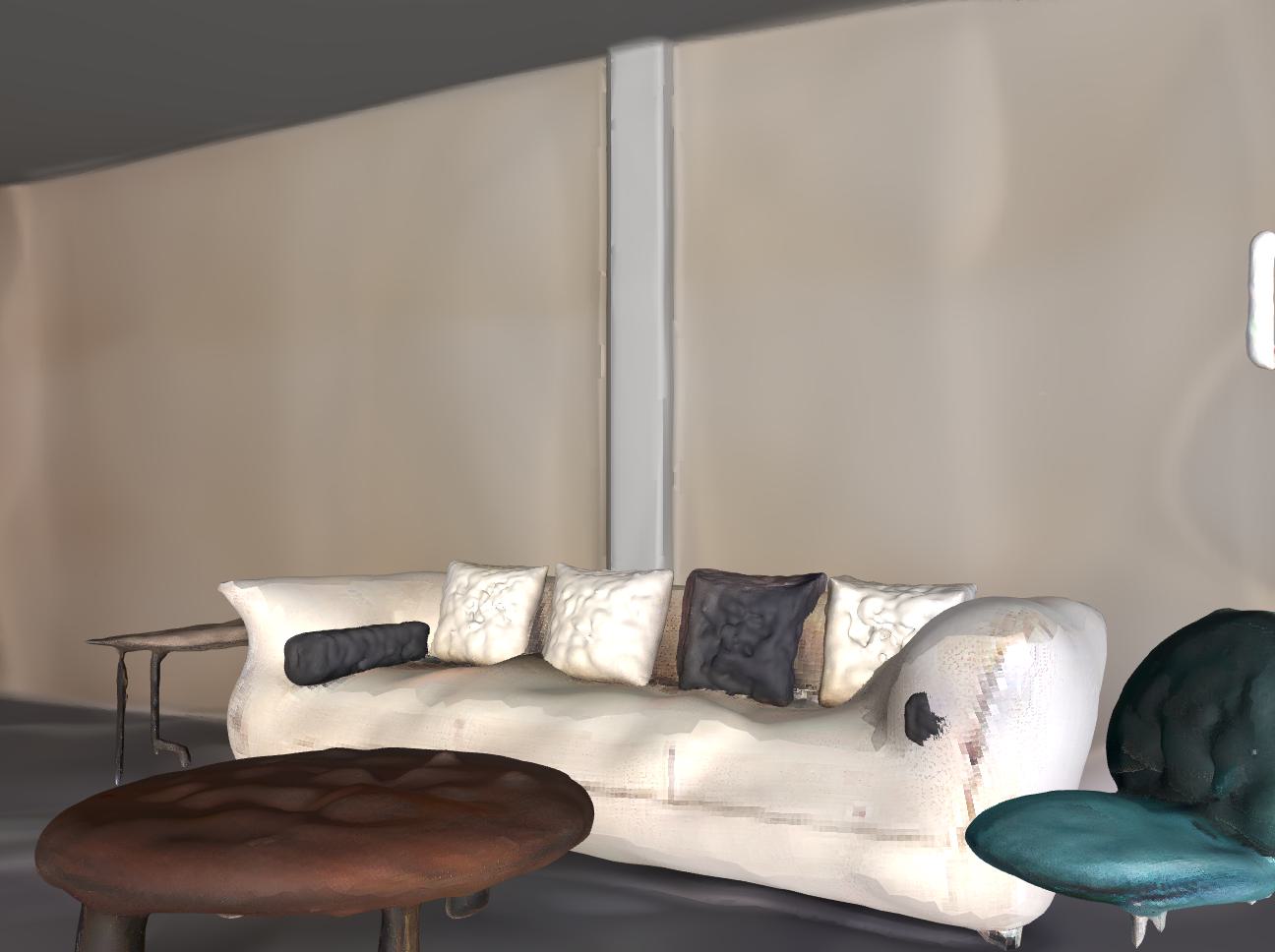} \\
        \includegraphics[width=\wid]{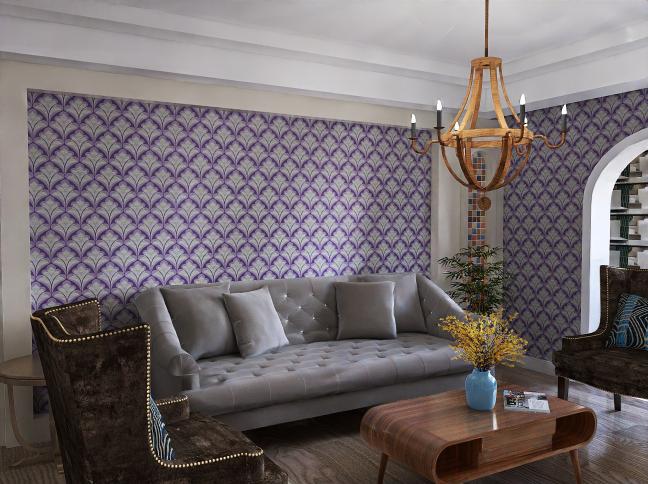} &
        \hspace{\mrg}
        \includegraphics[width=\wid]{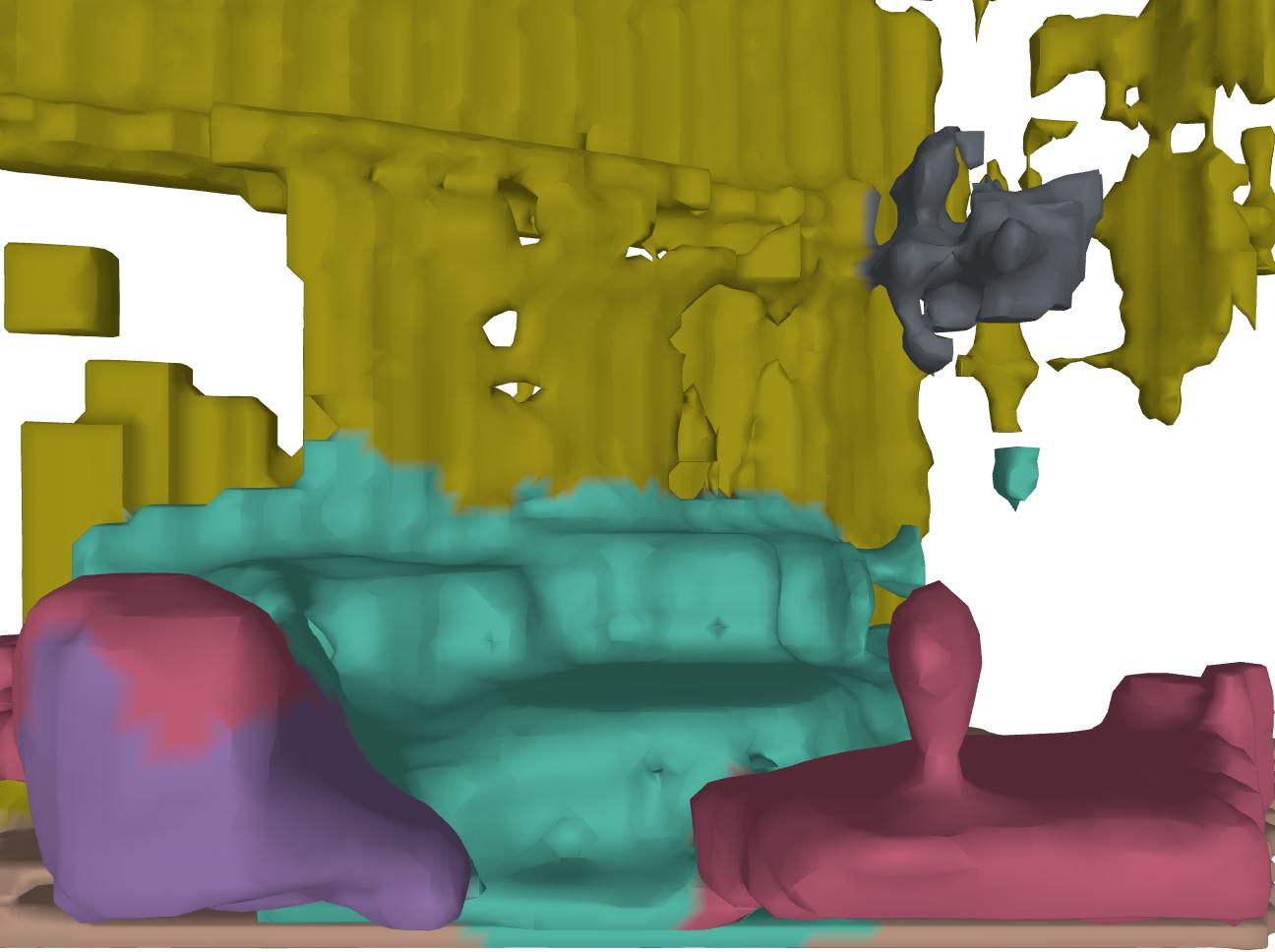} &
        \hspace{\mrg}
        \includegraphics[width=\wid]{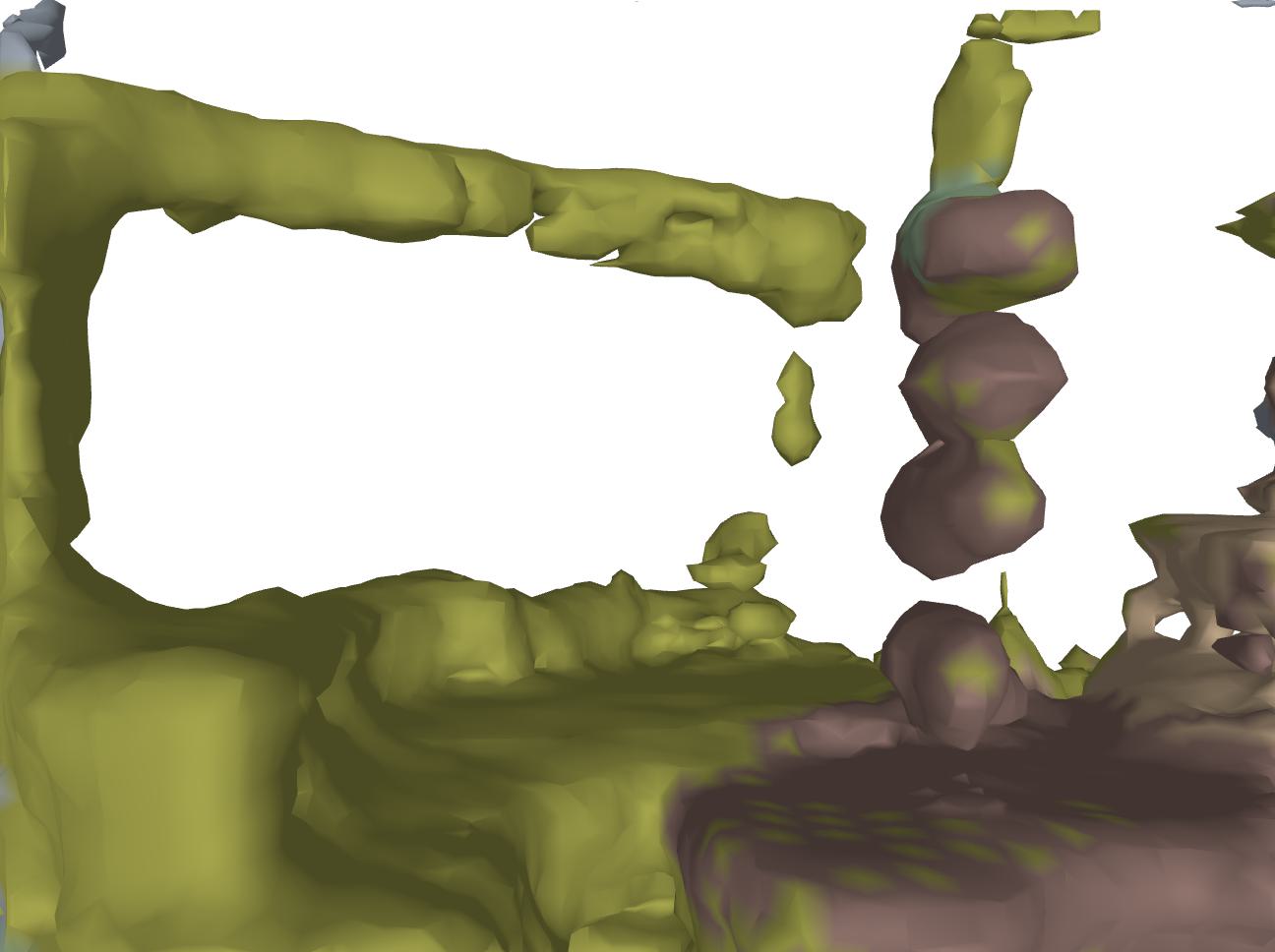} & 
        \hspace{\mrg}
        \includegraphics[width=\wid]{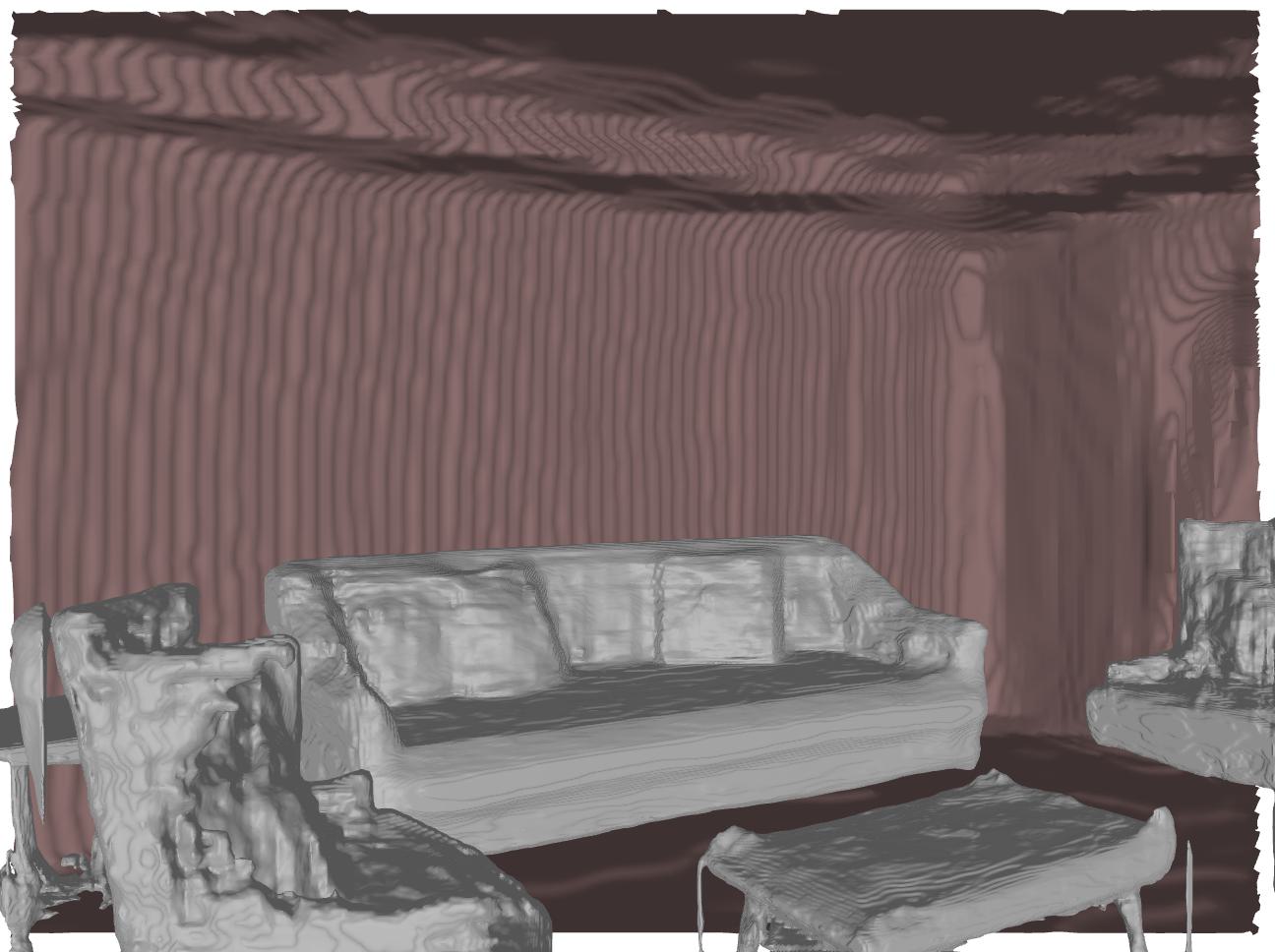} & 
        \hspace{\mrg}
        \includegraphics[width=\wid]{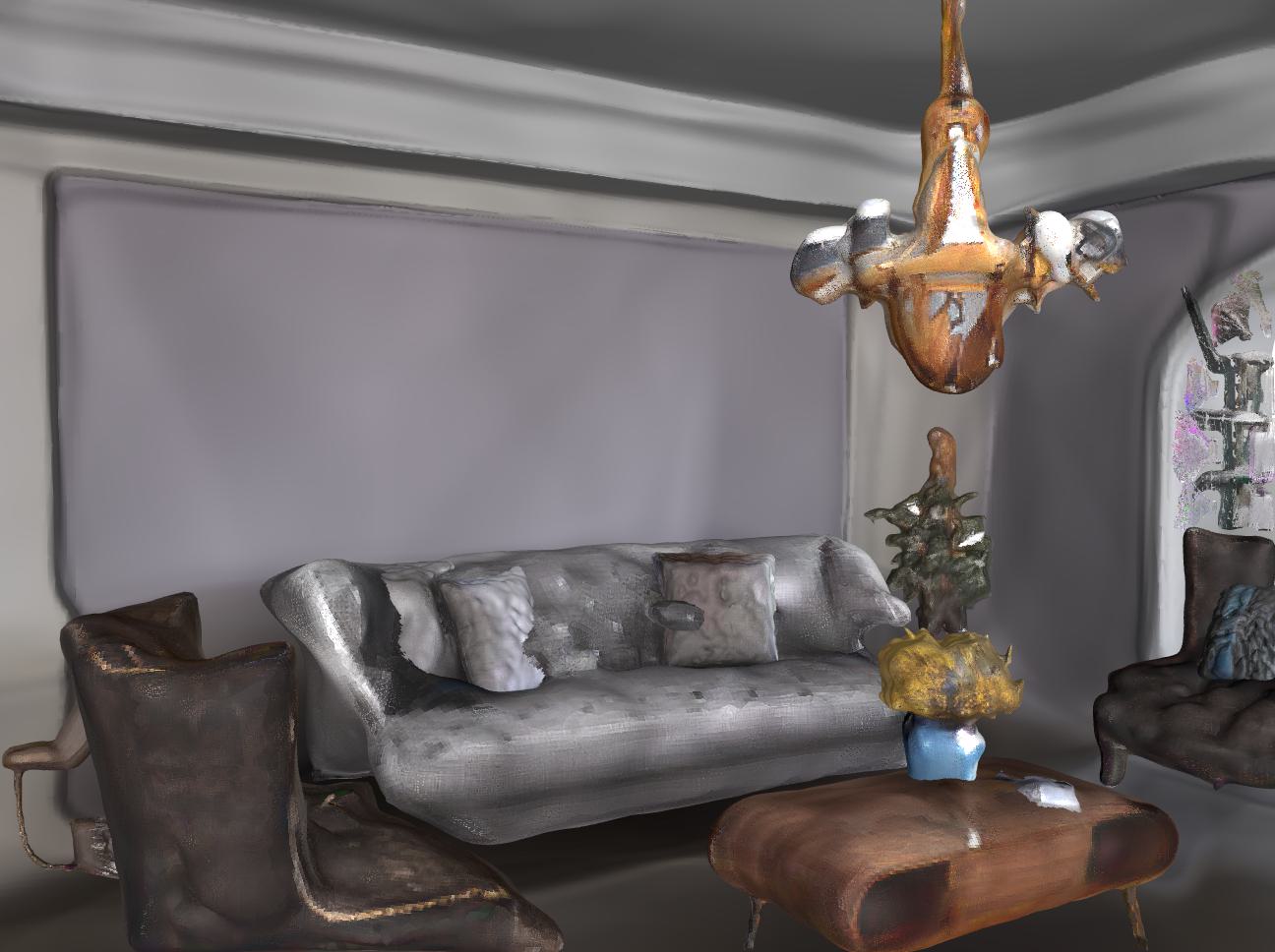} \\
        
        \vspace{\mrgv}
        Input image & \hspace{\mrg}
        BUOL \cite{chu2023buol} & \hspace{\mrg}
        Uni-3D \cite{zhang2023uni} & \hspace{\mrg}
        InstPIFu \cite{liu2022towards}  & \hspace{\mrg}
        Ours
    \end{tabular}
    \caption{Qualitative results on the 3D-FRONT dataset \cite{front20213d}. The methods considered reconstruct full scenes including background regions and foreground instances. 
    } 
    \label{fig:front_qual_bg}
\end{figure*}

\subsection{Implementation details}
As the proposed framework is not constrained to specific modules, we leverage the significant progress made by the computer vision community in the recent years towards solving the different sub-tasks described above. 

In the scene analysis stage, we rely on CropFormer~\cite{qi2022open} for entity segmentation, OneFormer~\cite{jain2023oneformer} to separate them in foreground instances and background entities, and Perspective Fields~\cite{jin2023perspective} to estimate the camera calibration. 
We mainly use Marigold~\cite{ke2023repurposing} for depth estimation. 
However, the model predicts affine-invariant depth which differs by an unknown image-level offset and scale from the absolute physical units. 
During evaluations, we estimate these factors based on the ground truth depth available in the datasets to ensure that the reconstructions align with the target. 
For in-the-wild predictions, we empirically found that estimating the two unknowns of Marigold output based on a metric depth estimation, in our case, DepthAnything~\cite{depthanything}, achieves better results than using the latter by itself. 

In the instance processing stage, we perform amodal completion on the reprojected crops using our model obtained by fine-tuning Stable Diffusion v1.5~\cite{Rombach2021HighResolutionIS}. We also sample several points in the instance's mask and feed them to OVSAM~\cite{yuan2024ovsam} together with the input image to obtain the text prompt for guiding the diffusion. The completed object is then reconstructed using DreamGaussian~\cite{tang2023dreamgaussian}.
We estimate the camera elevation required by DreamGaussian as in~\cite{liu2023one}.
Then, we find the 3D points of the reconstruction that correspond to the unprojected instance points, which serve as our layout guide, and compute the scale which aligns them. Further specifications regarding the integrated models, possible alternatives for some of the processing stages and an analysis of the inference time of our method are provided in Section \ref{sup:impl_details} of the supplementary material.

\subsection{Results}
\begin{table}
    \centering
    \setlength{\tabcolsep}{0.2em}
    \begin{tabular}{l ccc}
        Method & 3D-supervision-free & Zero-shot & Compositional  \\
        \hline
        InstPIFu \cite{liu2022towards}& \xmark & \xmark & \cmark \\
        USL~\cite{gkioxari2022learning} & \cmark & \xmark & \cmark \\
        
        BUOL~\cite{chu2023buol} & \xmark & \xmark & \xmark \\
        Uni-3D~\cite{zhang2023uni} & \xmark & \xmark & \xmark \\
        
        DreamG \cite{tang2023dreamgaussian} & \cmark & \cmark & \xmark \\
        Ours & \cmark & \cmark & \cmark \\
        \hline 
    \end{tabular}
    \vspace{-0.15cm}
    \caption{As opposed to existing approaches, our method is at the same time compositional, able to generalize in a zero-shot manner, and does not require training with 3D supervision.}
    \label{tab:comparing_capabilities}
    \vspace{-0.6cm}
\end{table}

We showcase the performance of our method using several datasets across diverse scenarios. For numerical evaluation we first consider 3D-FRONT \cite{front20213d, future20213d}, a synthetic dataset of indoor rooms with available ground truth geometry. Due to the large scale of the dataset, we manually sample 100 images from the test split of \cite{liu2022towards}, avoiding the images with heavy scene occlusions (\eg, camera positioned behind a plant), intersecting objects, and scenes with very few objects. In addition, we use the 10 validation images of HOPE-Image \cite{tyree2022hope} dataset containing household objects captured under two scenarios. Though the dataset is intended for object pose evaluation, we find the ground-truth object alignment to match the input images well-enough for our purpose.  

We report the quantitative results using the widely-employed metrics for 3D reconstruction: Chamfer Distance \cite{barrow1977parametric} and F-Score \cite{knapitsch2017tanks}. Both are computed between densely sampled sets of points from the reconstructed meshes and the ground truth respectively. As we focus the evaluation on whole scenes, the points are uniformly sampled from the entire geometry of a scene. 

We compare our method against several solutions covering multiple approaches for 3D scene reconstruction. An overview of their capabilities is presented in Table~\ref{tab:comparing_capabilities}. 
BUOL~\cite{chu2023buol} and Uni-3D \cite{zhang2023uni} are both feed-forward scene reconstruction methods that have been trained with 3D supervision on the 3D-FRONT dataset. 
While we evaluate the methods on the same dataset they were trained on, we use a more realistic rendering, following \cite{liu2022towards}. 
Even under this minor change, the methods' performance degrades significantly, as can be seen in Table~\ref{tab:front_quant} and Figure~\ref{fig:front_qual_bg}, showing their lack of generalization.

\begin{figure*}
    \centering    
    \setlength{\wid}{0.195\textwidth}
    \setlength{\mrg}{-0.35cm}
    \setlength{\mrgv}{-0.1cm}
    \begin{tabular}{ccccc}
        \includegraphics[width=\wid]{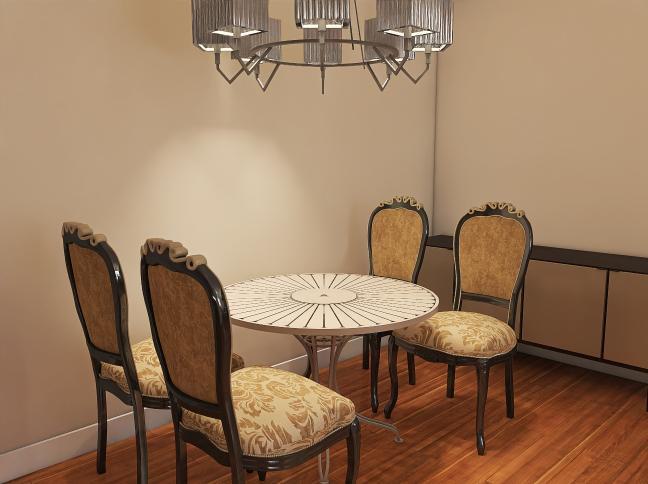} &
        \hspace{\mrg}
        \includegraphics[width=\wid]{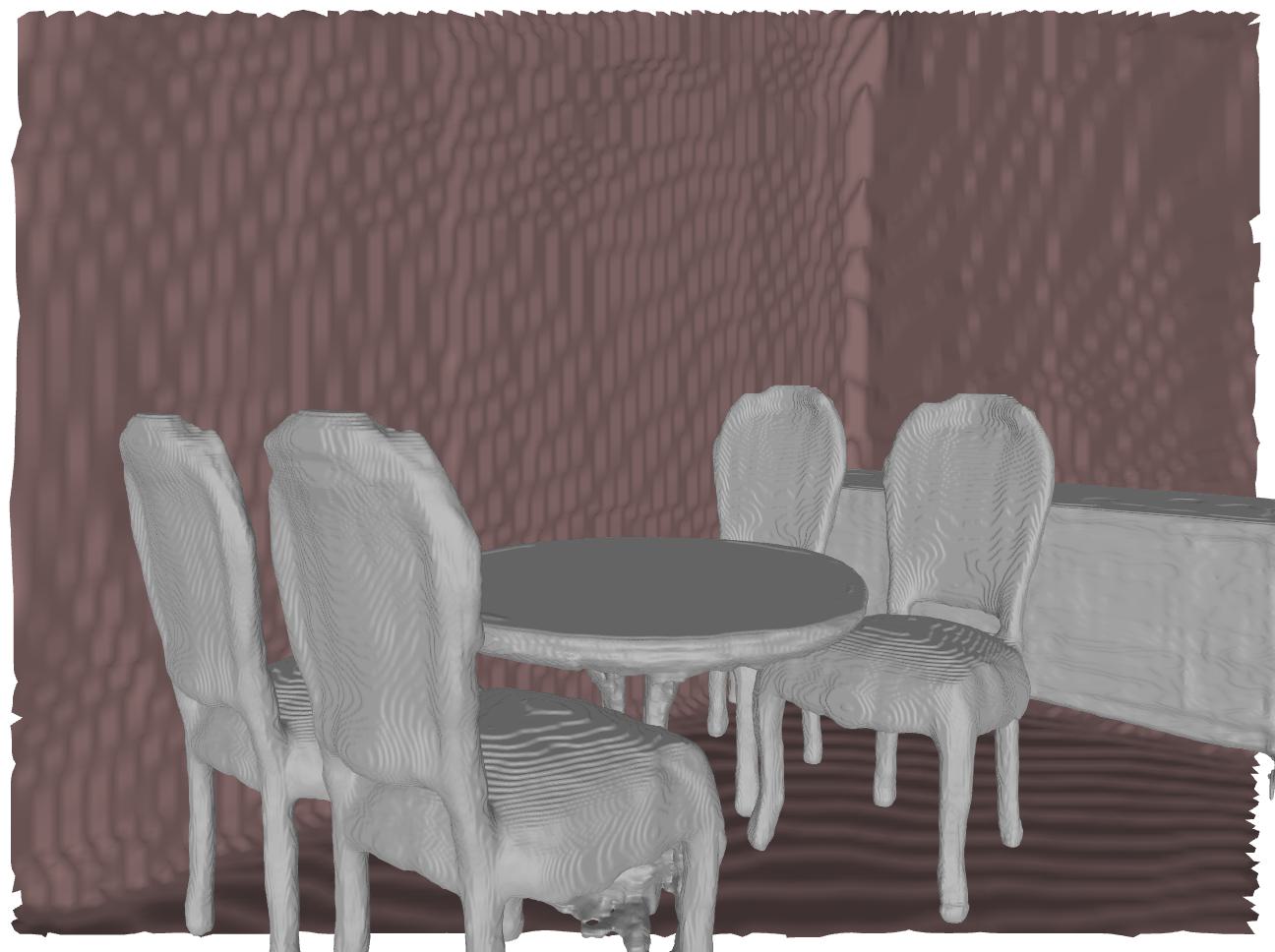} &
        \hspace{\mrg}
        \includegraphics[width=\wid]{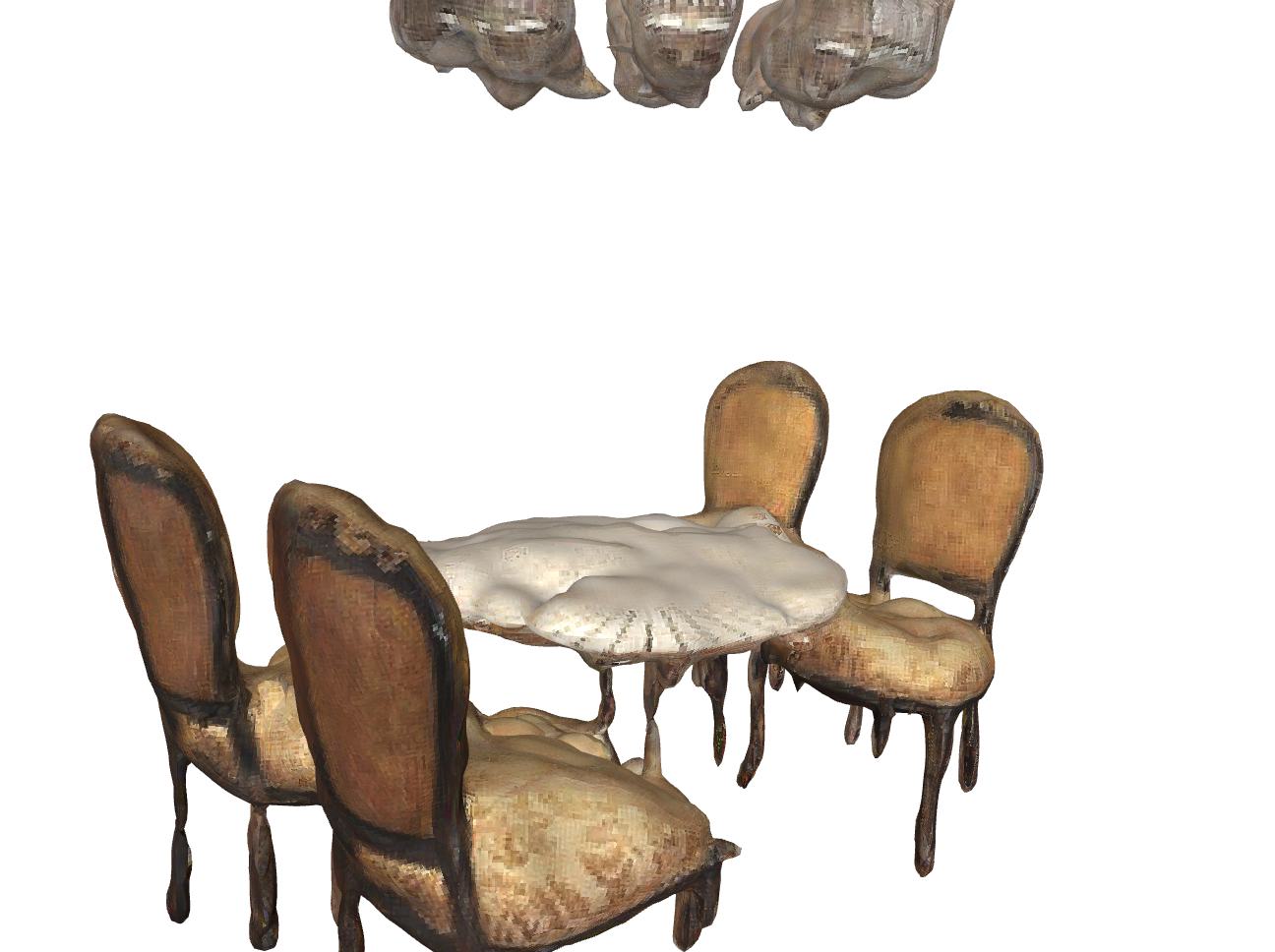} & 
        \hspace{\mrg}
        \includegraphics[width=\wid]{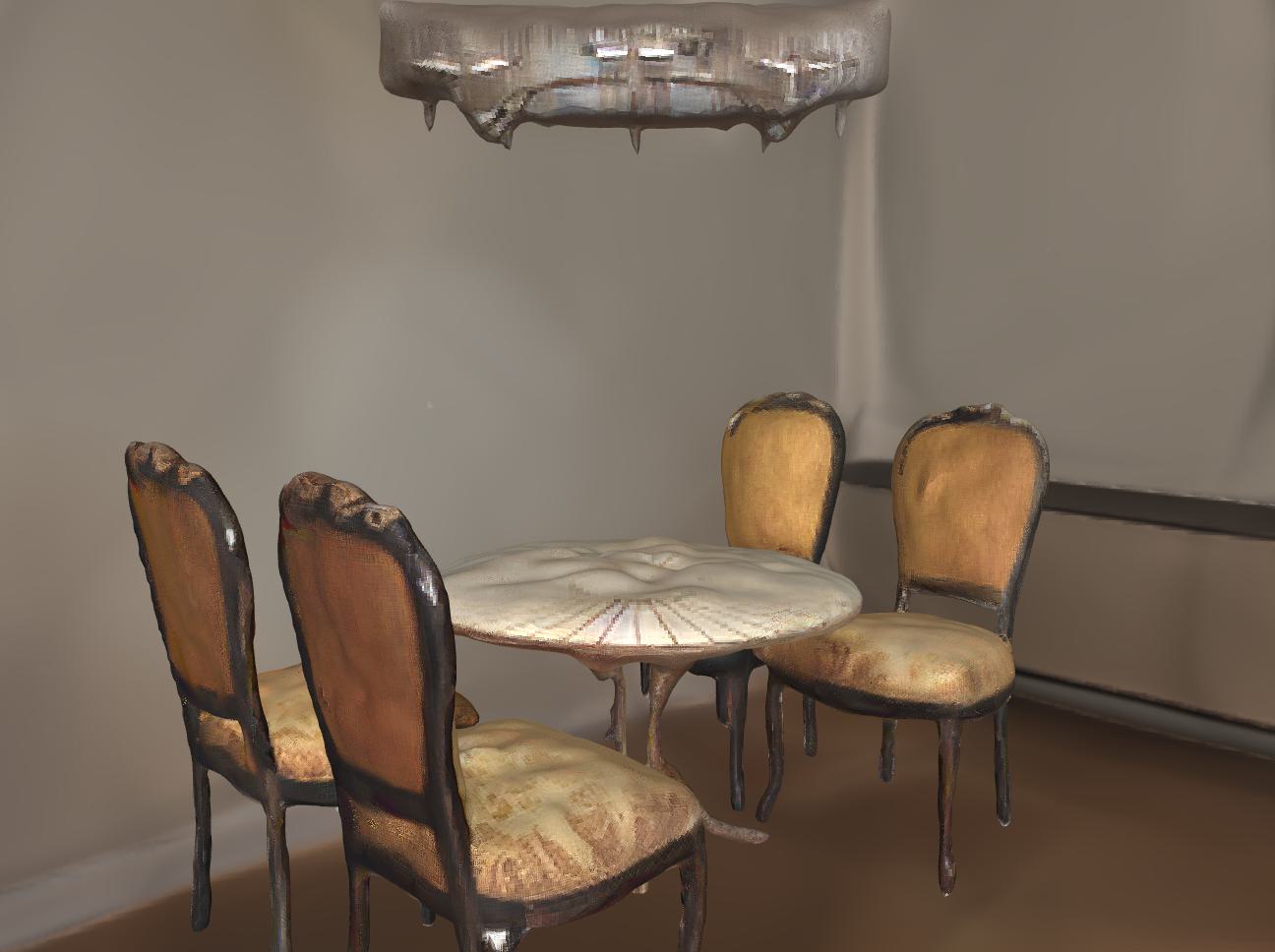} & 
        \hspace{\mrg}
        \includegraphics[width=\wid]{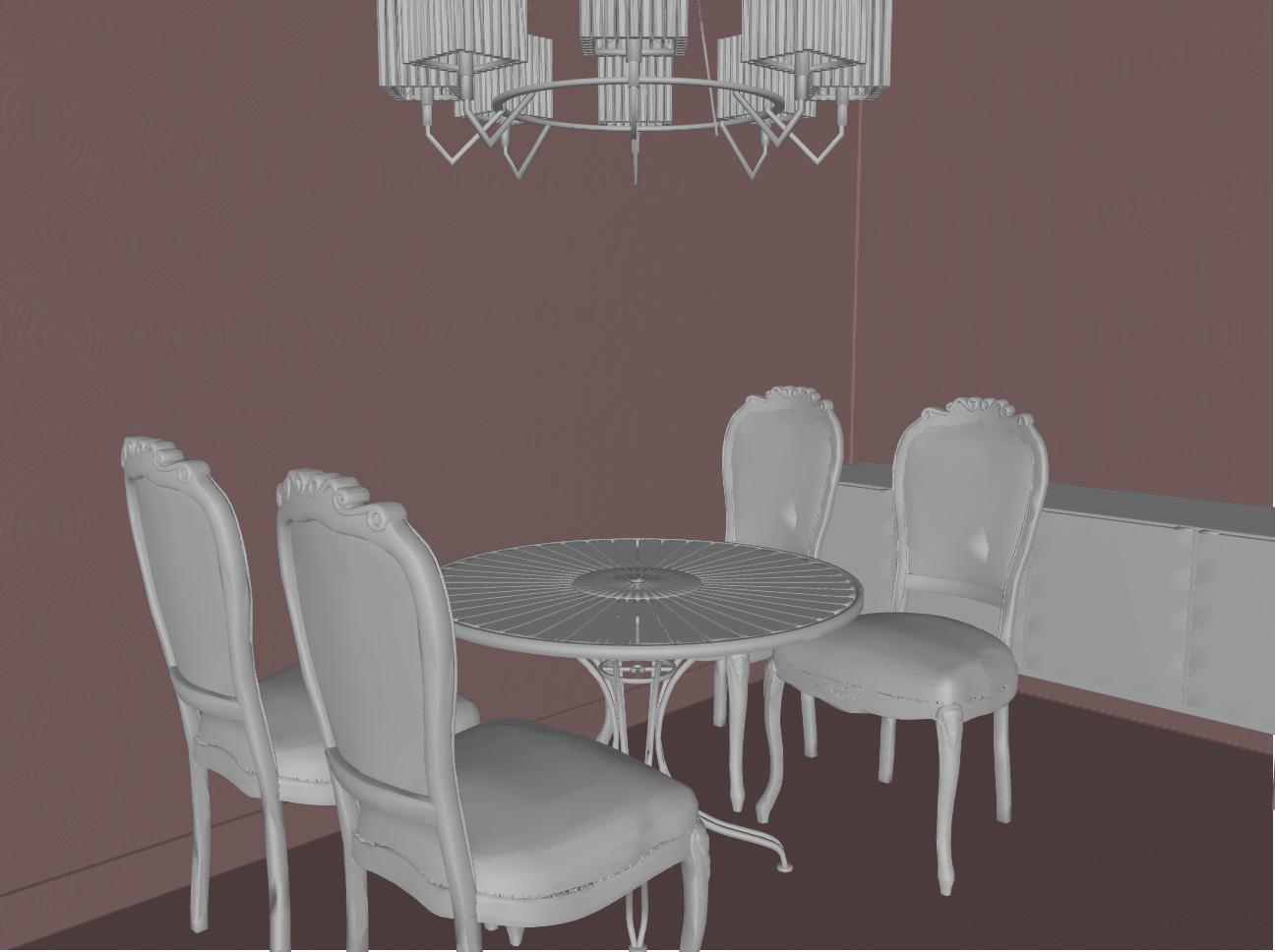} 
        \\ 
         &
        \hspace{\mrg}
        \includegraphics[width=\wid]{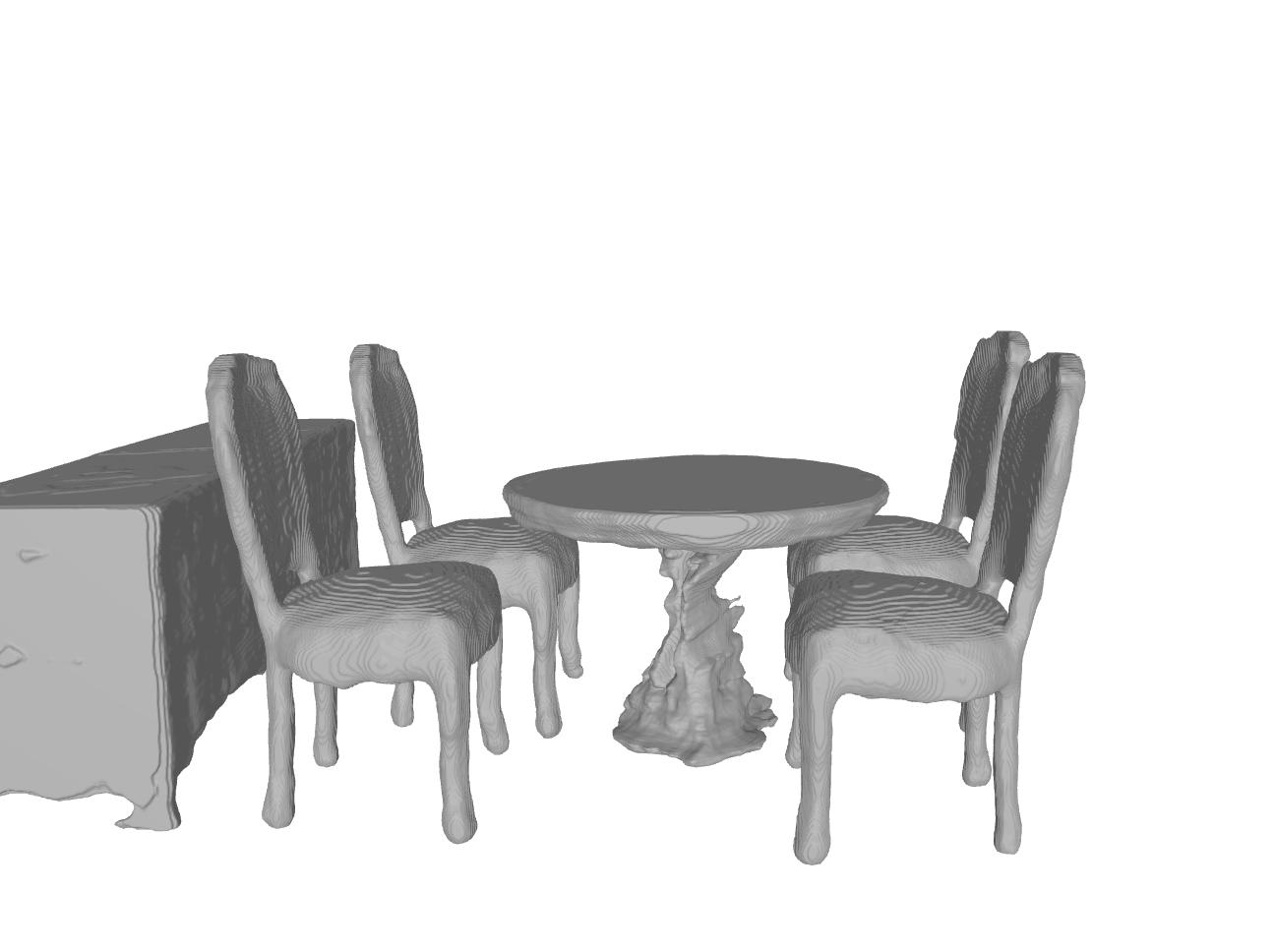} &
        \hspace{\mrg}
        \includegraphics[width=\wid]{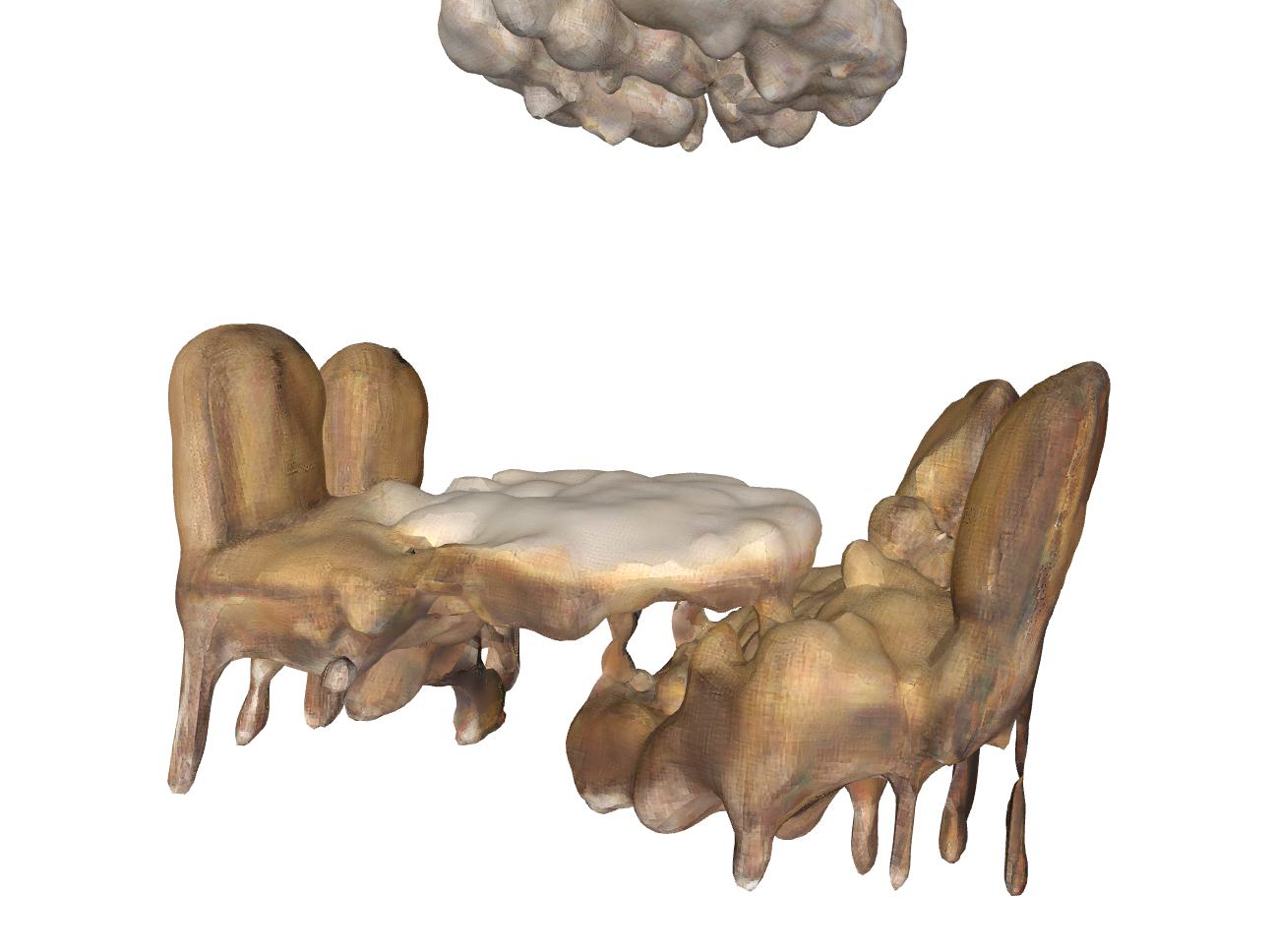} & 
        \hspace{\mrg}
        \includegraphics[width=\wid]{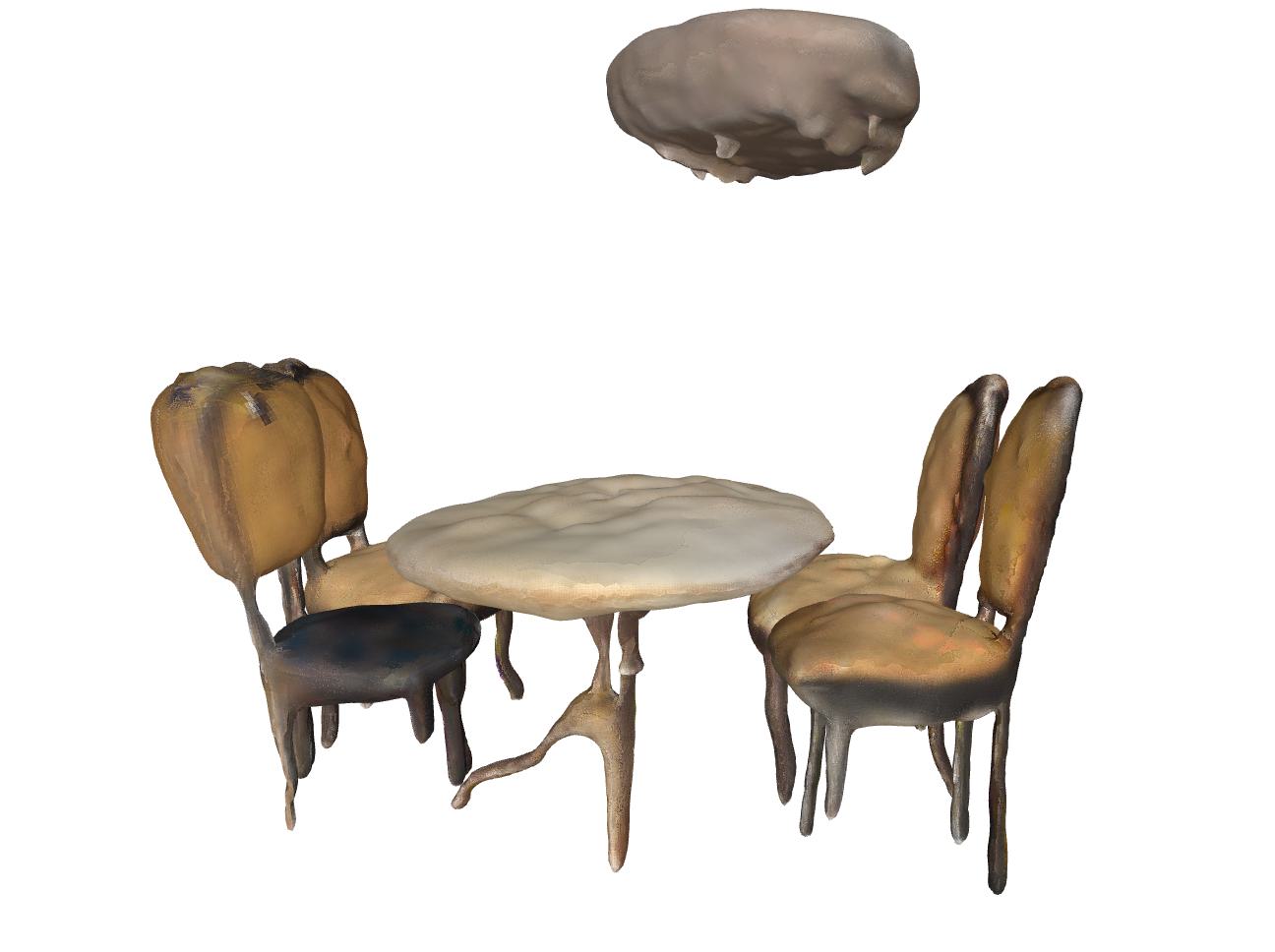} & 
        \hspace{\mrg}
        \includegraphics[width=\wid]{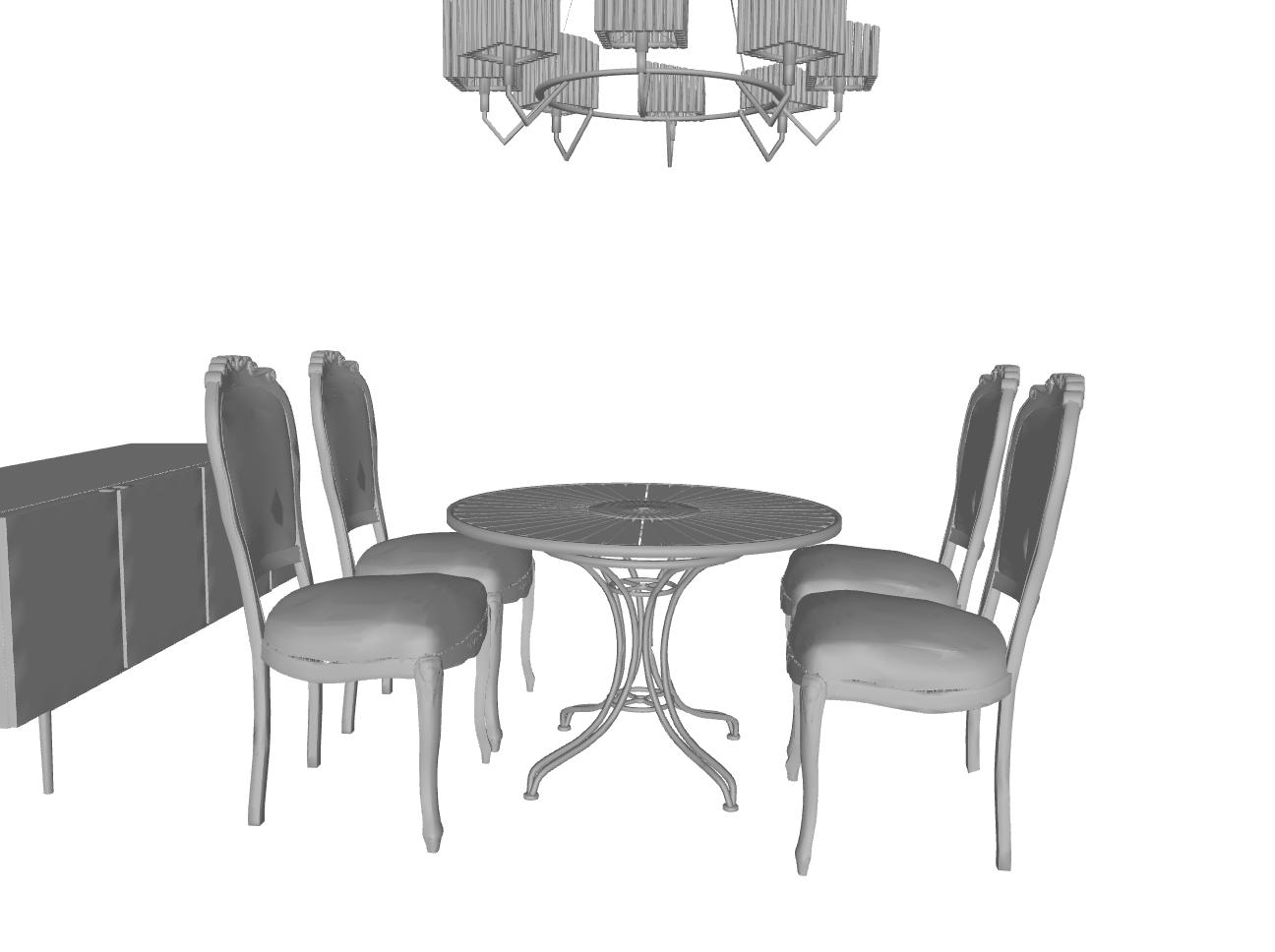} \\
        \includegraphics[width=\wid]{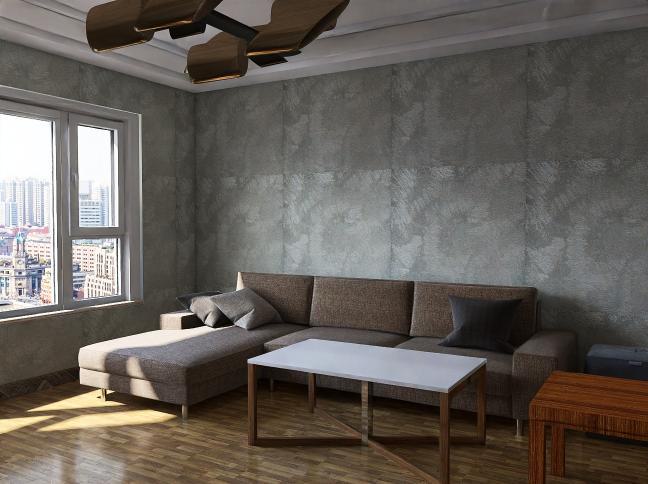} &
        \hspace{\mrg}
        \includegraphics[width=\wid]{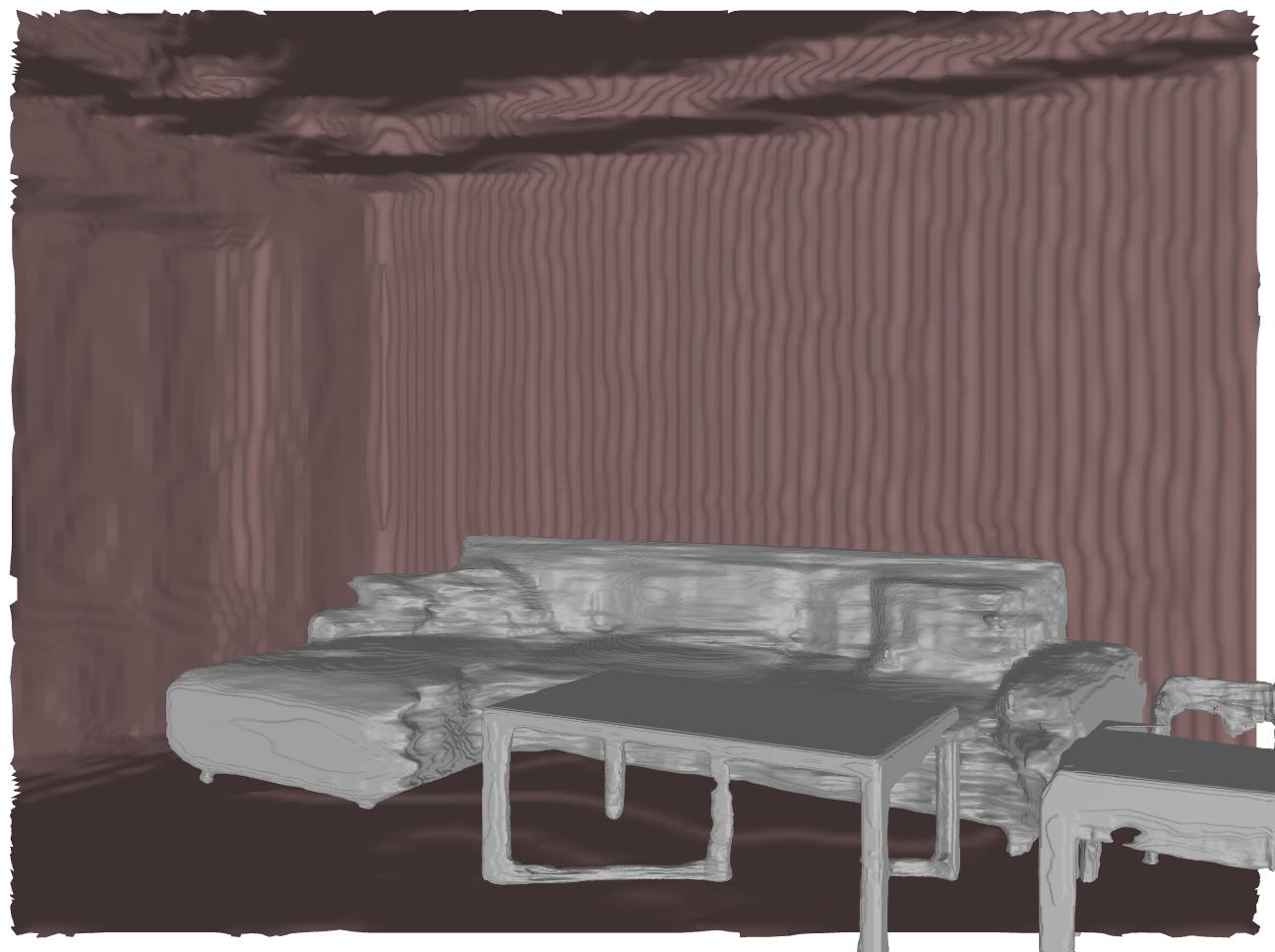} &
        \hspace{\mrg}
        \includegraphics[width=\wid]{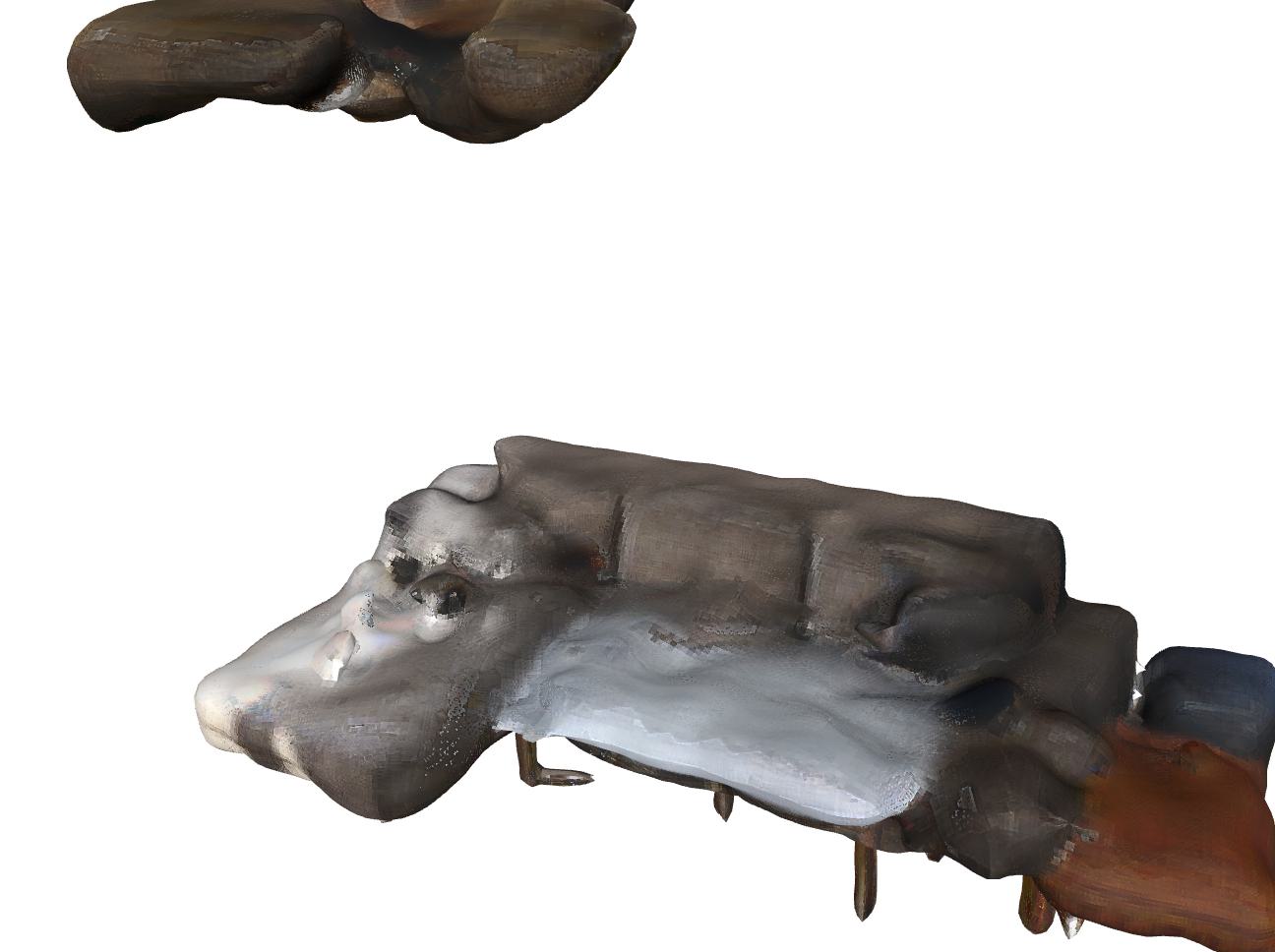} & 
        \hspace{\mrg}
        \includegraphics[width=\wid]{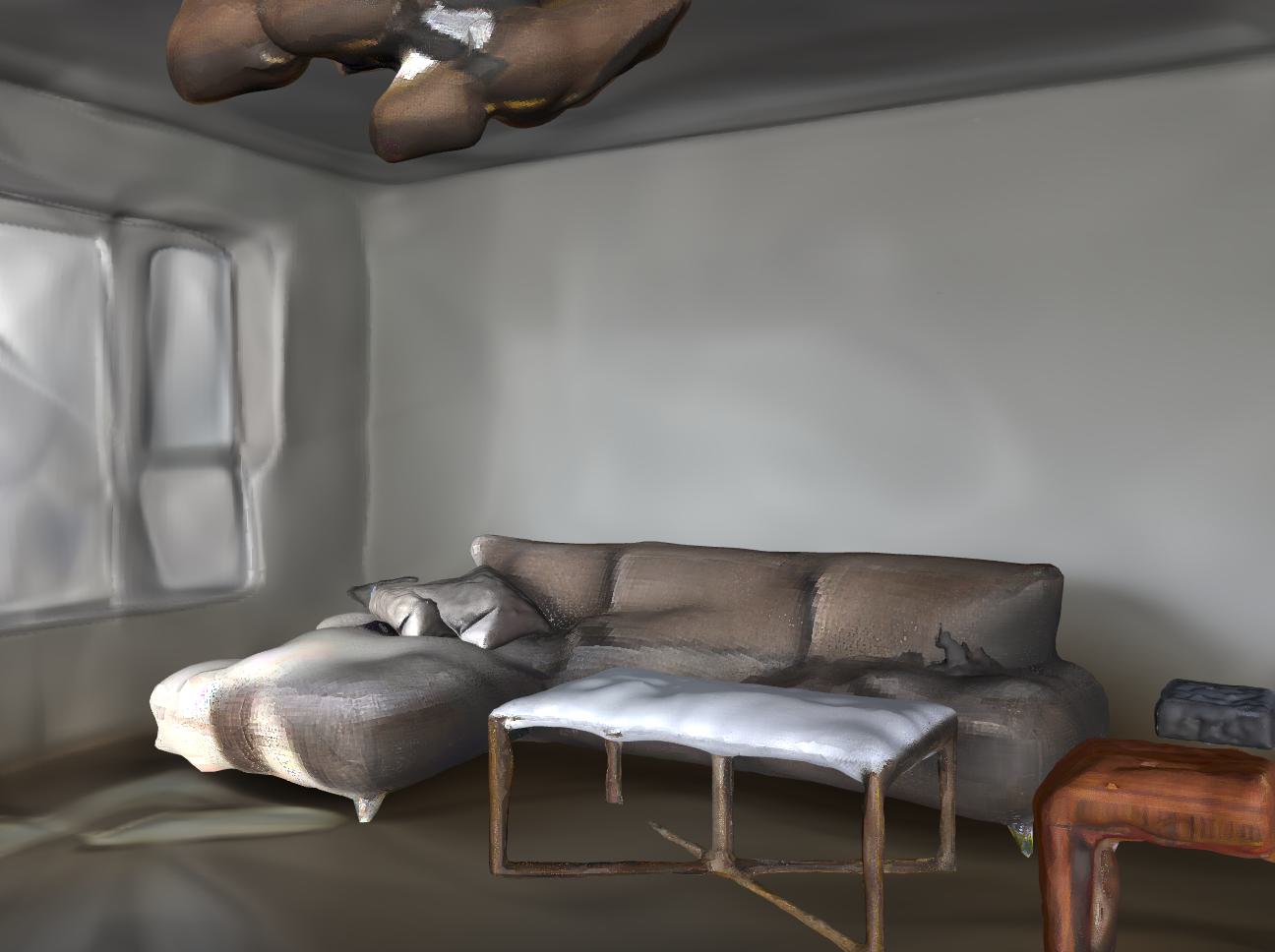} & 
        \hspace{\mrg}
        \includegraphics[width=\wid]{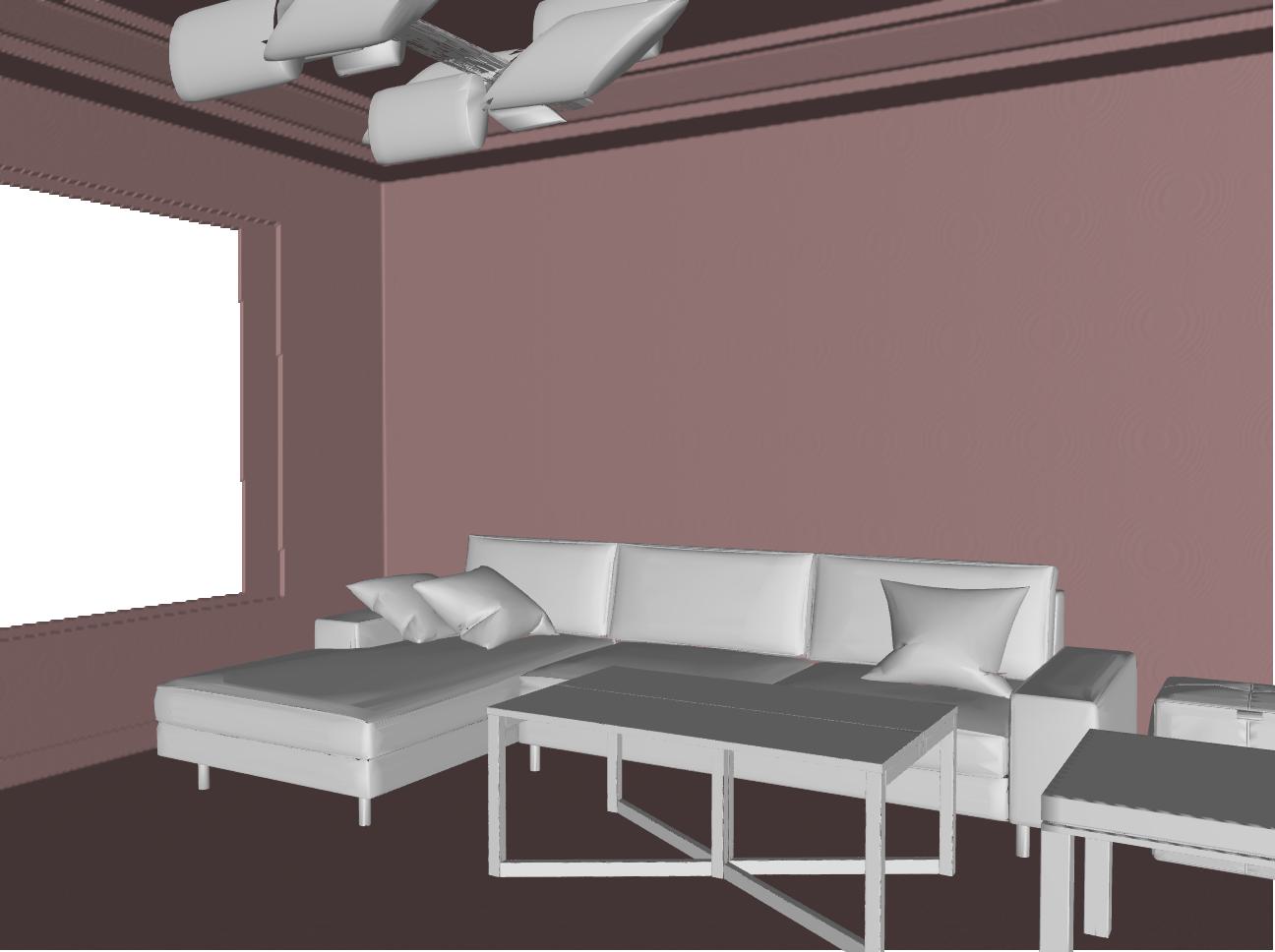} \\
         &
        \hspace{\mrg}
        \includegraphics[width=\wid]{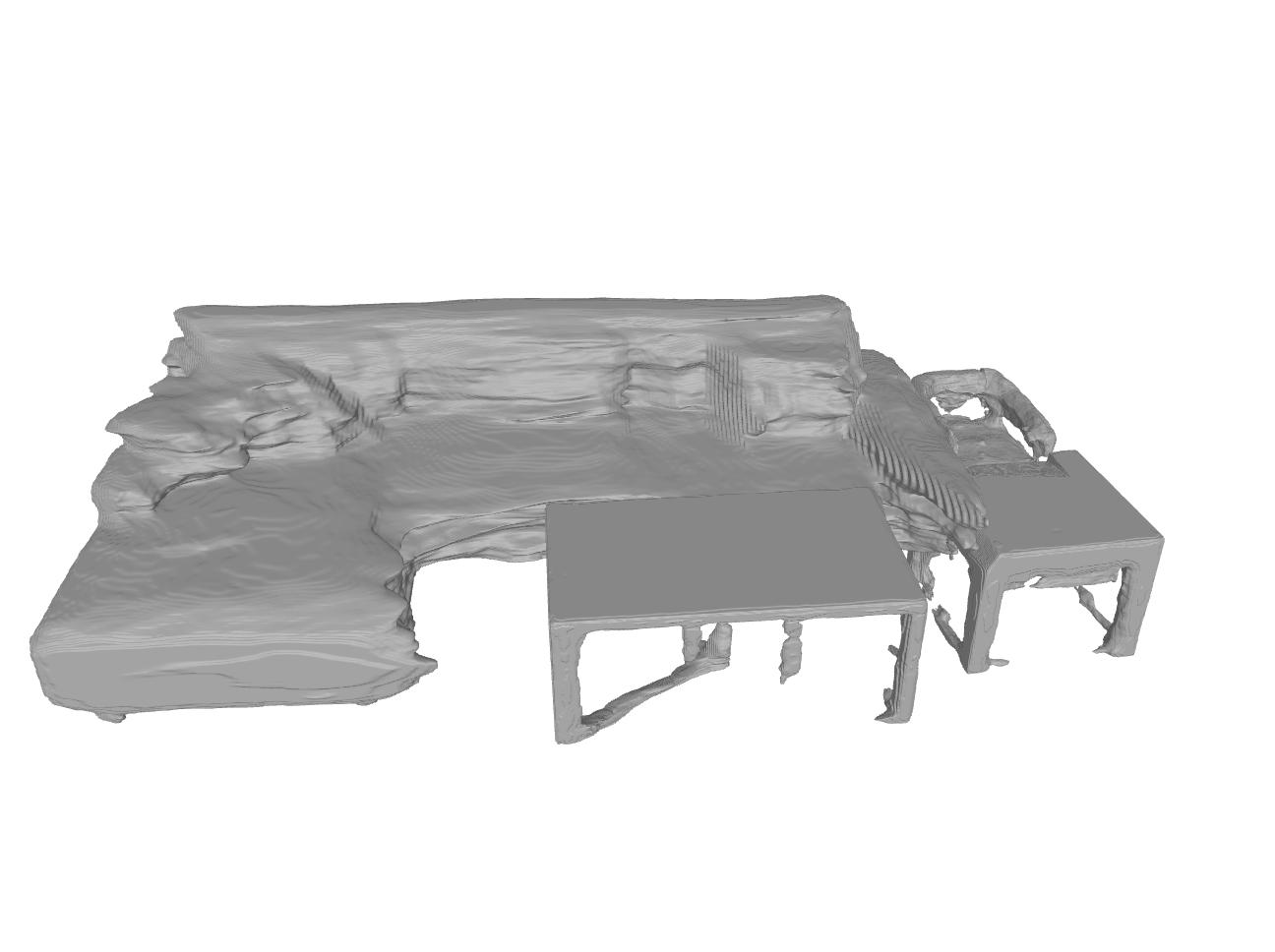} &
        \hspace{\mrg}
        \includegraphics[width=\wid]{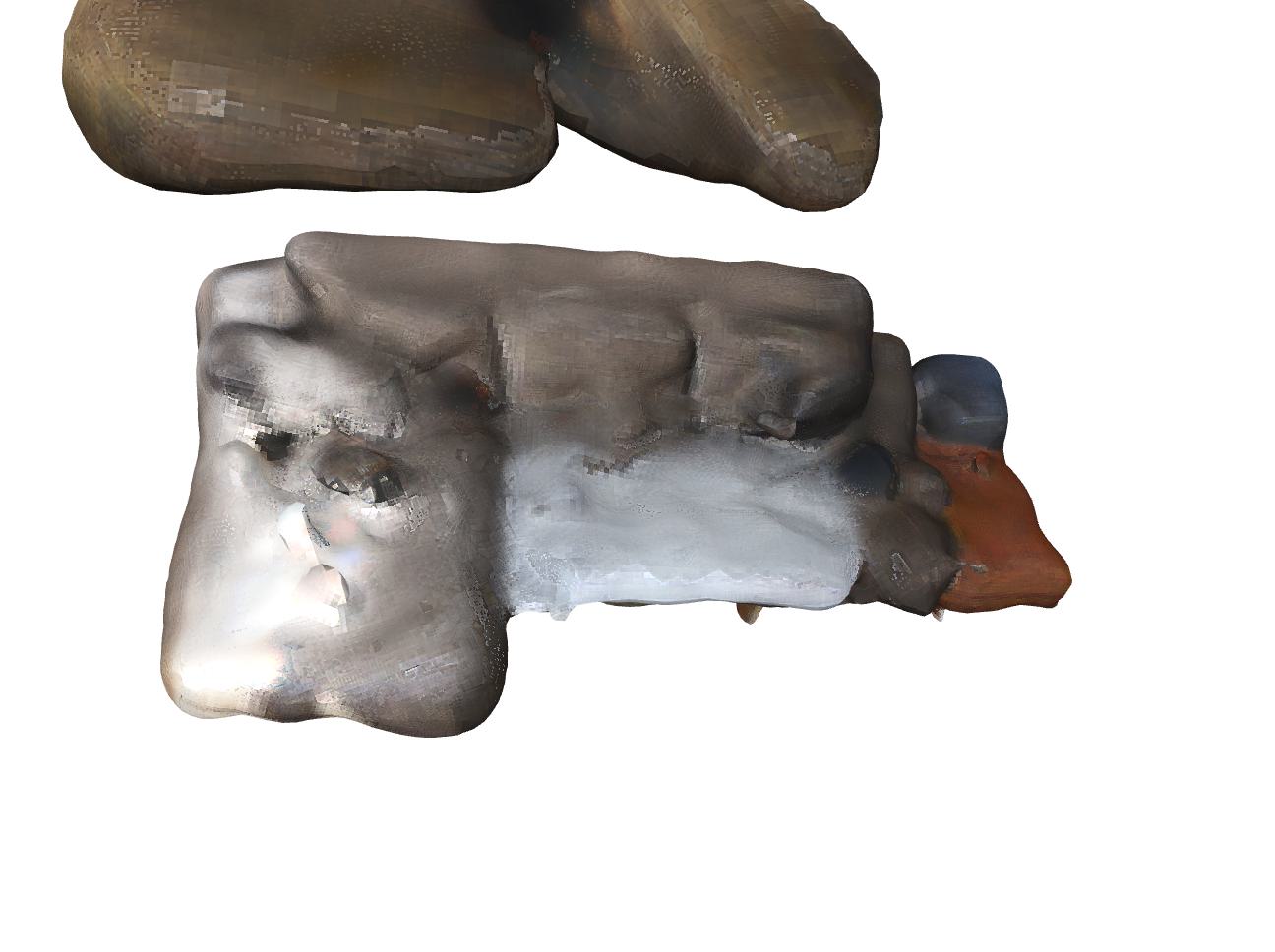} & 
        \hspace{\mrg}
        \includegraphics[width=\wid]{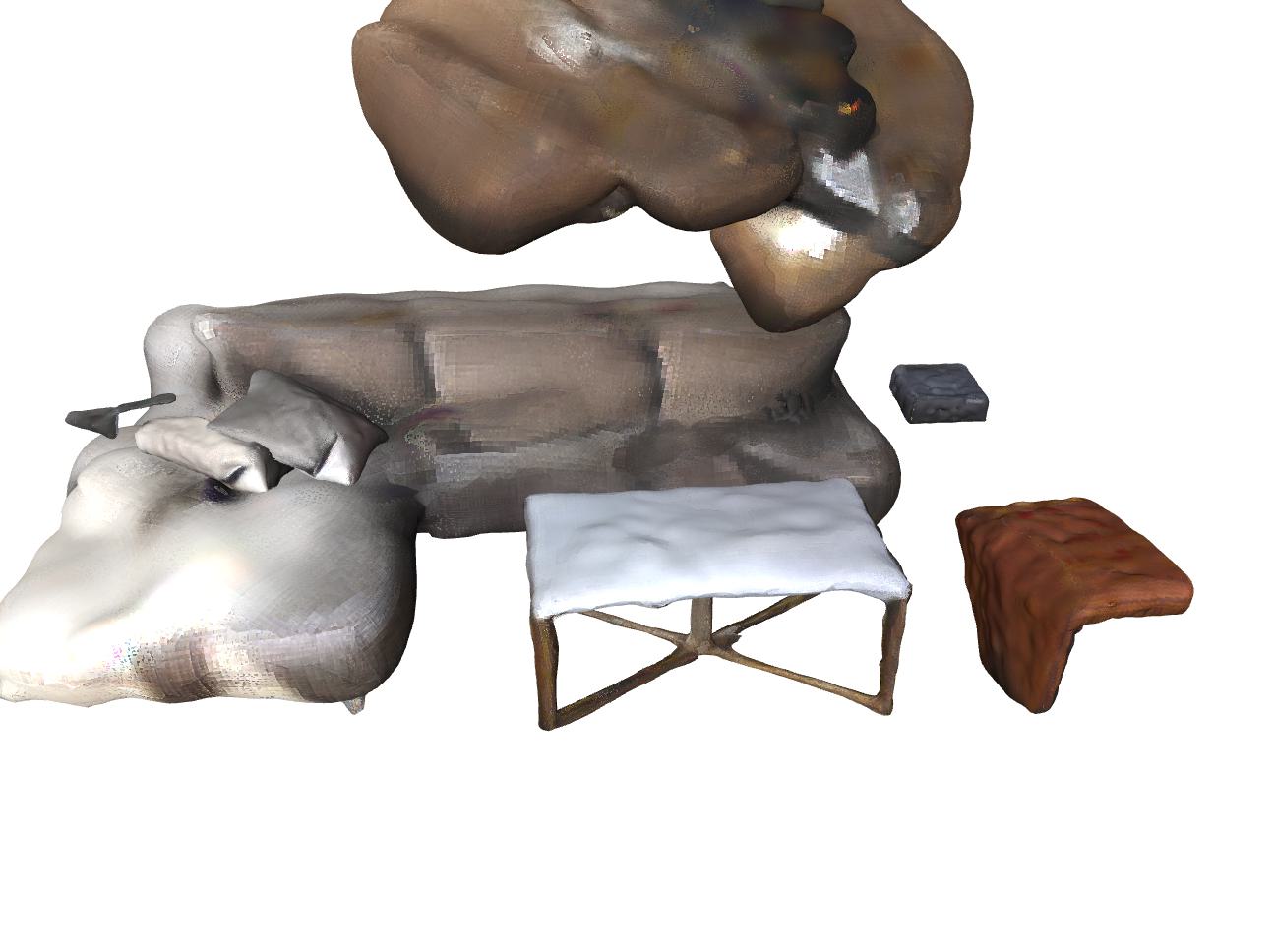} & 
        \hspace{\mrg}
        \includegraphics[width=\wid]{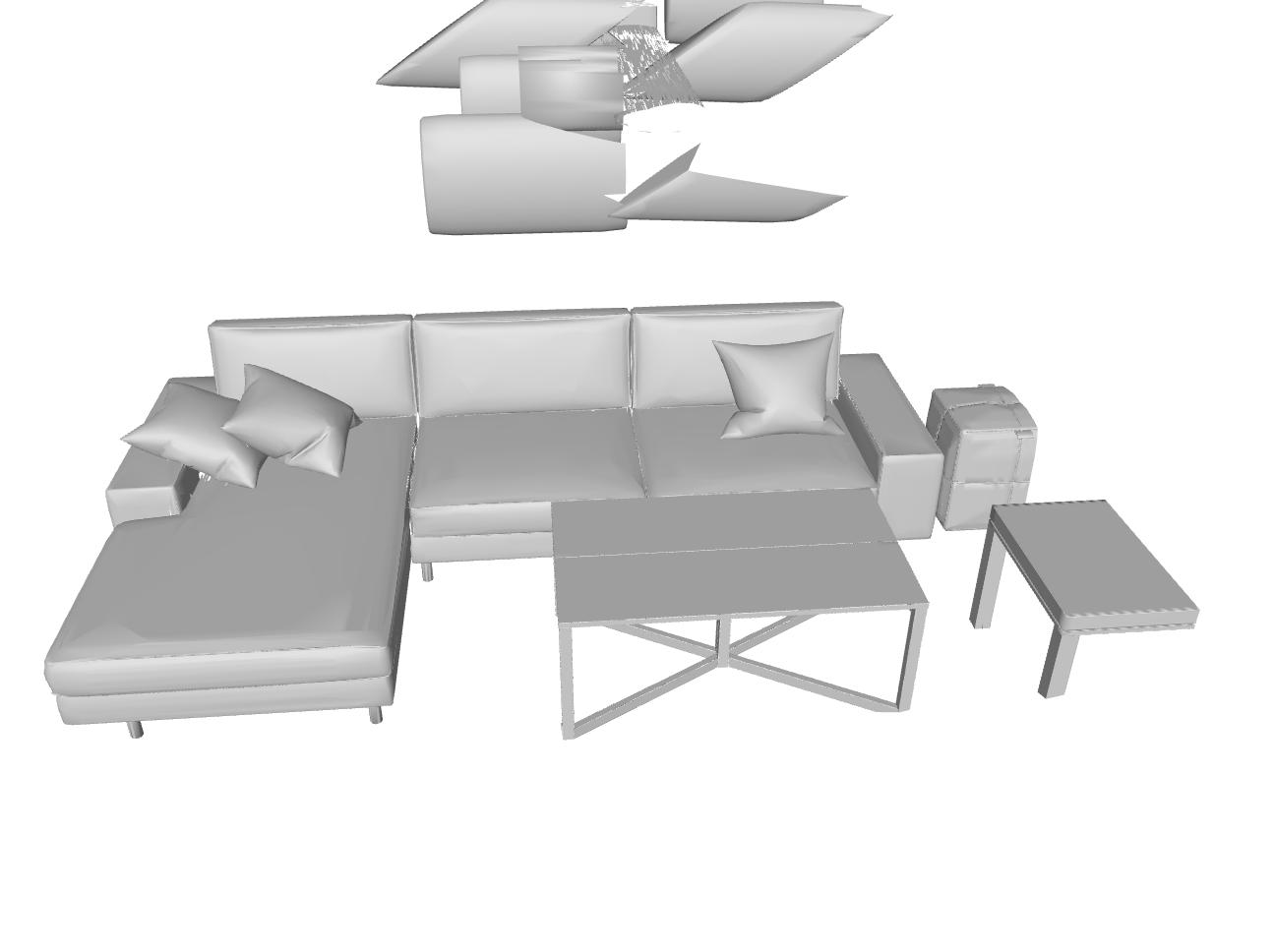} \\
        \vspace{\mrgv}
        Input image & \hspace{\mrg}
        InstPIFu \cite{liu2022towards} & \hspace{\mrg}DreamGaussian \cite{tang2023dreamgaussian} & \hspace{\mrg}
        Ours  
        & \hspace{\mrg}
        GT geometry
    \end{tabular}
    \caption{Qualitative results on 3D-FRONT \cite{front20213d} illustrated under the input view and a second one chosen to highlight the reconstruction performance. In contrast to Ours and InstPIFu, DreamGaussian reconstructs all the objects at once and does not model the background. 
    } 
    \label{fig:front_qual_fg}
    \vspace{-0.3cm}
\end{figure*}

\begin{table}
    \centering
    \setlength{\tabcolsep}{0.2em}
    \begin{tabular}{l cc|cc}
        \\
        & \multicolumn{2}{c}{3D-FRONT \cite{front20213d}} & \multicolumn{2}{c}{HOPE-Image \cite{tyree2022hope}}
        \\
        Method & Chamfer $\downarrow$ & F-Score $\uparrow$ & Chamfer $\downarrow$ & F-Score $\uparrow$
        \\
        \hline
        InstPIFu \cite{liu2022towards} & 0.124 & \textbf{74.14}  & - & - \\
        DreamG \cite{tang2023dreamgaussian} & 0.207 & 41.09 & 3.038 & 29.57 \\
        \setlength{\tabcolsep}{-0.25em}
        \begin{tabular}{l}DreamG $+$\\ reprojection \end{tabular} & 0.187 & 49.98 & 4.059 & 30.62 \\
        Ours & \textbf{0.120} & 68.82 & \textbf{1.446} & \textbf{54.82} \\
        \hline
    \end{tabular}
    \vspace{-0.15cm}
    \caption{Quantitative results on foreground instances reconstruction. 
    On 3D-FRONT, we are on par with InstPIFu, despite it being trained with 3D data on this dataset. On HOPE-Image, InstPIFu fails to recognize any of the evaluated objects.
    On both datasets, we outperform the non-compositional approach, DreamGaussian.
    }
    \label{tab:hope_quant}
    \vspace{-0.6cm}
\end{table}

\begin{table}
    \centering
    \begin{tabular}{l cc}
        \\
        Method & Chamfer $\downarrow$ & F-Score $\uparrow$
        \\
        \hline
        InstPIFu \cite{liu2022towards} & 0.119 & 70.63 \\
        BUOL \cite{chu2023buol} & 0.294 & 37.04 \\
        Uni-3D \cite{zhang2023uni} & 0.448 & 32.97 \\
        Ours & \textbf{0.099} & \textbf{75.33} \\
        \hline
    \end{tabular}
    \vspace{-0.15cm}
    \caption{Quantitative evaluation on full scene reconstruction including background and foreground elements on the 3D-FRONT dataset \cite{front20213d}. Our method, which performs in a zero-shot fashion, outperforms all the baselines, which have been trained using 3D supervision on a version of this dataset. 
    }
    \label{tab:front_quant}
    \vspace{-0.6cm}
\end{table}

\begin{figure*}
    \centering    
    \setlength{\wid}{0.215\textwidth}
    \setlength{\mrg}{-0.3cm}
    \setlength{\mrgv}{0cm}
    \begin{tabular}{cccc}
        \includegraphics[width=\wid]{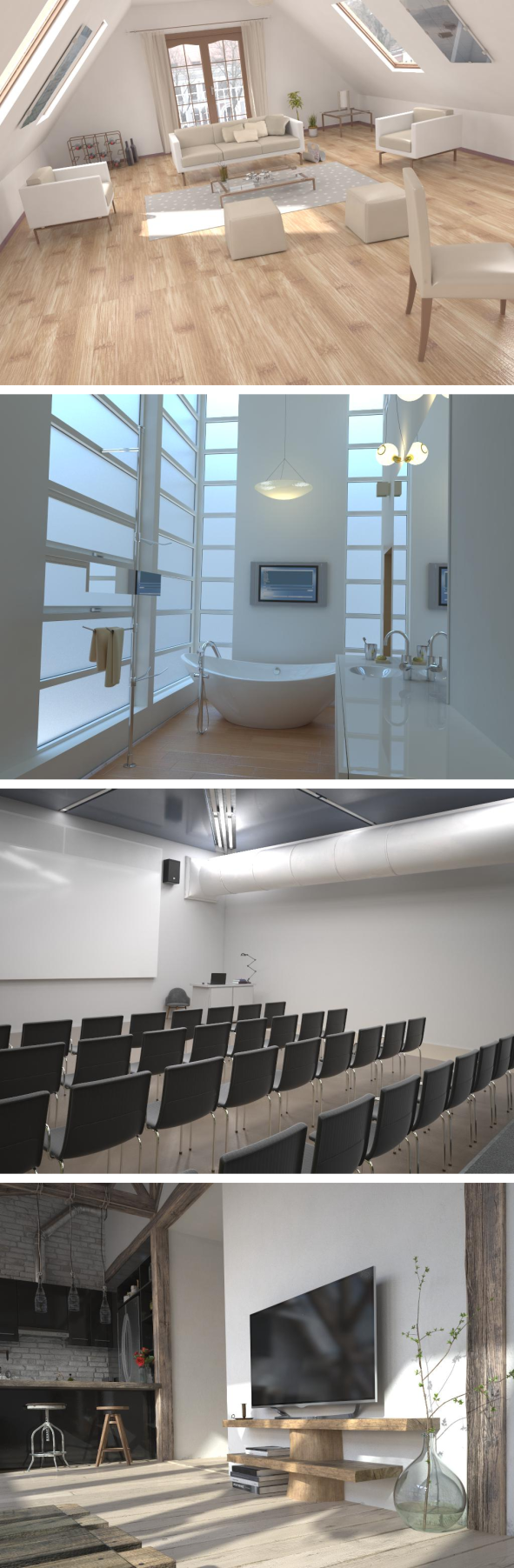} &
        \hspace{\mrg}
        \includegraphics[width=\wid]{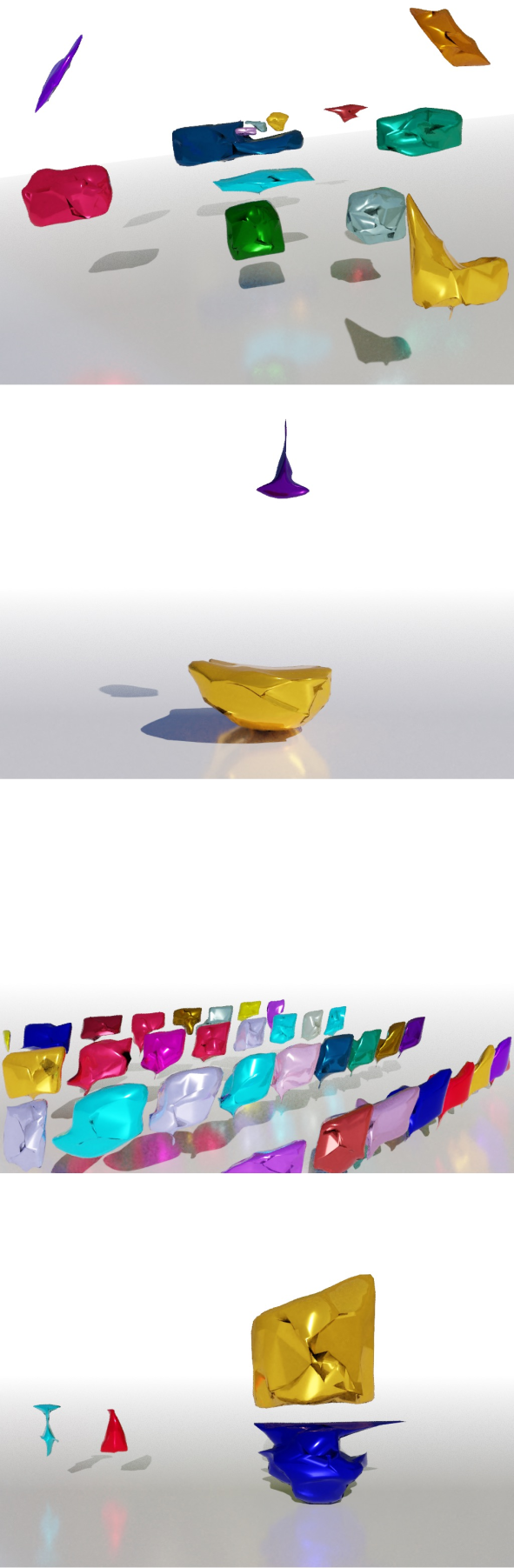} &
        \hspace{\mrg}
        \includegraphics[width=\wid]{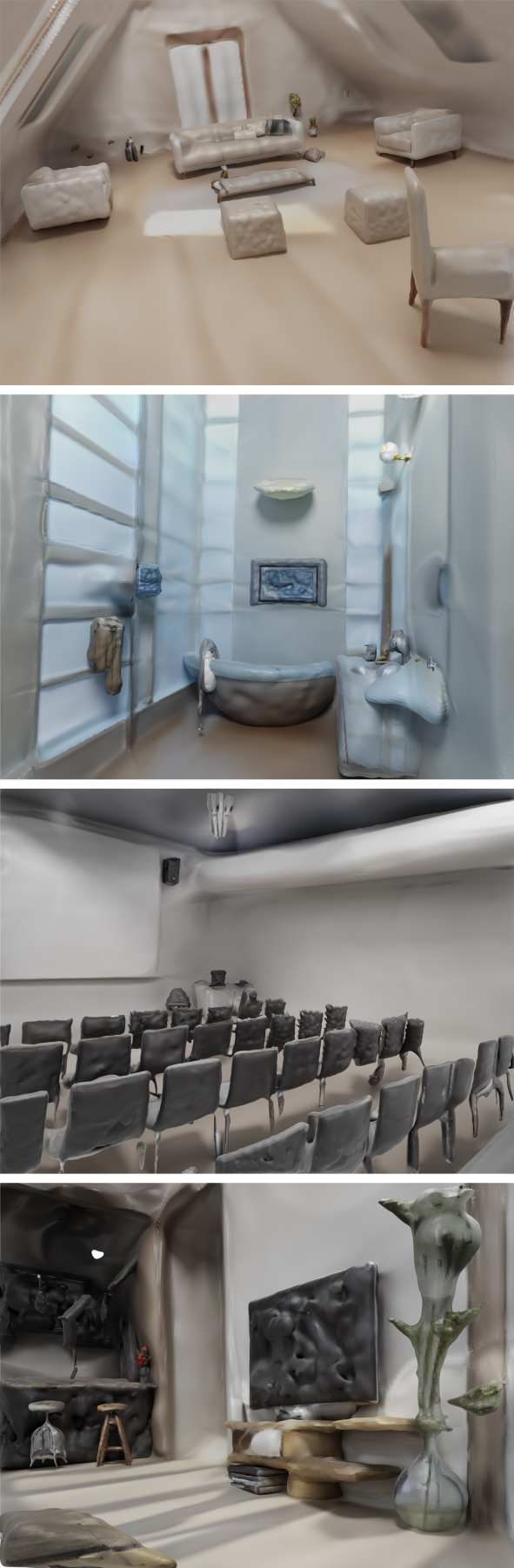} & 
        \hspace{\mrg}
        \includegraphics[width=\wid]{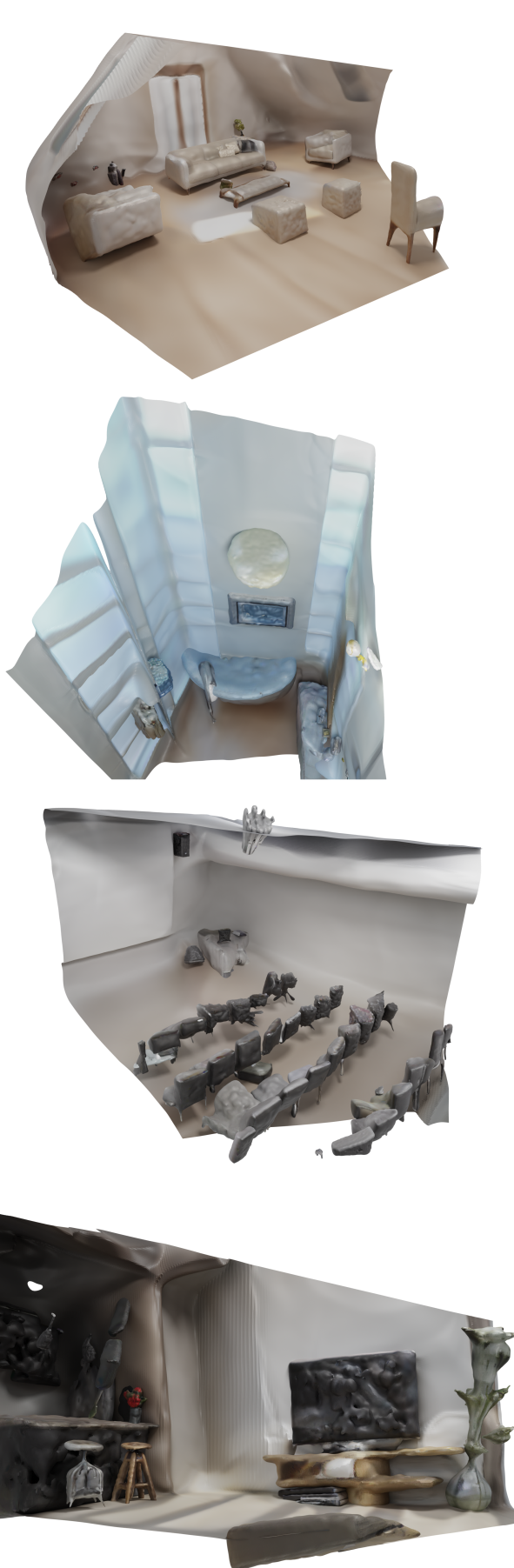} 
        \\ 
        \vspace{\mrgv}
        Input image & \hspace{\mrg}
        USL \cite{gkioxari2022learning} & \hspace{\mrg}
        Ours (input view) & \hspace{\mrg}
        Ours (another view)  
    \end{tabular}
    \vspace{-0.2cm}
    \caption{Qualitative results on the Hypersim dataset~\cite{roberts_2021_hyper}. We compare against the USL method which also does not require 3D data for supervision. Our scenes are more complete, with significantly more objects being recognized and reconstructed.
    } 
    \label{fig:hyper_qual}
    \vspace{-0.3cm}
\end{figure*}

\begin{figure*}
    \centering    
    \setlength{\wid}{0.215\textwidth}
    \setlength{\mrg}{-0.3cm}
    \setlength{\mrgv}{0cm}
    \begin{tabular}{cccc}
        \includegraphics[width=\wid]{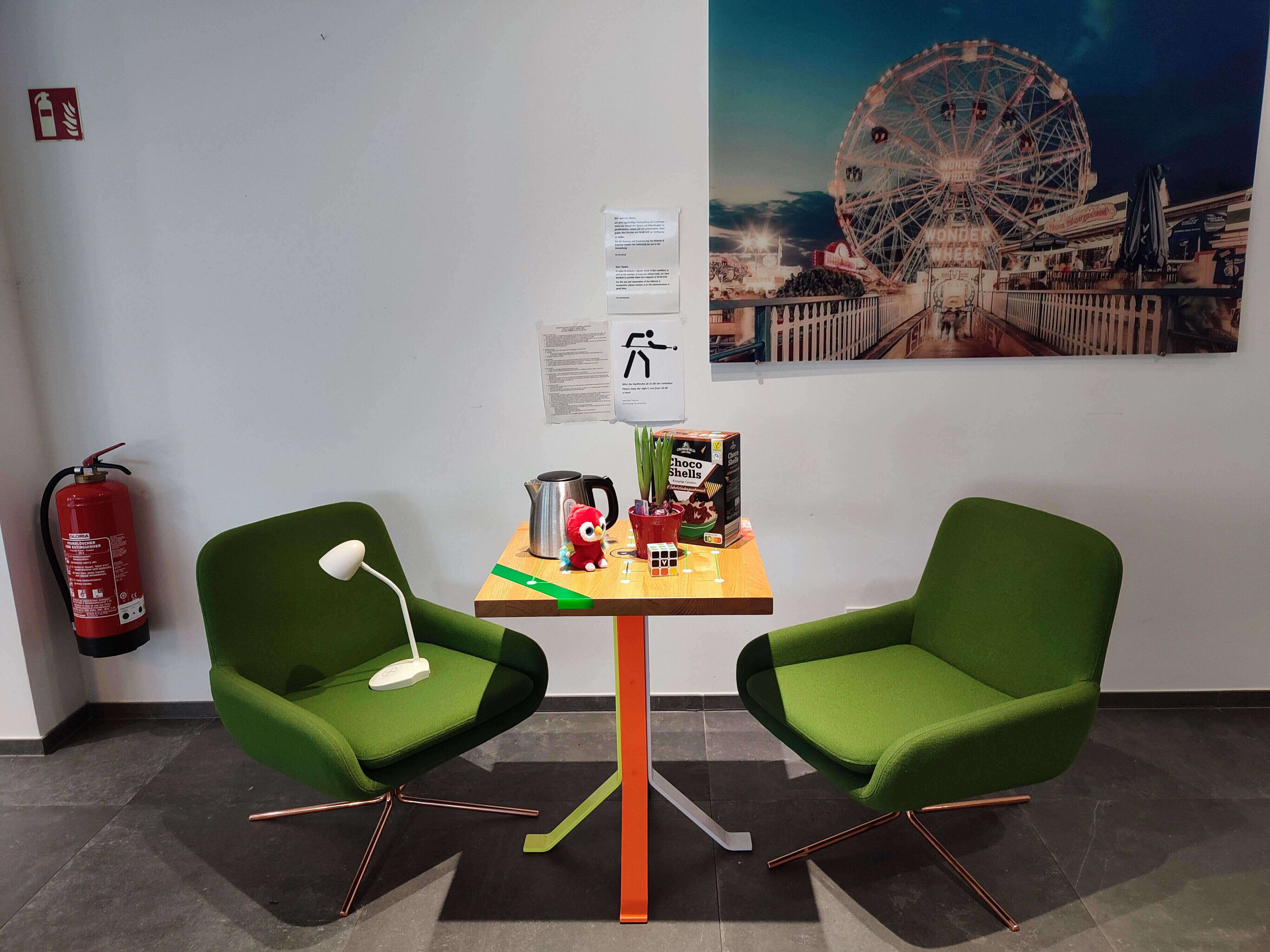} &
        \hspace{\mrg}
        \includegraphics[width=\wid]{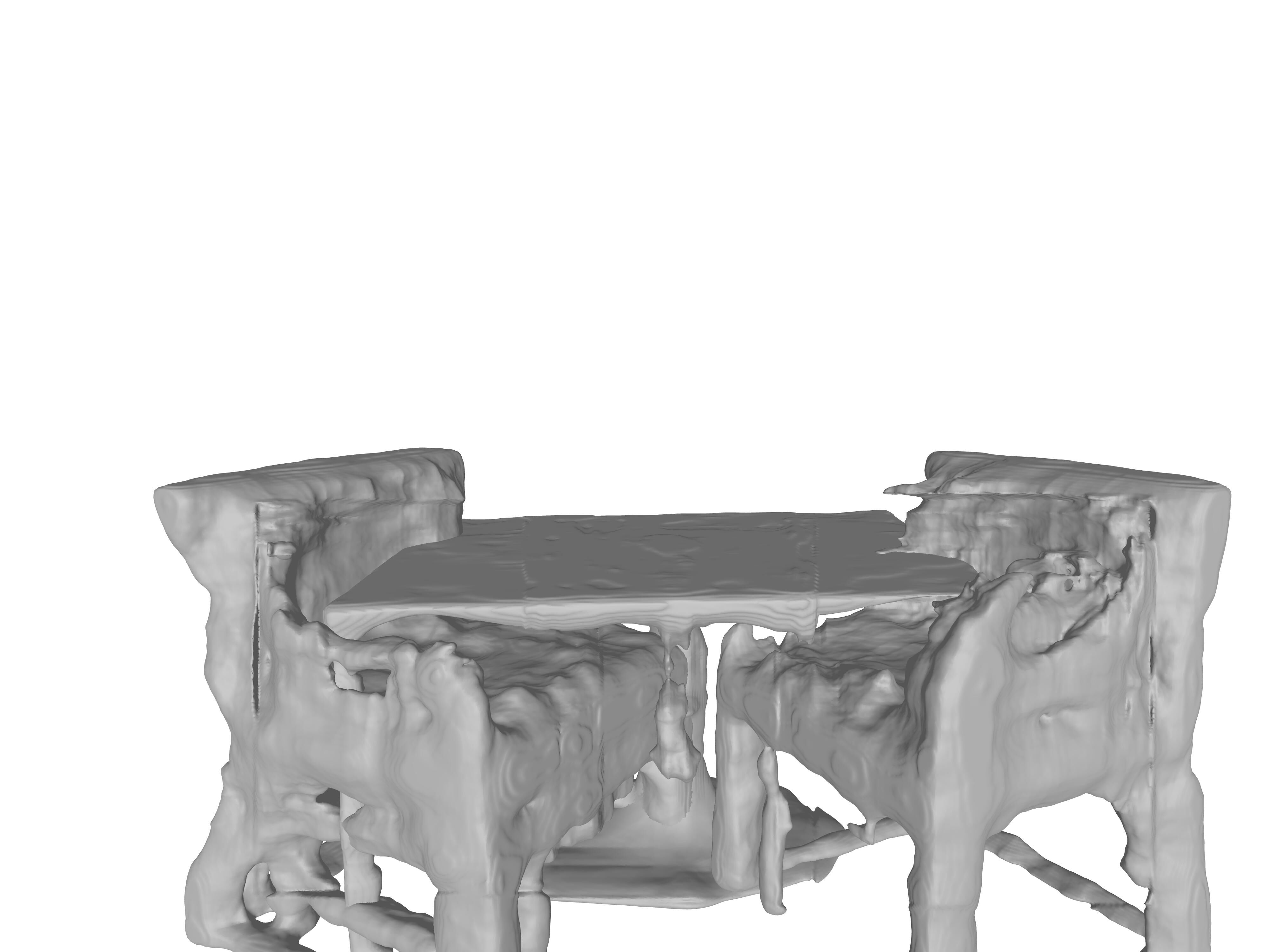}  &
        \hspace{\mrg}
        \includegraphics[width=\wid]{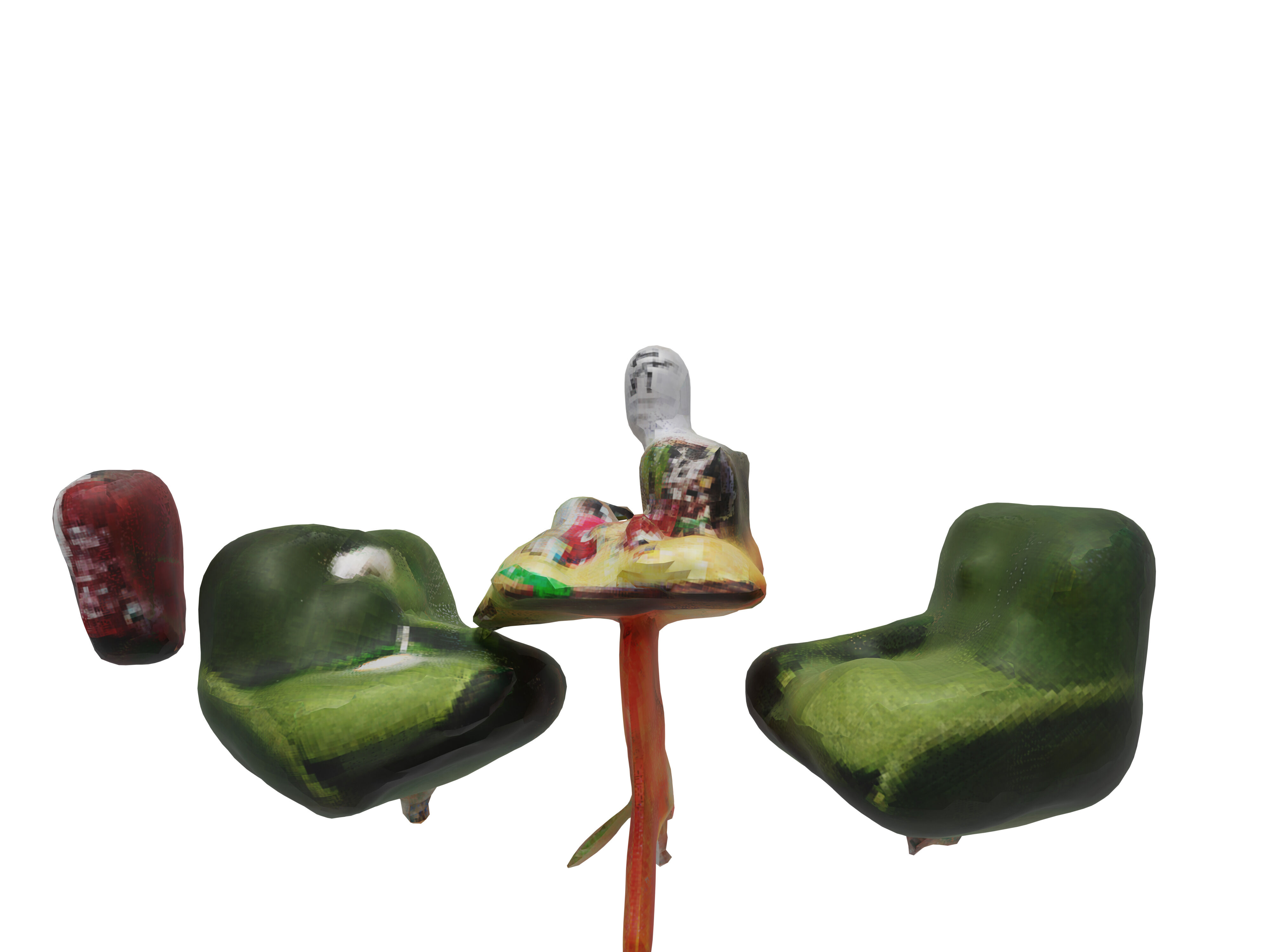} & 
        \hspace{\mrg}
        \includegraphics[width=\wid]{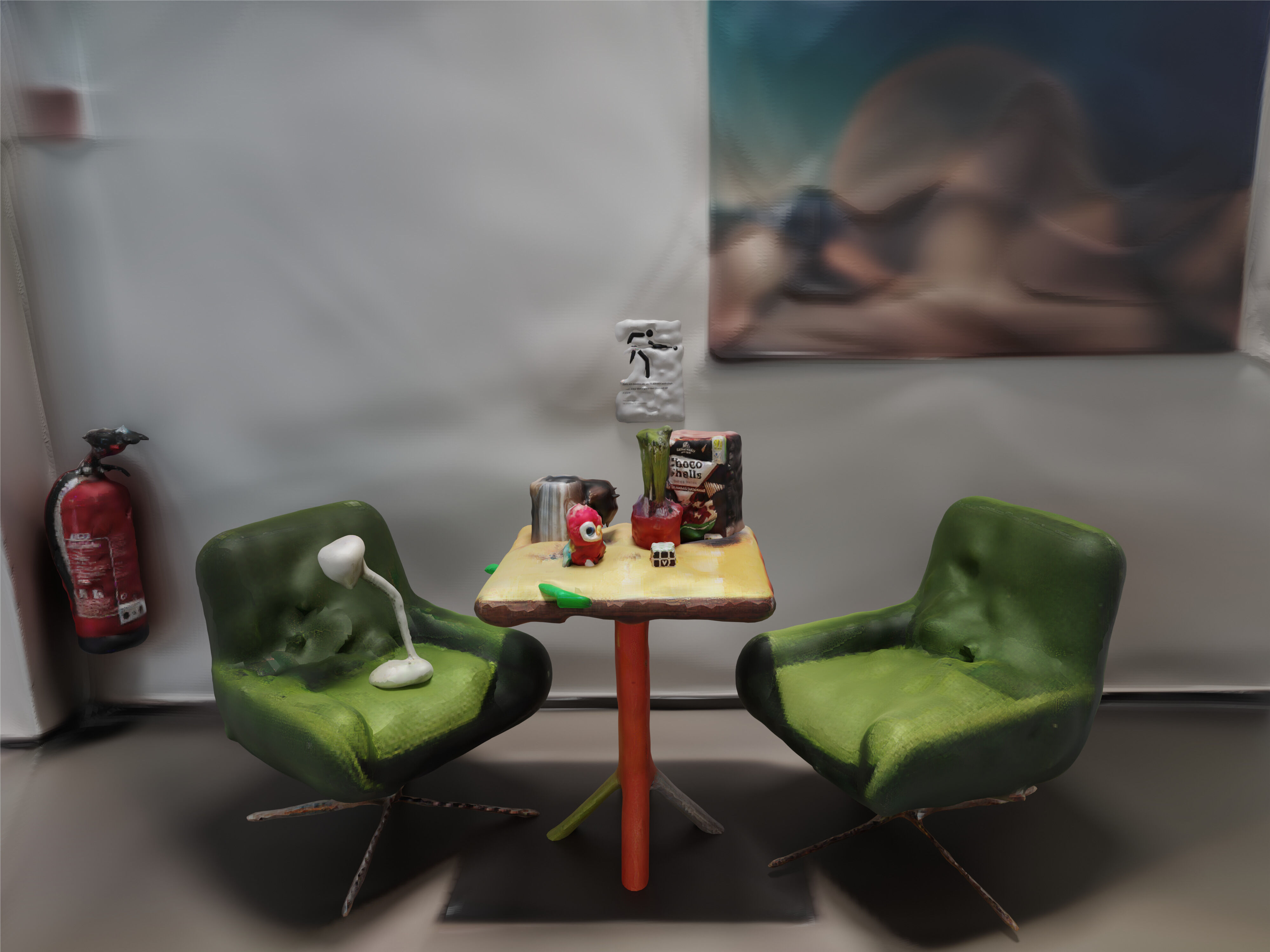} 
        \\ 
        \includegraphics[width=\wid]{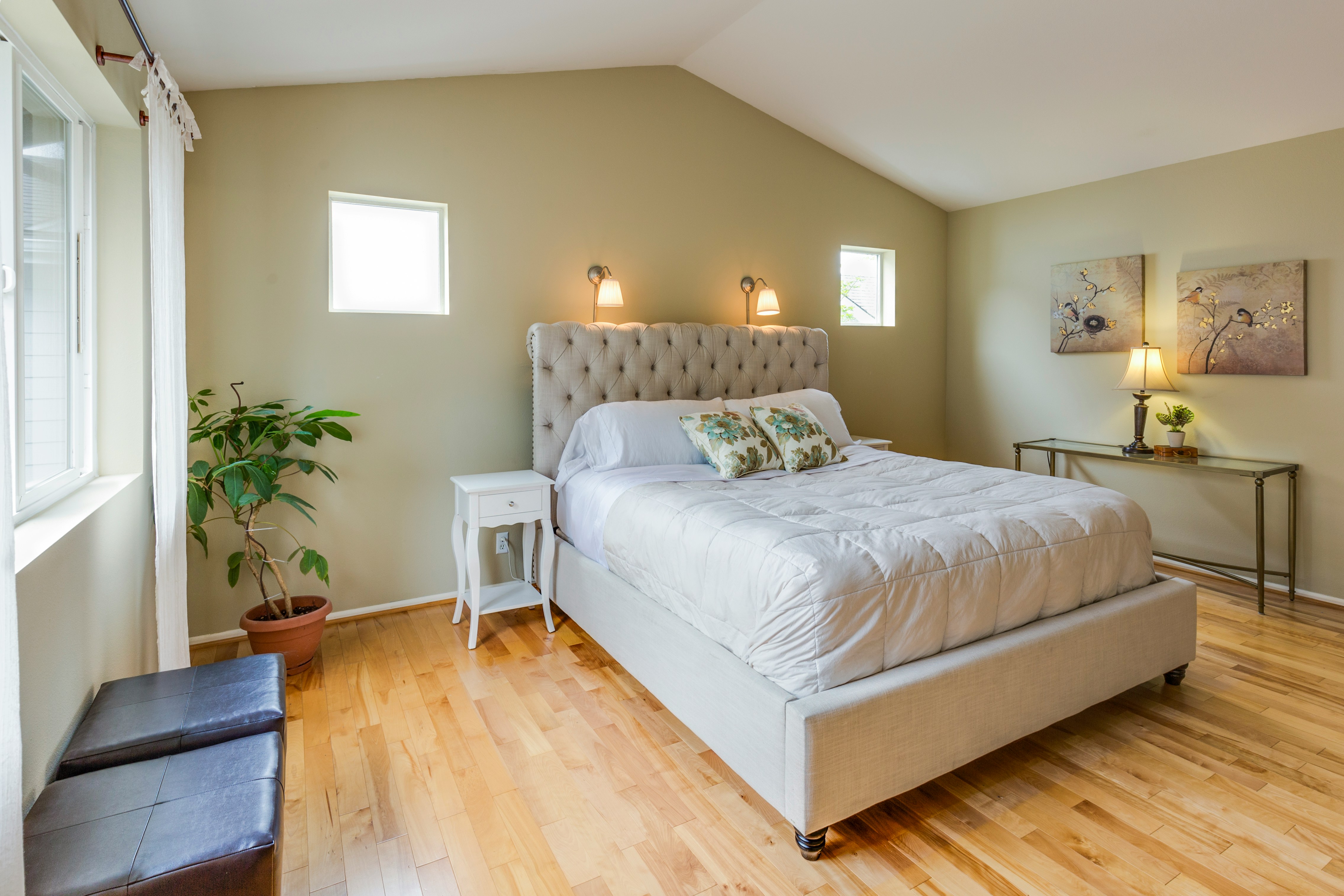} &
        \hspace{\mrg}
        \includegraphics[width=0.85\wid]{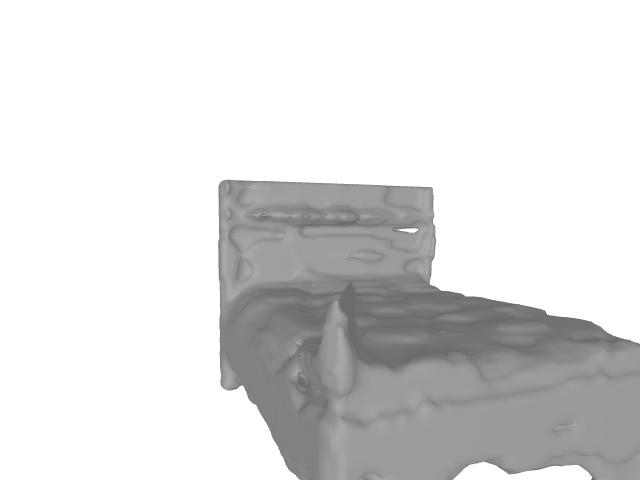}  &
        \hspace{\mrg}
        \includegraphics[width=\wid]{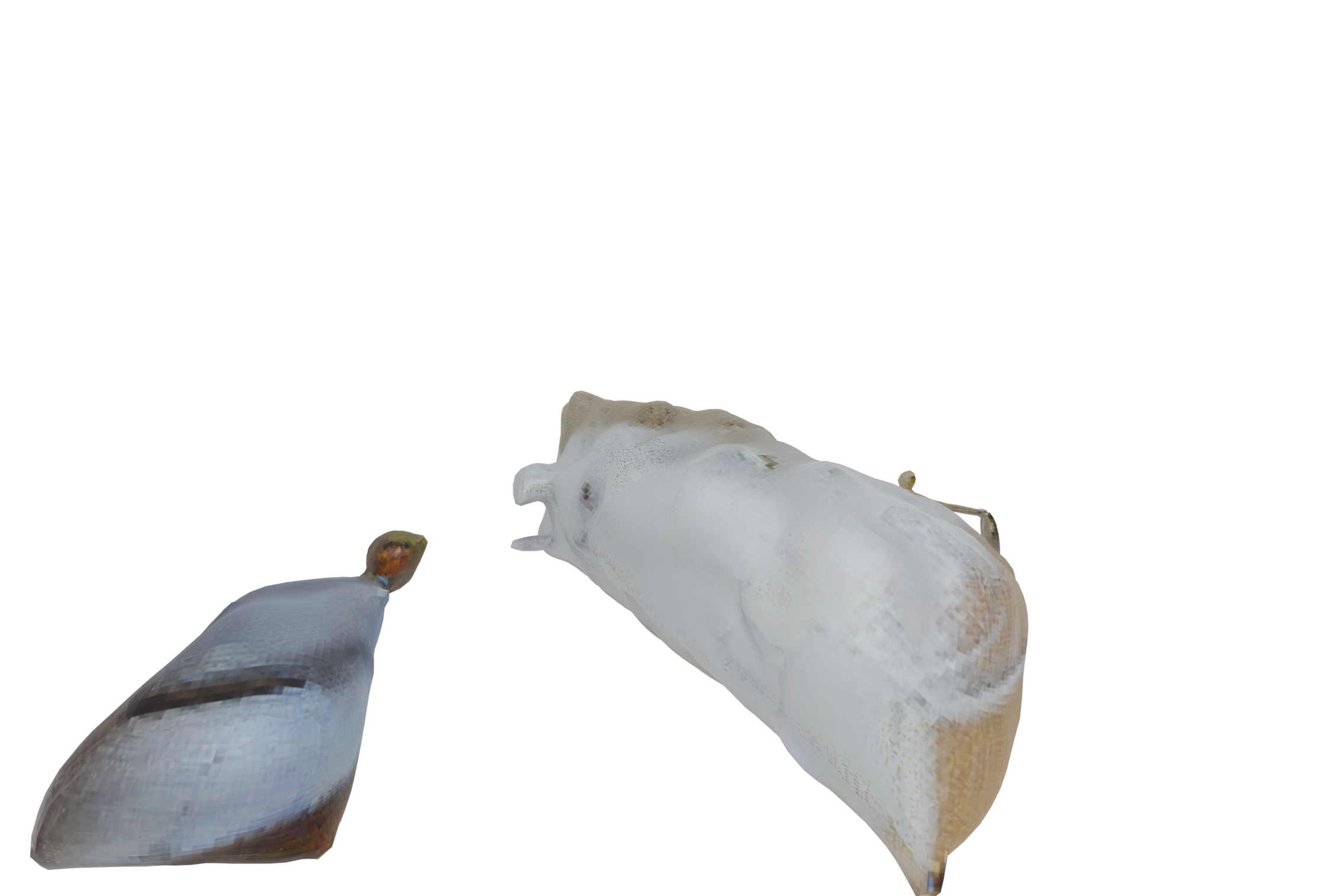} & 
        \hspace{\mrg}
        \includegraphics[width=\wid]{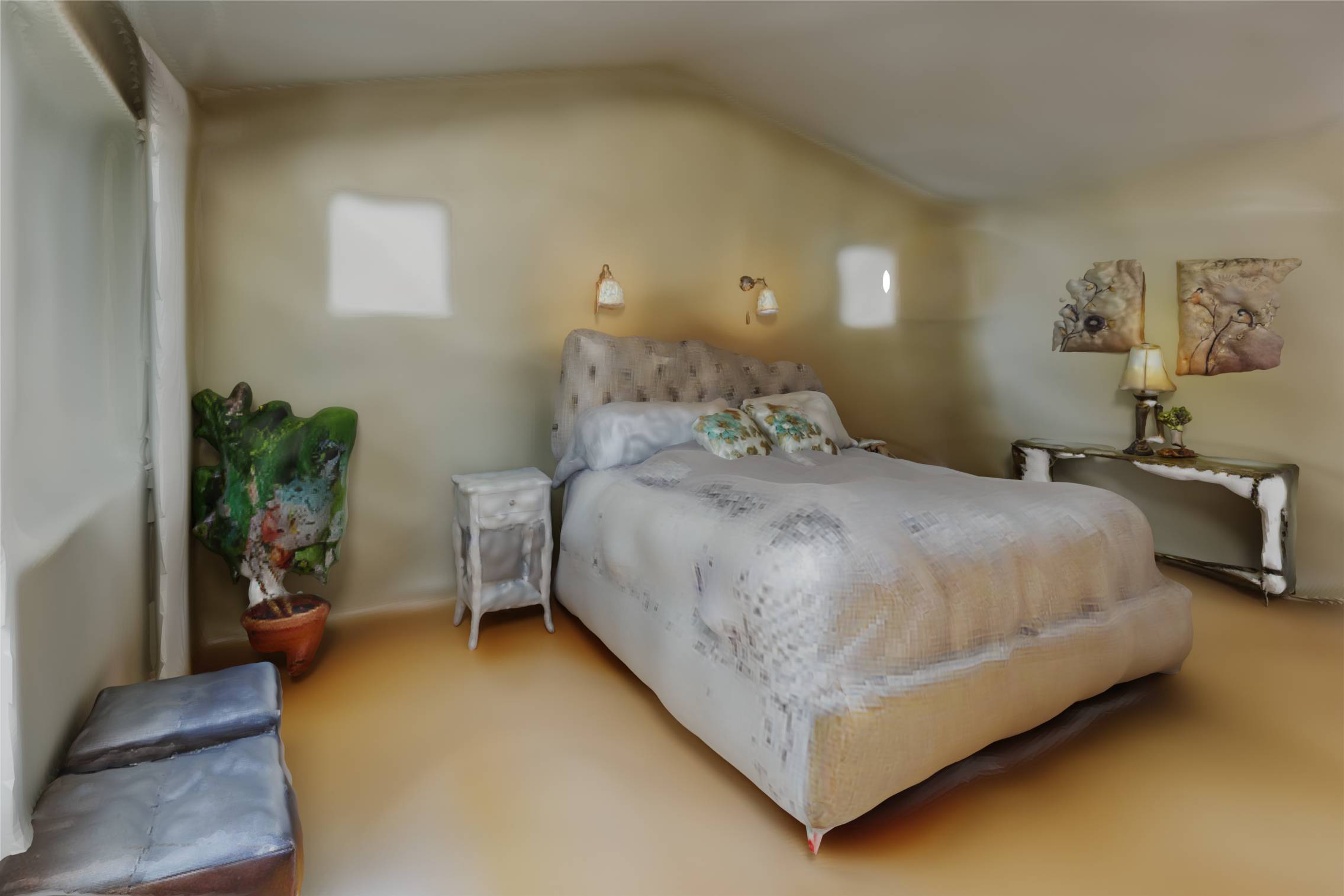}  
        \\ 
        \begin{tabular}{c}
        \includegraphics[width=0.49\wid]{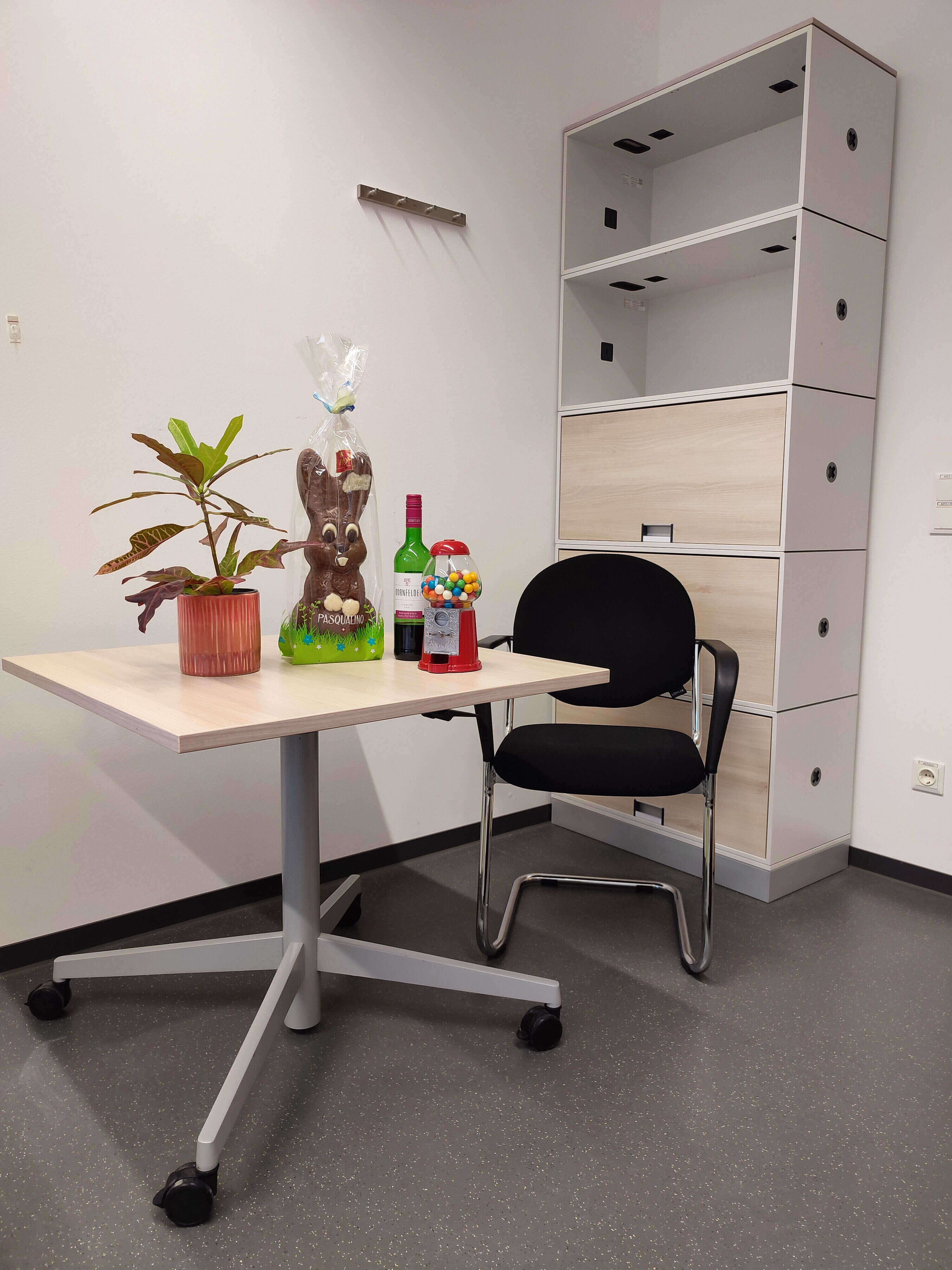} 
        \end{tabular}  &
        \hspace{\mrg}
        \begin{tabular}{c}
        \includegraphics[width=0.49\wid]{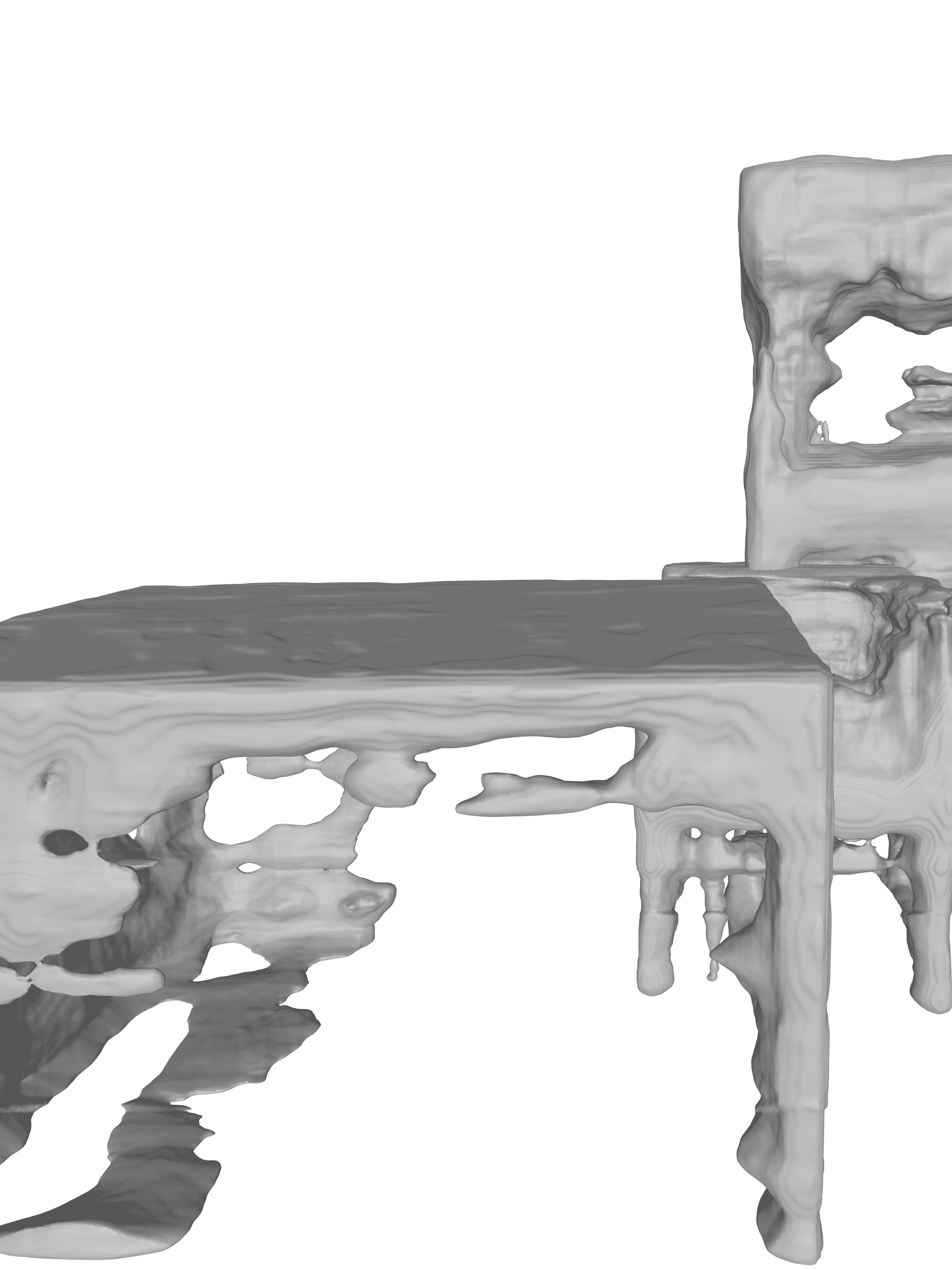} 
        \end{tabular}  &
        \hspace{\mrg}
        \begin{tabular}{c}
        \includegraphics[width=0.49\wid]{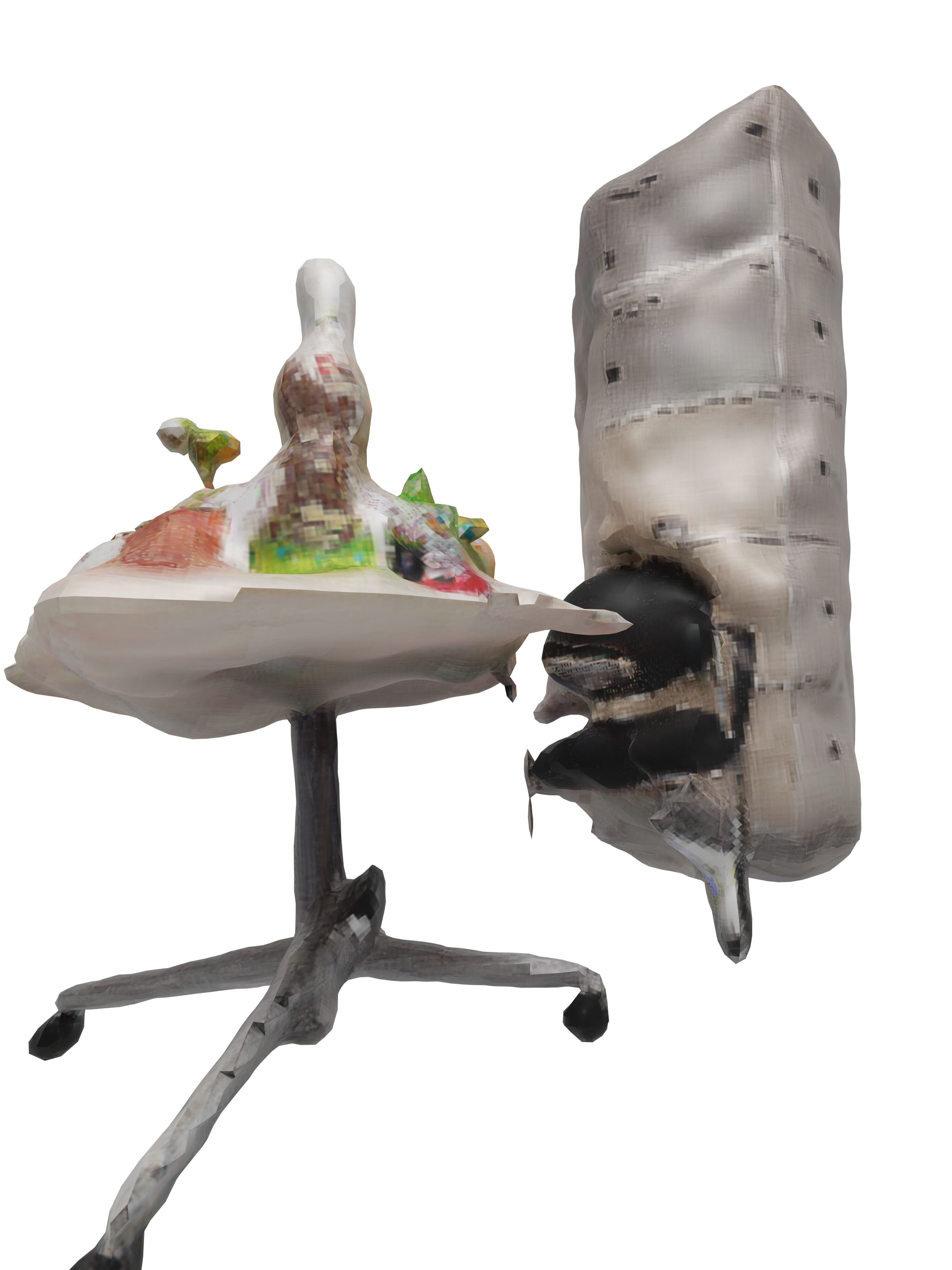} 
        \end{tabular} & 
        \hspace{\mrg}
        \begin{tabular}{c}
        \includegraphics[width=0.49\wid]{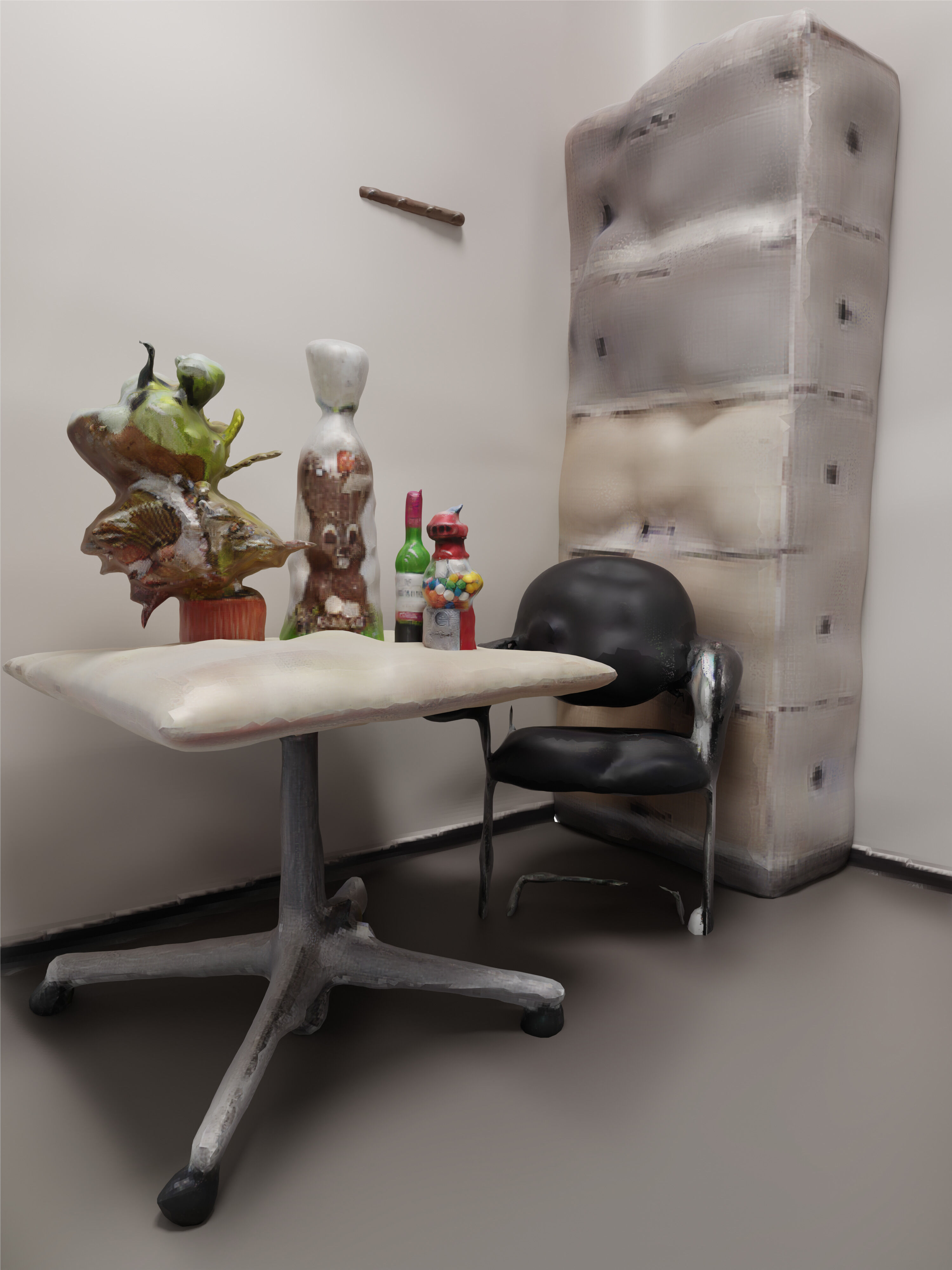} 
        \end{tabular}
        \\ 
        \vspace{\mrgv}
        Input image & \hspace{\mrg}    InstPIFu~\cite{liu2022towards} & \hspace{\mrg}    DreamGaussian~\cite{tang2023dreamgaussian} & \hspace{\mrg}
        Ours  
    \end{tabular}
    \vspace{-0.2cm}
    \caption{Qualitative results on real-world data. Interactive visualizations are included on our project page.
    } 
    \label{fig:real_qual}
    \vspace{-0.4cm}
\end{figure*}

InstPIFu \cite{liu2022towards} and USL \cite{gkioxari2022learning} are both compositional approaches that rely on 2D and 3D object detectors for identifying the objects to be reconstructed and aligning them in the scene. However, InstPIFu requires direct 3D supervision and USL is trained end-to-end. This limits their application domain and generalizability. 
Furthermore, as seen in Tables~\ref{tab:front_quant} and \ref{tab:hope_quant}, our method is able to quantitatively match the performance of InstPIFu even when evaluated on the 3D-FRONT dataset (which InstPIFu used for training).
The qualitative results in Figures \ref{fig:front_qual_bg} and \ref{fig:front_qual_fg} show that the method successfully reconstructs large furniture pieces and arranges them in a good layout; however, several objects such as plants and chandeliers are missing.
The results of InstPIFu further degrade when evaluated out-of-distribution, as can be seen 
on real-world images in Figure \ref{fig:real_qual}.
Since the implementation of USL~\cite{gkioxari2022learning} is not publicly available, we only compare our visual results on the Hypersim dataset \cite{roberts_2021_hyper} in Figure~\ref{fig:hyper_qual}. USL also misses many objects, and the reconstructed components have a simplified geometry with no texture. In contrast, our results are more realistic and have higher visual quality.

Additionally, 
we compare our method with DreamGaussian~\cite{tang2023dreamgaussian} by itself in a non-compositional approach. To apply the model, we treat all instances in a scene as a single object and reconstruct them together. As the model has seen several scenes composed of more than one object during its training \cite{deitke2023objaversexl}, it performs reasonably under this setting. Given the different camera intrinsics in the evaluation, we also compare the results of applying the model on the images after using a reprojection similar to the one used in our pipeline. This further boosts its performance as measured in Table~\ref{tab:hope_quant}. Still, its results are worse compared to our compositional approach, which can be analyzed in the qualitative results in Figure \ref{fig:front_qual_fg} and in the supplementary material. 

Lastly, we evaluate in Figure~\ref{fig:real_qual} the methods' performance on real-world data with diverse scenarios. The results show that the proposed solution reconstructs complex scenes well, while overcoming many limitations of prior works. We briefly address the reconstruction of outdoor scenes in the Section \ref{sec:outdoor} of the supplementary material.

\subsection{Ablations}

The two most important interfaces in combining the depth estimation and the single-view object reconstruction components are our reprojection and amodal completion.
We present an ablation of these components in Table~\ref{tab:front_ablation}.
Ablating the reprojection amounts to simply using image crops. This ignores projective geometry properties and leads to deformed reconstructions.
Amodal completion is necessary to contend with the
various occlusions that appear in the input view. This step boosts the performance on the 3D-FRONT dataset, but does not improve the numerical evaluation on HOPE-Image dataset, since most of the objects in the scenes are not occluded. 
More ablation results are included in the Section \ref{sec:qual_ablations} of the supplementary material. 
\begin{table}
    \centering
    \setlength{\tabcolsep}{0.15em}
    \begin{tabular}{l cc | cc}
        \\
        & \multicolumn{2}{c}{3D-FRONT \cite{front20213d}} & \multicolumn{2}{c}{HOPE-Image \cite{tyree2022hope}} \\
        Method & Chamfer $\downarrow$ & F-Score $\uparrow$ & Chamfer $\downarrow$ & F-Score $\uparrow$
        \\
        \hline
        Ours (full) & \textbf{0.120} & \textbf{68.82} & \textbf{1.446} & 54.82\\
        w/o reprojection 
         & 0.155 & 58.56 & 1.505 & 51.89\\
         w/o amodal comp.
         & 0.122 & 66.81 & 1.450 & \textbf{54.99} \\
        \hline
    \end{tabular}
    \vspace{-0.1cm}
    \caption{Ablation results for our reprojection and amodal completion modules; both contribute to the performance of our method.}
    \label{tab:front_ablation}
    \vspace{-0.6cm}
\end{table}

\subsection{Limitations}
The proposed method has certain shortcomings and there is significant room for improvement, especially for in-the-wild predictions. By design, the failure cases of the individual modules (depth estimation, camera calibration, elevation estimation, etc.) become limitations of our framework. Since we do not train an end-to-end system, errors can propagate from one stage to the next. Therefore, the performance of the overall pipeline is limited by its weakest link.

We believe that most of the current limitations can be overcome by improving the implementation of some of the particular modules in our framework and by enhancing their interoperability: using estimated depth in 3D object reconstruction or global image context for amodal completion. We further discuss the method's limitations and provide concrete examples in the Section \ref{sup:limit} of the supplementary material.

\section{Conclusion}
In this work, we introduce a modular framework for reconstructing complex 3D scenes from an image. We prioritize generalization by taking a divide-and-conquer approach rather than end-to-end. Our decomposing into multiple entities benefits from existing components, which effectively solve the established subtasks. We develop the necessary interfaces that enable the modules to function properly and finally yield a full 3D reconstruction.
Our experiments decidedly show the advantage of the proposed method on various types of scenes.
Considering the illustrated performance for diverse scenarios, we believe that our approach is a strong baseline and a stepping stone towards generalizable full 3D scene reconstruction from a single image.

\vspace{0.5\baselineskip}
{
\footnotesize{
\textbf{Acknowledgement.}
This work was funded by the German Federal Ministry of Education and Research (BMBF), FKZ: 01IS22082 (IRRW). The authors are responsible for the content of this publication.
The authors appreciate the scientific support and HPC resources provided by the Erlangen National High Performance Computing Center (NHR@FAU) of the Friedrich-Alexander-Universität Erlangen-Nürnberg (FAU) under the NHR project b112dc IRRW. NHR funding is provided by federal and Bavarian state authorities. NHR@FAU hardware is partially funded by the German Research Foundation (DFG) – 440719683.
}
}
{
    \small
    \bibliographystyle{ieeenat_fullname}
    \bibliography{main}
}
\clearpage
\setcounter{page}{1}
\maketitlesupplementary

\section*{Overview}
This supplementary material includes additional details and results which provide insights into our novel approach for generalizable 3D scene reconstruction.
In section \ref{sup:impl_details}, we present supporting information for the reproducibility of our method and the evaluation setup. Section \ref{sec:additional_exp} extends the ablation study in the main paper and shows the effect of replacing different modules in the pipeline (\ref{sec:alternative}), presents a qualitative comparison for the ablations (\ref{sec:qual_ablations}), and discusses the application on outdoor scenes (\ref{sec:outdoor}). Then, in section \ref{sup:limit}, we elaborate and illustrate the limitations of our method. Additional results and our code base can be found on the project page \footnote{\url{https://andreeadogaru.github.io/Gen3DSR}}. 

\section{Implementation details} 
\label{sup:impl_details}
\subsection{Amodal completion}
\label{sec:comp}
\begin{figure}
    \centering
    \includegraphics[width=0.49\textwidth]{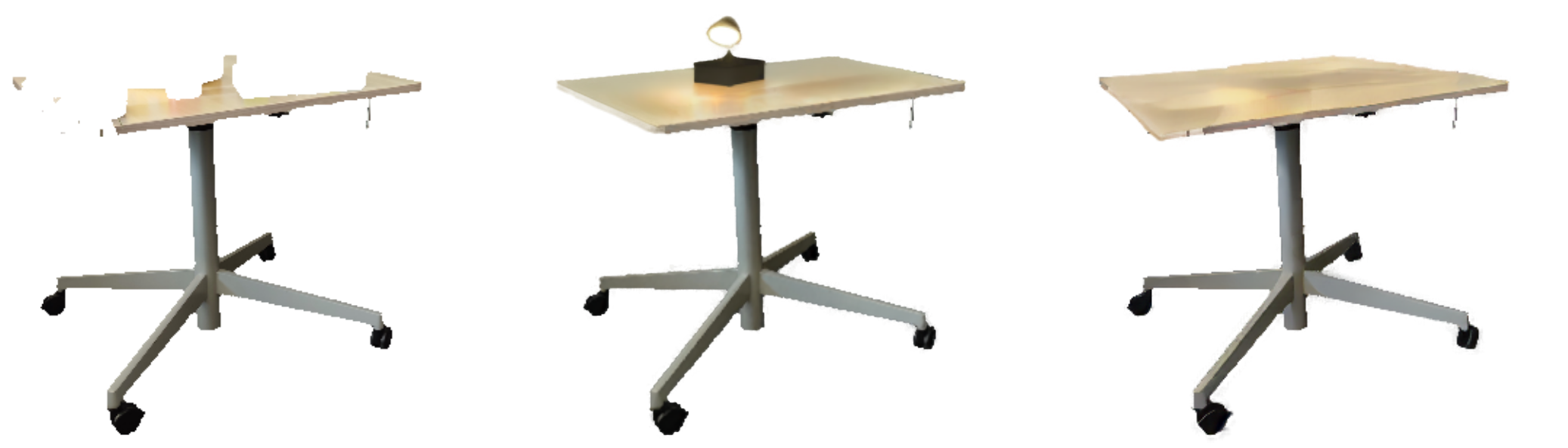}
    \vspace{-0.2cm}
    \begin{tabularx}{0.49\textwidth}{XXX}
        \centering Input crop & \centering SD XL inpainting & \centering Our amodal completion
    \end{tabularx}
    \caption{In comparison to our amodal completion, inpainting models tend to hallucinate new objects when filling in the holes. Both models use \texttt{desk} as prompt.}
    \label{fig:am_in}
    \vspace{-0.3cm}
\end{figure}

Amodal completion differs from inpainting by ensuring that the model does not hallucinate new objects when recovering the missing parts, as shown in Figure \ref{fig:am_in}. 

\begin{figure*}
    \centering
    \includegraphics[width=0.99\textwidth]{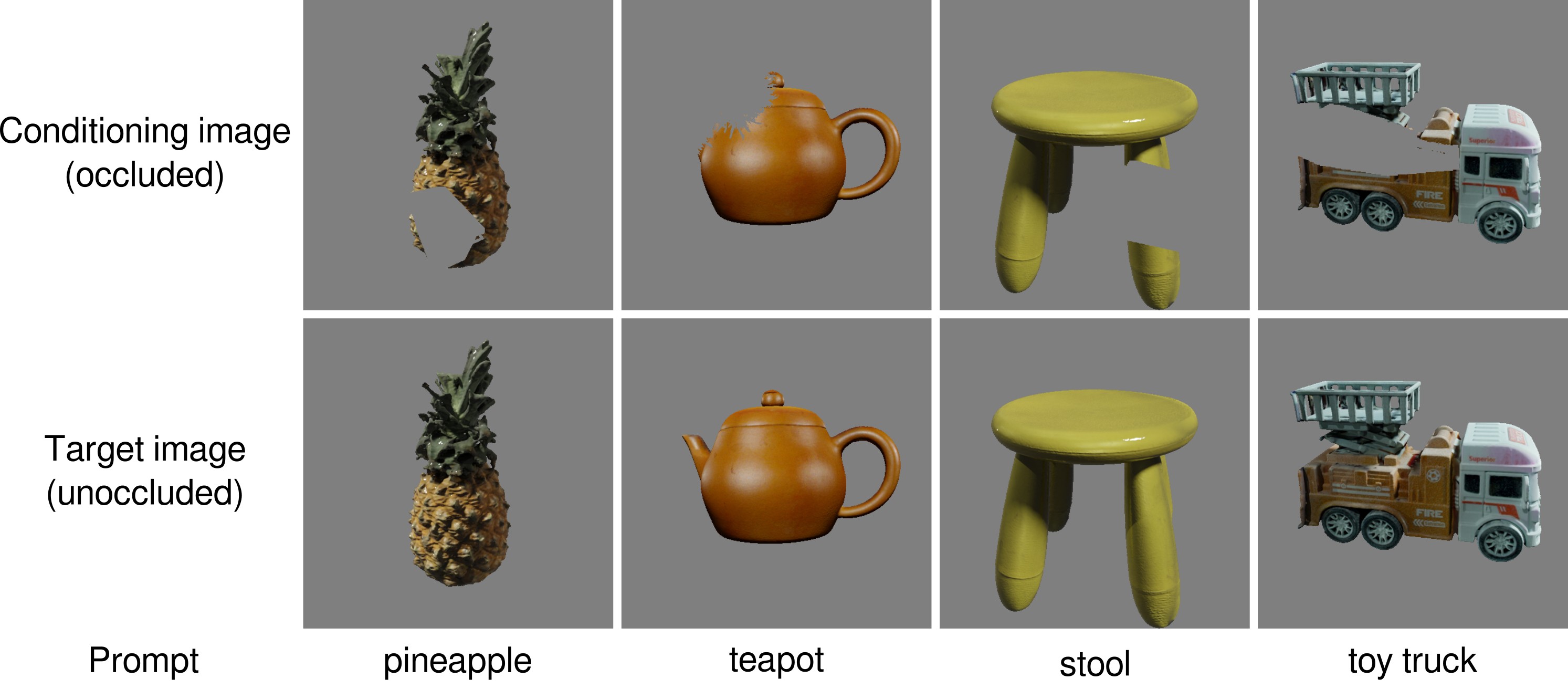}
    \caption{Training samples from the amodal completion synthetic dataset with objects from OmniObject3D \cite{wu2023omniobject3d}.}
    \label{fig:amodal_data}
    \vspace{-0.3cm}
\end{figure*}

The amodal completion step in our framework is fulfilled by an image-to-image Stable Diffusion \cite{Rombach2021HighResolutionIS} model fine-tuned on a custom dataset. We create this dataset starting from OmniObject3D \cite{wu2023omniobject3d} which contains $6000$ high-quality real-scanned 3D objects. 
Though the model only needs 2D images for training, we found it beneficial to render 3D objects instead of using real images. This allows us to obtain well-segmented singular objects. 
The Blender-rendered images of the objects are used as the target images. 
To obtain the conditioning view, we mask-out parts of the object by randomly overlaying the silhouette of an arbitrary object sampled from the Objaverse dataset \cite{deitke2023objaverse}. 
All the channels of the background and the occluded pixels are set to the same value of $127.5$ (equivalent to zero after image  normalization).  
The category labels provided in OmniObject3D are used as prompts to guide the diffusion process. 
Training samples from the dataset are provided in Figure \ref{fig:amodal_data}.
We train the model at a resolution of $512 \times 512$ for $25000$ iterations with a batch size of $16$. 

As our amodal completion model does not explicitly infer the amodal mask of the completed object, we obtain it by using a foreground segmentation model \cite{qin2022}. 

\subsection{Reprojection and view-space alignment}
Utilizing the estimated depth map $D$ alongside the camera calibration $K_{img}$, we perform pixel unprojection from the input image, resulting a point cloud $P^{view}$ within the view space. 
We refer to $P^{view}$ as our layout guide; this serves as a pivotal reference, ensuring the alignment of all individually reconstructed components within the view space, thus forming the complete scene. While the background modeling step directly fits a SDF to the corresponding background points $P_{bg}^{view}$, the instance processing step uses $P_i^{view}$ for two purposes: reprojection and alignment. 

The straightforward approach of simply cropping the instances out of the original image and feeding them to the single-shot object reconstruction method $\mathcal{R}$ leads to deformed reconstructions. This happens because the view-conditioned diffusion model $\mathcal{Z}$, which is used for generating novel views of the object, assumes images are captured under a predefined setup \cite{liu2023zero}. Specifically, the object should be in the center of the image, captured by a camera with field of view of $49.1^{\circ}$, positioned at a distance between $[1.0, 1.7]$ from the object which is normalized to fit into the unit cube. As these conditions are not satisfied by arbitrary crops within the image, we project $P_i^{view}$ into a crop $C_i$ with similar properties. The virtual camera employed for projection uses the $\mathcal{Z}$-compatible camera setup; \ie, a camera positioned at a distance of $1.5$ from the normalized $P_i^{view}$ and oriented towards its center, with intrinsics $K_{crop}$. The reconstructed object might have a different scale compared to $P_i^{view}$. Therefore we estimate the scale factor $s_i$ that aligns the reconstructed object with $P_i^{view}$ using a RANSAC-based approach \cite{fischler1981random}, accounting for the possible mismatches.

\subsection{Background modeling}
The MLP used to reconstruct the background has 4 hidden layers with
128 neurons each and Softplus activations. The network is
trained from scratch for each scene using points sampled
along background camera rays. The SDF of the points is computed as the distance to the unprojected estimated depth.
As we do not sample points in the regions originally occluded, the network implicitly interpolates the missing areas based on the visible surrounding regions. 

\subsection{Integrated models}
\label{sec:integrated}
Our method ensembles several models, each tackling a different sub-tasks as follows:
\begin{itemize}
    \item Camera calibration (estimation of field of view and principal point): Perspective Fields~\cite{jin2023perspective} trained on 360cities~\cite{360cities} and EDINA~\cite{Do_2022_EgoSceneMSR} datasets. 
    \item Entity segmentation: CropFormer~\cite{Qi_2023_ICCV} with Hornet-L backbone trained on EntitySeg dataset~\cite{Qi_2023_ICCV}.
    \item stuff-thing segmentation: OneFormer~\cite{jain2023oneformer} with DiNAT-L backbone trained on ADE20K~\cite{zhou2019ade20k}. The model predicts 150 classes which are grouped in stuff and thing. We make some modifications to the original grouping; \textit{thing} $\to$ \textit{stuff}: window, door, curtain, mirror, fence, rail, column, stairs, screen door, balustrade, step; \textit{stuff} $\to$ \textit{thing}: plant, tent, crt screen, cradle, blanket. 
    \item Monocular depth estimation: Depth Anything~\cite{depthanything} fine-tuned on NYUv2~\cite{Silberman_ECCV12} (metric depth) and Marigold~\cite{ke2023repurposing} with Stable Diffusion v2 backbone trained on two synthetic datasets (affine-invariant depth).
    \item Object recognition: OVSAM \cite{yuan2024ovsam} which combines SAM~\cite{Kirillov_2023_segmenta} with CLIP\cite{jia2021clip} and is trained on COCO~\cite{lin2014microsoft} and LVIS~\cite{gupta2019lvis} datasets. For each object we sample 5 points inside the eroded instance mask to prompt the model. 
    \item Amodal completion: our model based on Stable Diffusion v1.5 and trained on the synthethic dataset described in section \ref{sec:comp} of this supplementary material.
    \item Singe-image object reconstruction: DreamGaussian~\cite{tang2023dreamgaussian} which reconstructs the object using 3D Gaussians~\cite{kerbl3Dgaussians} and employs Zero-1-to-3 XL \cite{liu2023zero} trained on ObjaverseXL~\cite{deitke2023objaversexl} as the 2D diffusion prior.
\end{itemize}

We show variants of our pipeline with different modules for some of the processing steps (depth estimation, stuff-thing segmentation, and object reconstruction) in section \ref{sec:alternative}.

\subsection{Inference time}
Using the default pipeline configuration on an NVIDIA RTX
A5000, for an image with resolution $1500 \times 1500$, the scene
analysis stage takes $\approx 80$ sec and for each instance specific
processing additional $\approx 90$ sec are required. Similar to other compositional methods (\eg, InstPIFu \cite{liu2022towards}), the inference time increases linearly with the complexity of the scene (number of objects). This limitation can be mitigated by paralellizing the reconstruction of the objects, provided the available hardware resources allow it. Still, the time per instance is quite high, mostly due to the inference time of DreamGaussian \cite{tang2023dreamgaussian}. This particular method optimizes 3D Gaussians to views of the object and has an additional step for texture refinement, adding to the overall runtime. However, as we designed a modular pipeline, we can replace DreamGaussian \cite{tang2023dreamgaussian} with new, improved methods for single-view 3D object reconstruction, such as LaRa \cite{LaRa}. This very recent method enables us to reduce the processing time per instance to $\approx 50$ sec. 
Finally, the processing time could be further improved by streamlining the pipeline for a specific configuration of modules. 

\subsection{Evaluation details}

For 3D-FRONT~\cite{front20213d, future20213d} we report the metrics at the original scale of the geometry in the dataset using 1000000 points sampled from both the reconstructed meshes and the ground truth geometry. The F-Score is computed at a threshold of $0.1$. As most of the modules in our pipeline operate at a high-resolution, the original resolution of the images ($648 \times 484$) is insufficient. Therefore, we first increase their size to $1296 \times 968$ using a super-resolution method \cite{wu2023seesr}. 

For HOPE-Image~\cite{tyree2022hope}, we compose the ground truth geometry using the provided pose annotations. Additionally, we found that a scale of $0.1$ must first be applied to the original object meshes to match with the images. The metrics are reported at this scale based on 500000 sampled points and the threshold used for F-Score is $1.0$. On this dataset, we only evaluate the reconstruction of the foreground instances, as there is no ground truth background geometry.

\section{Additional experiments}
\label{sec:additional_exp}

\subsection{HOPE-Image}
\begin{figure*}
    \centering    
    \setlength{\wid}{0.24\textwidth}
    \setlength{\mrg}{-0.35cm}
    \setlength{\mrgv}{0cm}
    \begin{tabular}{cccc}
        \vcentered{\includegraphics[width=\wid]{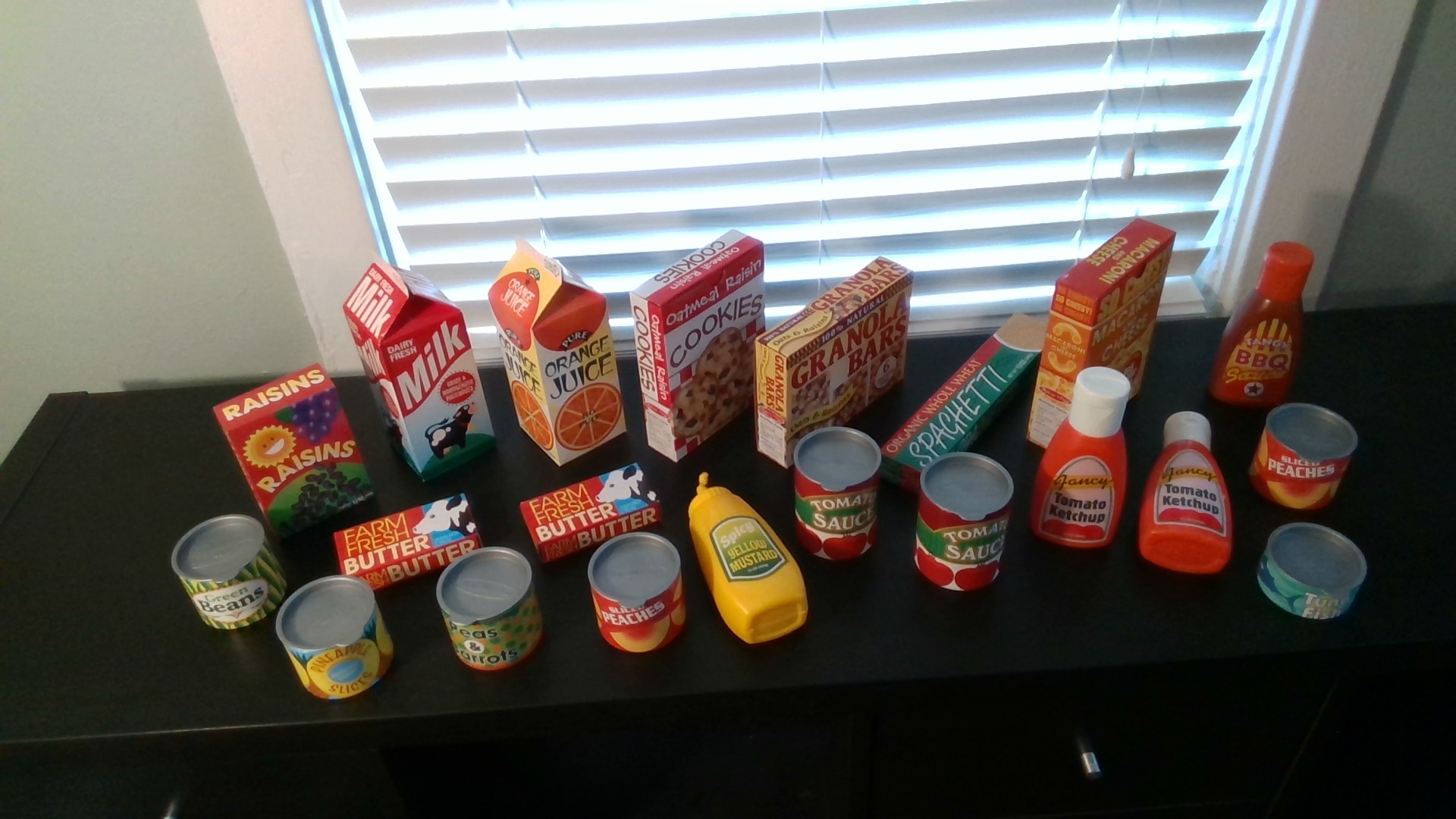}} &
        \hspace{\mrg}
        \begin{tabular}{c}
        \includegraphics[width=\wid]{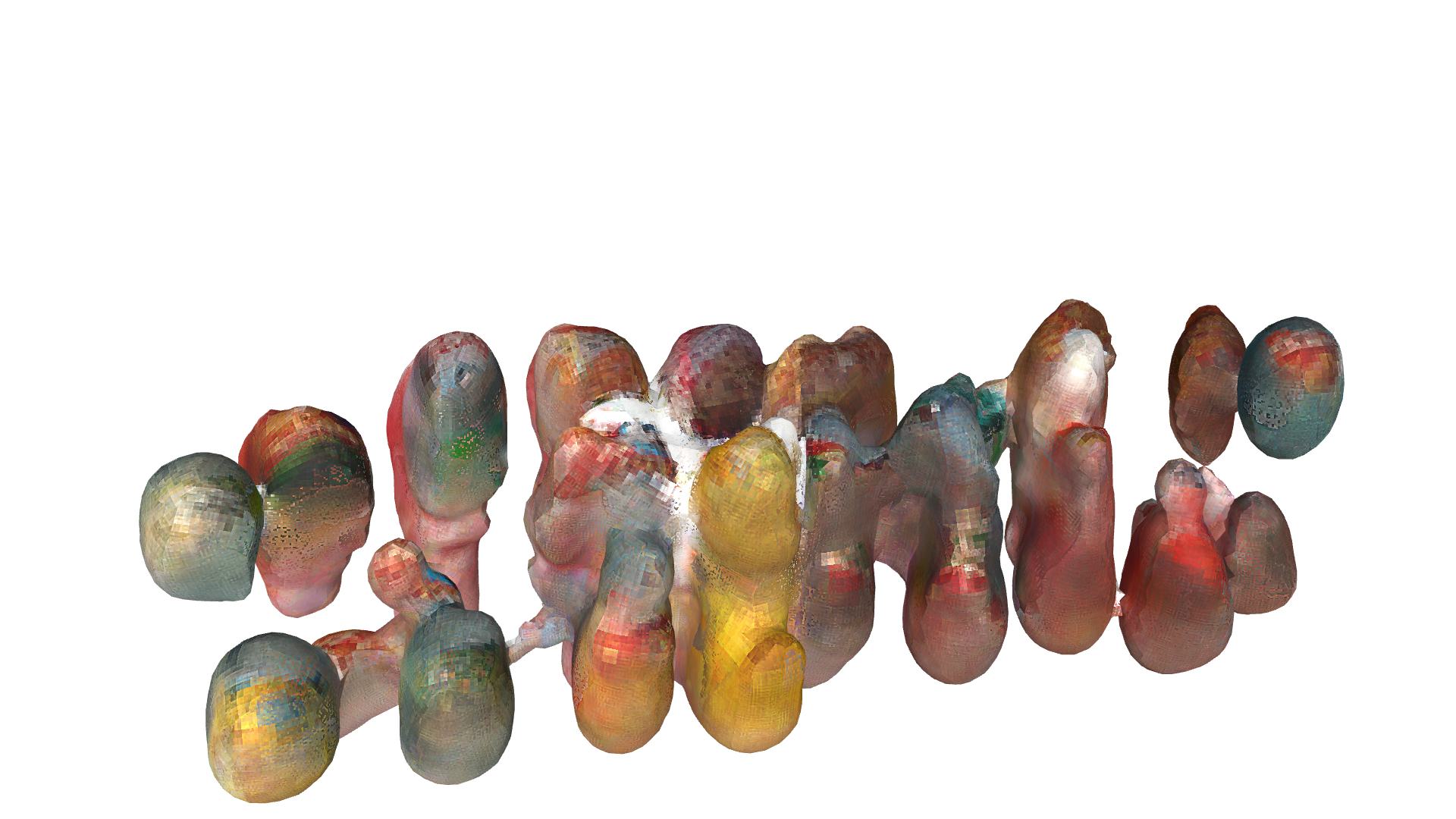}  \\
        \includegraphics[width=\wid]{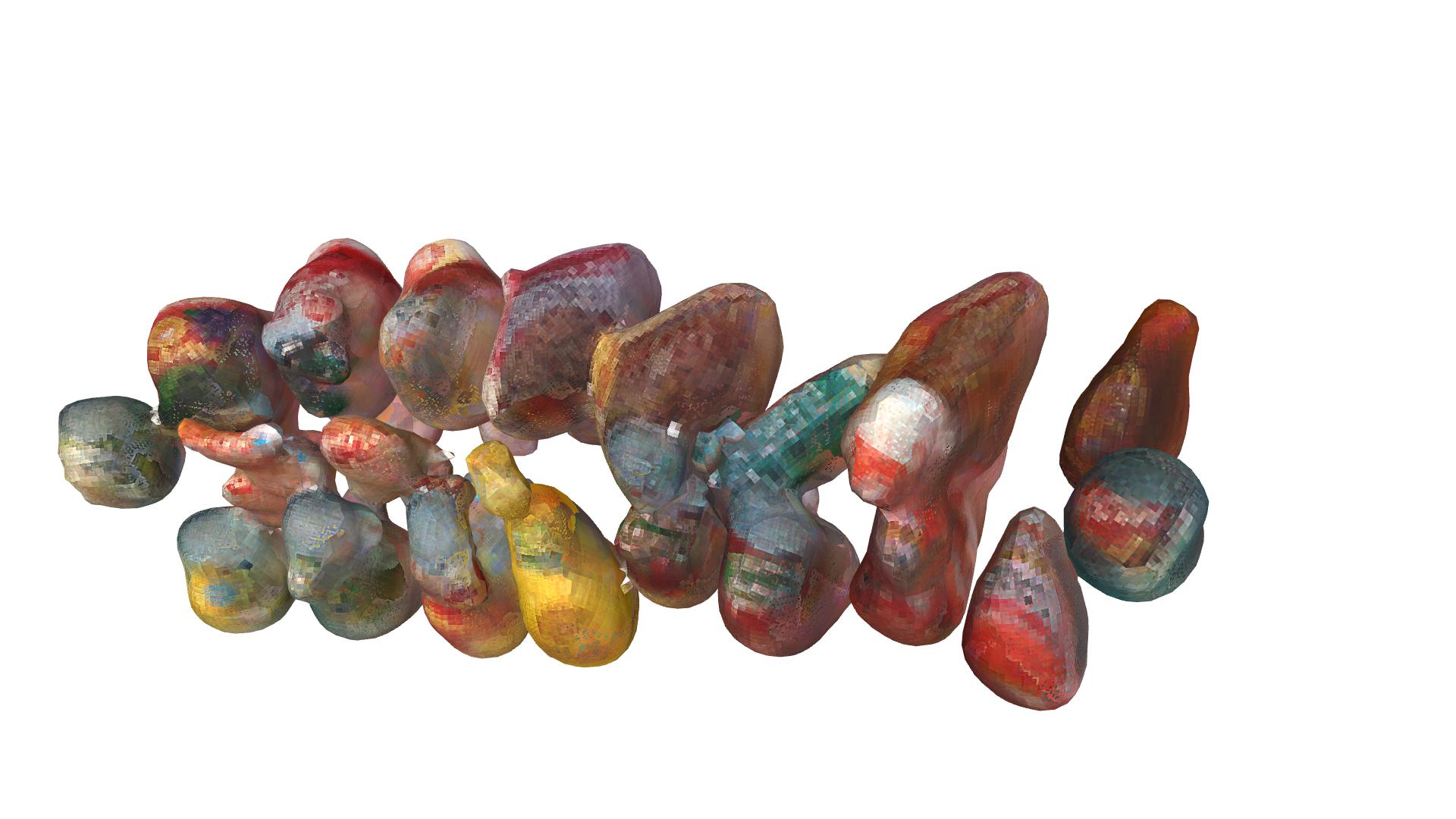}
        \end{tabular}
         & 
        \hspace{\mrg}
        \begin{tabular}{c}
        \includegraphics[width=\wid]{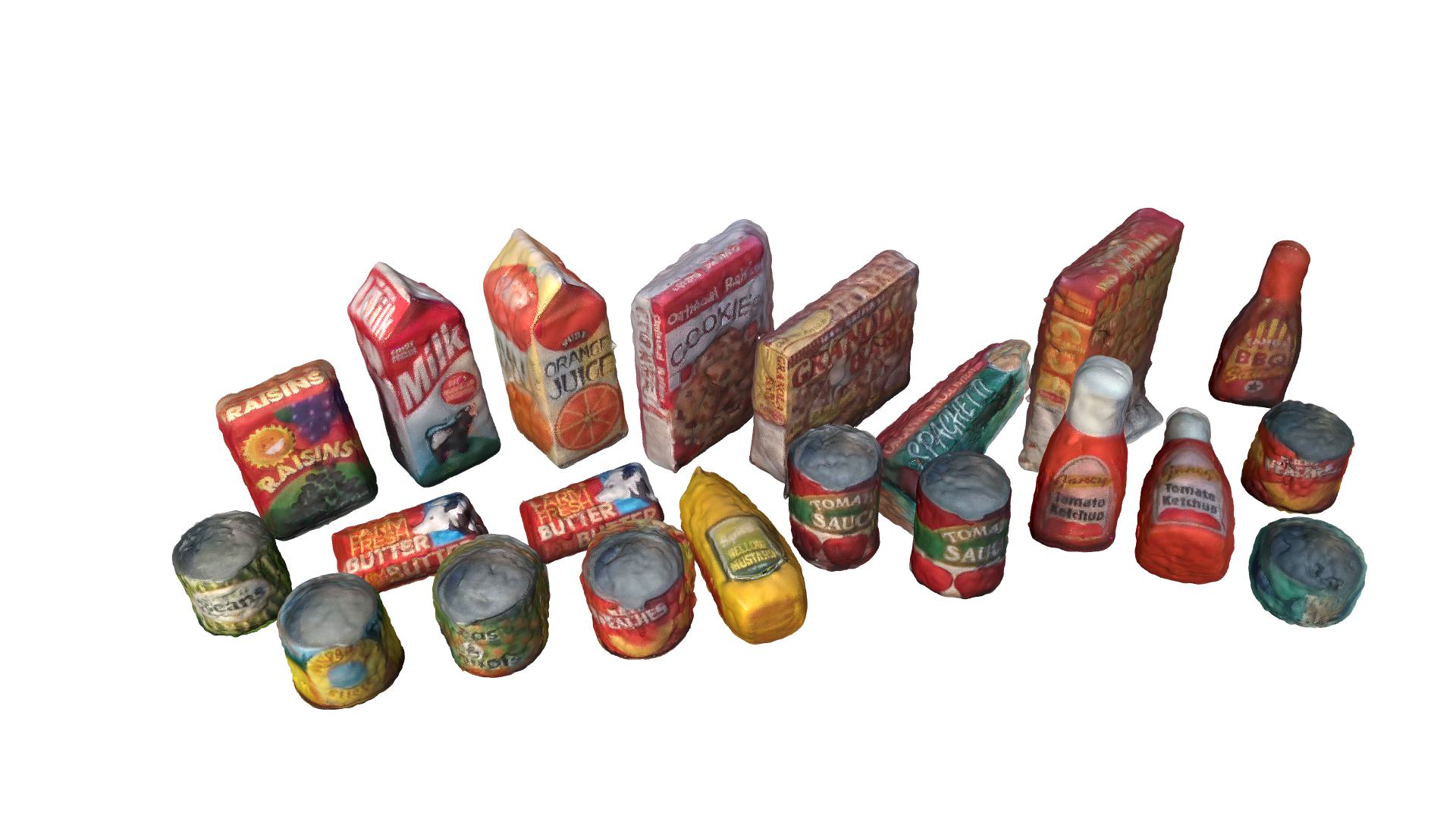} \\
        \includegraphics[width=\wid]{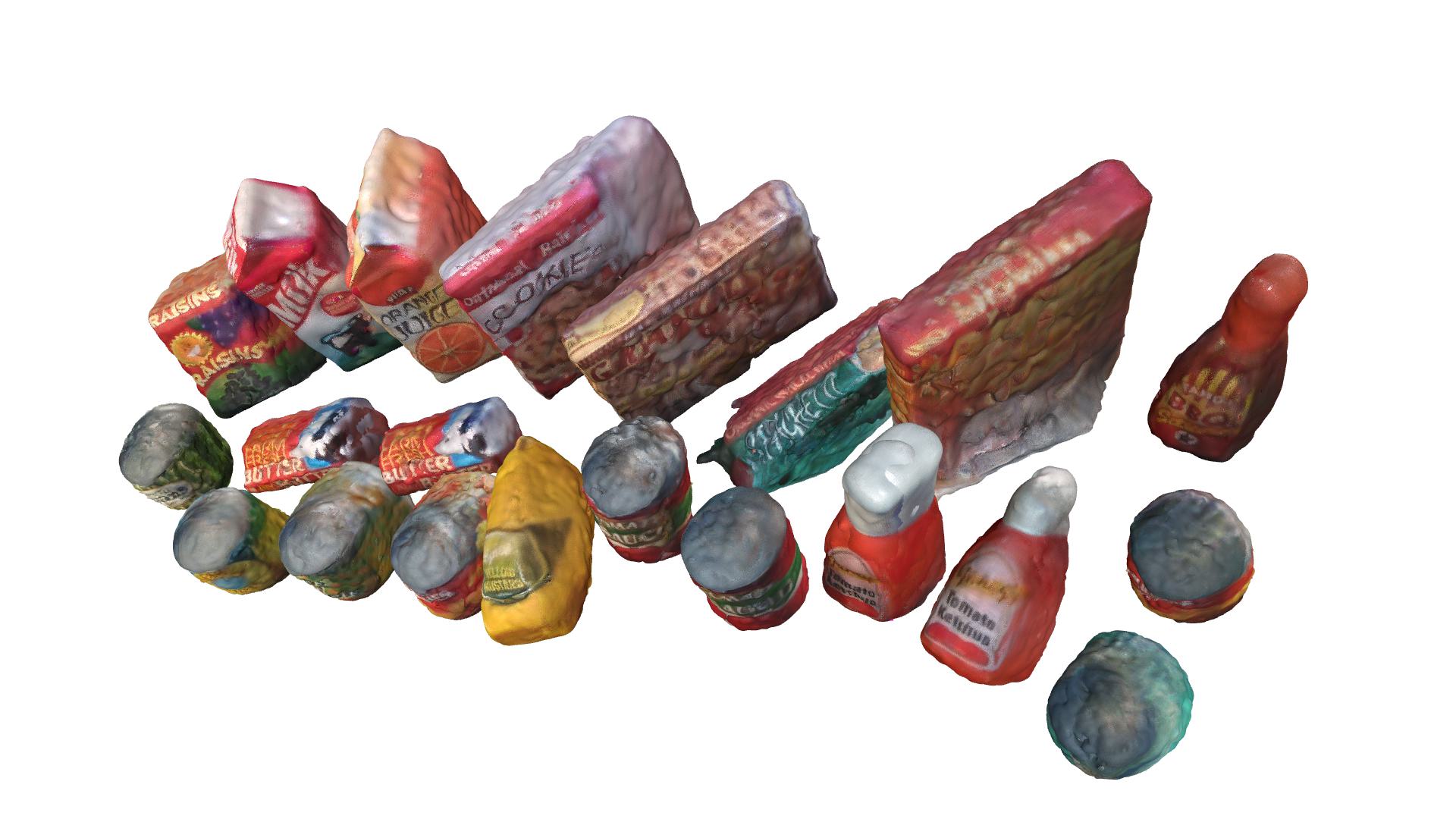} 
        \end{tabular}
        & 
        \hspace{\mrg}
        \begin{tabular}{c}
        \includegraphics[width=\wid]{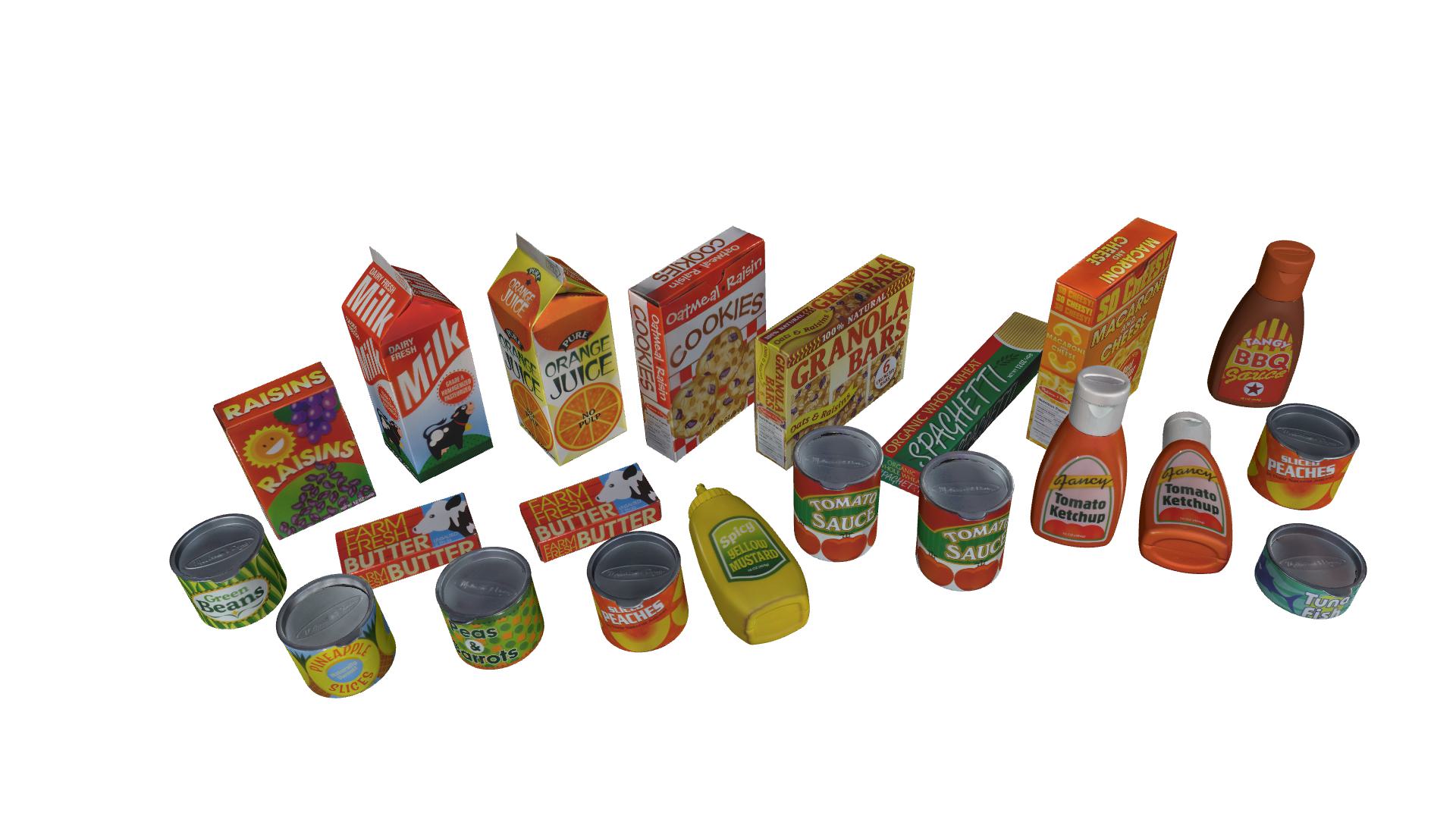} 
        \\
        \includegraphics[width=\wid]{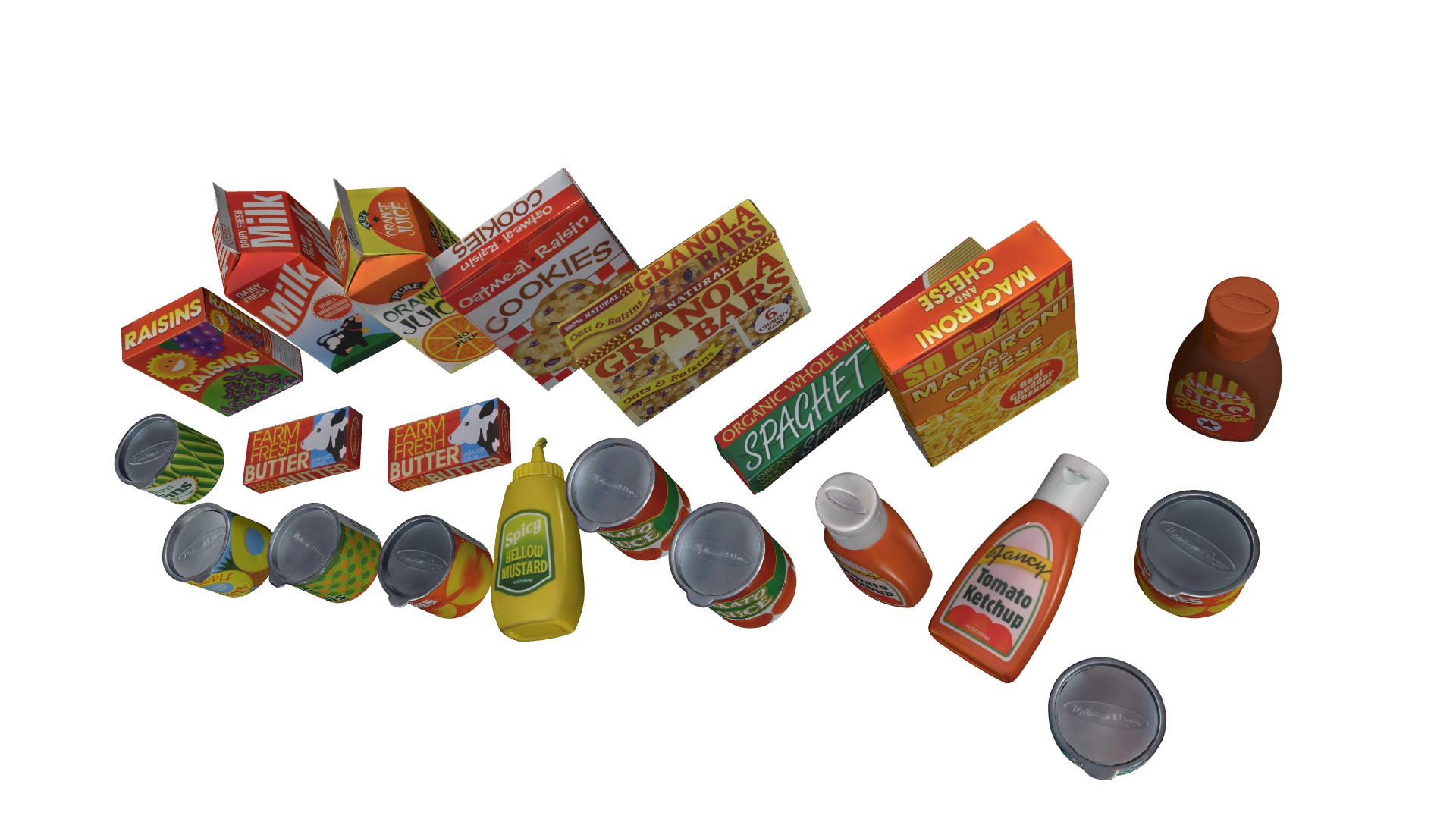}
        \end{tabular}
         \\
        \vcentered{\includegraphics[width=\wid]{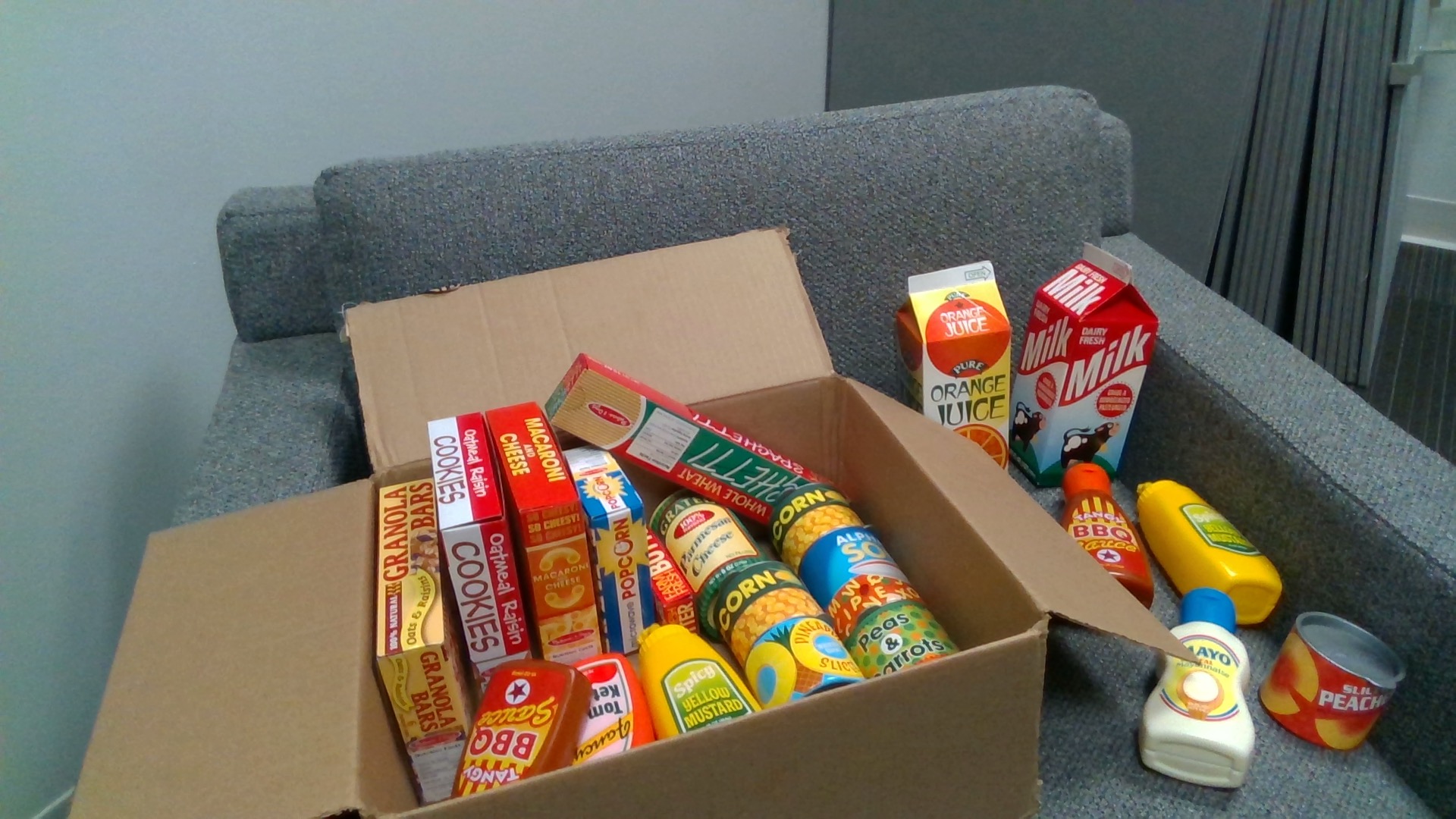}} &
        \hspace{\mrg}
        \begin{tabular}{c}
            \includegraphics[width=\wid]{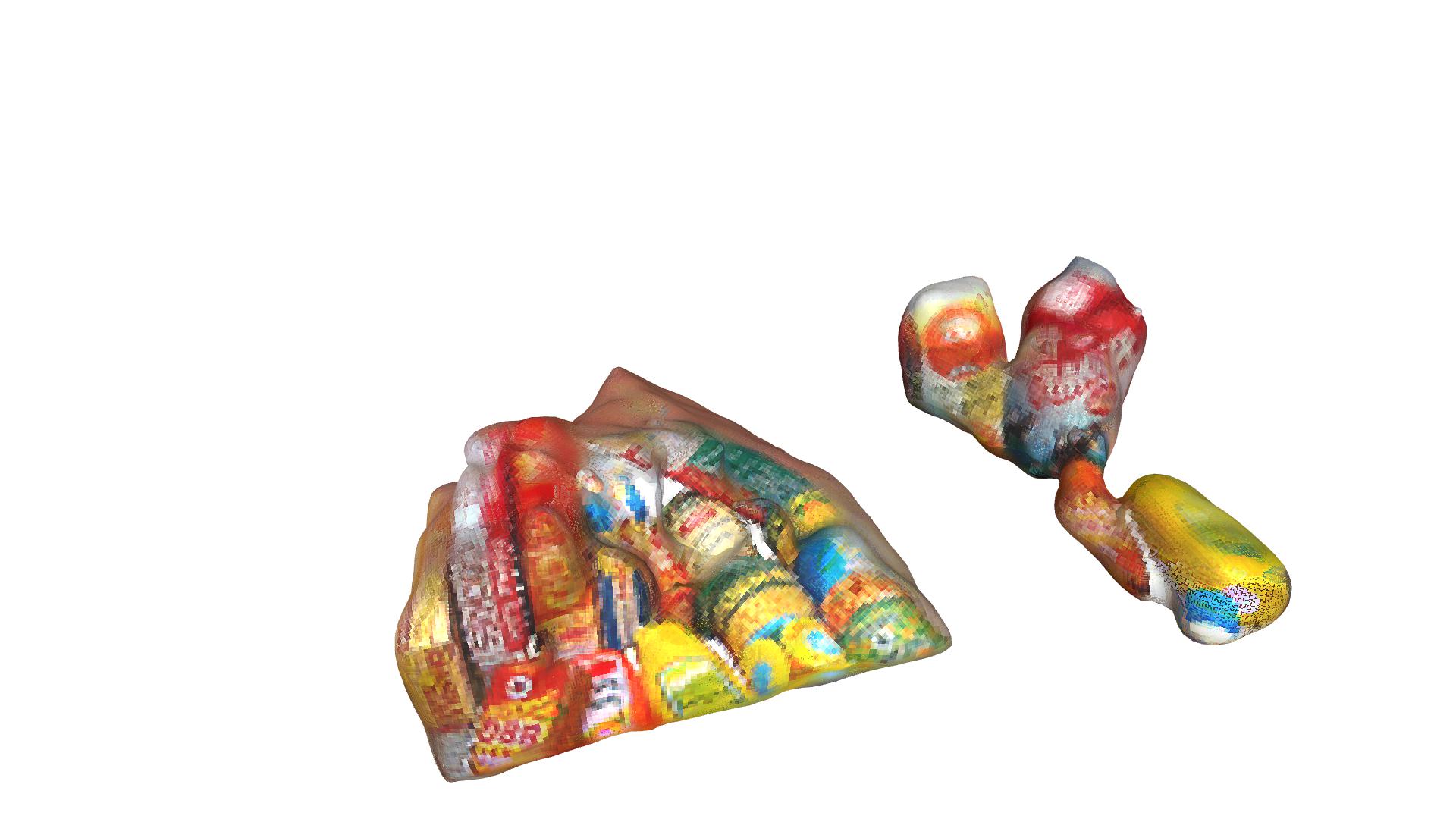} \\
            \includegraphics[width=\wid]{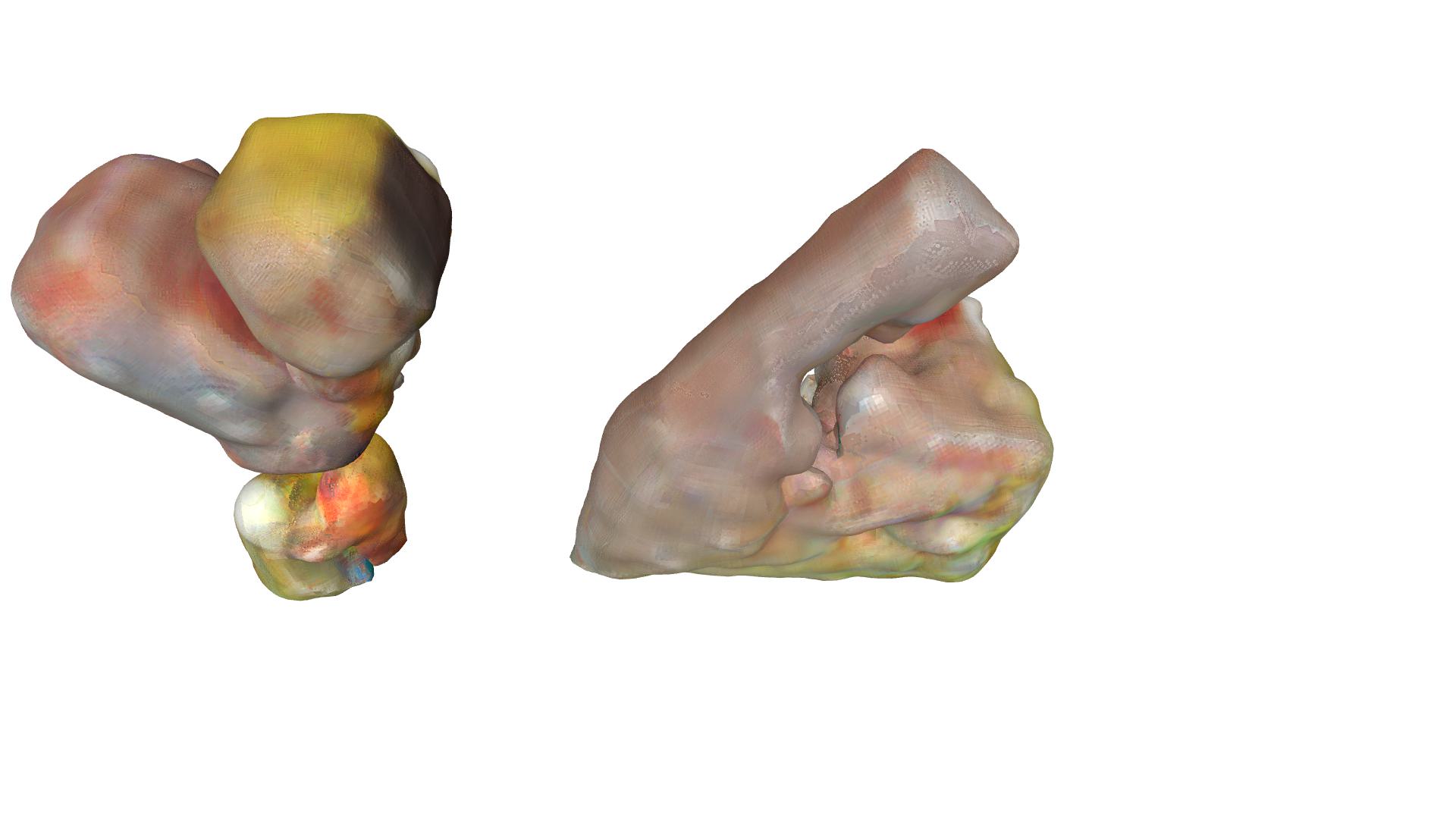}
        \end{tabular}
         &
        \hspace{\mrg}
        \begin{tabular}{c}
            \includegraphics[width=\wid]{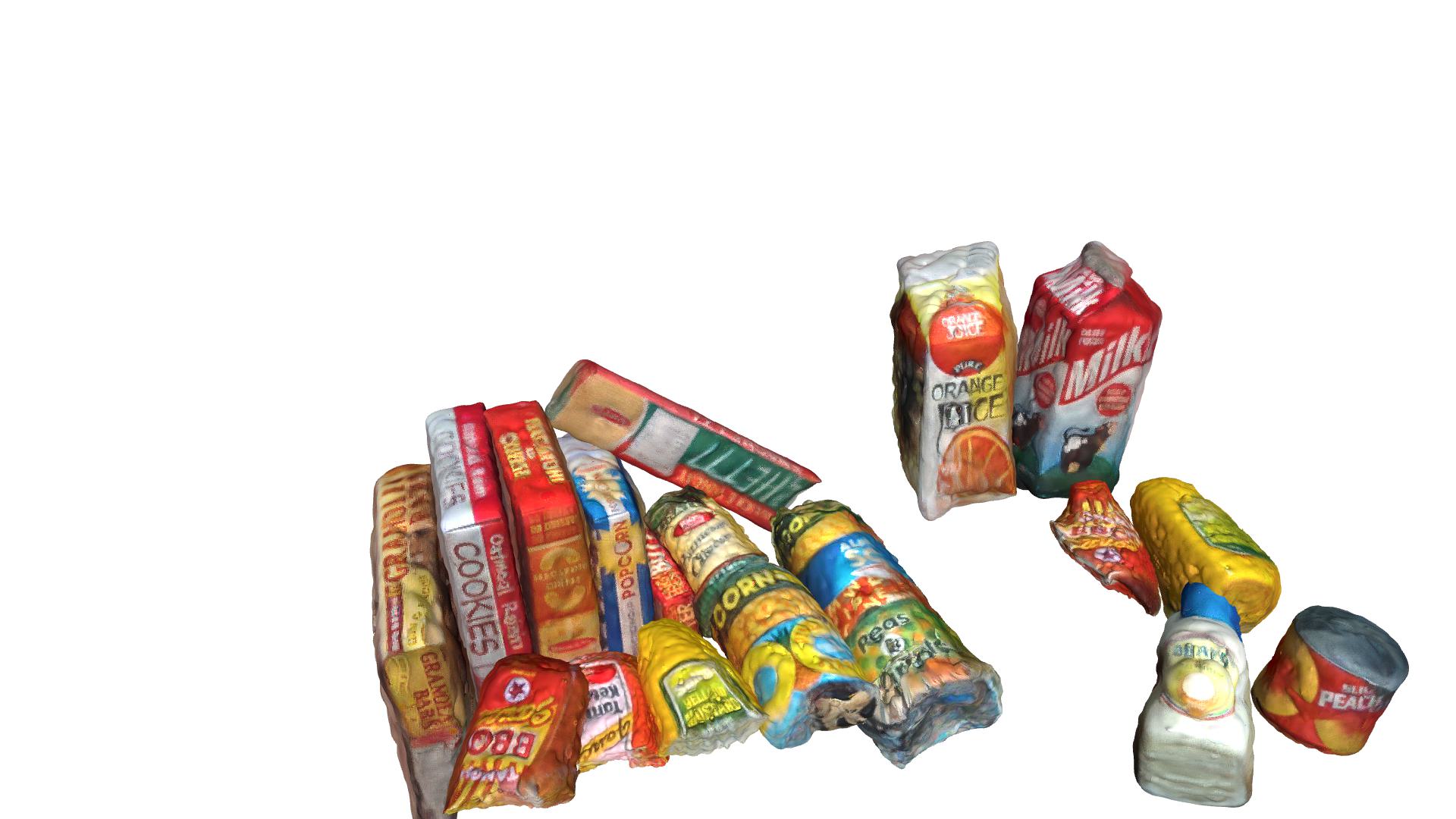} \\
            \includegraphics[width=\wid]{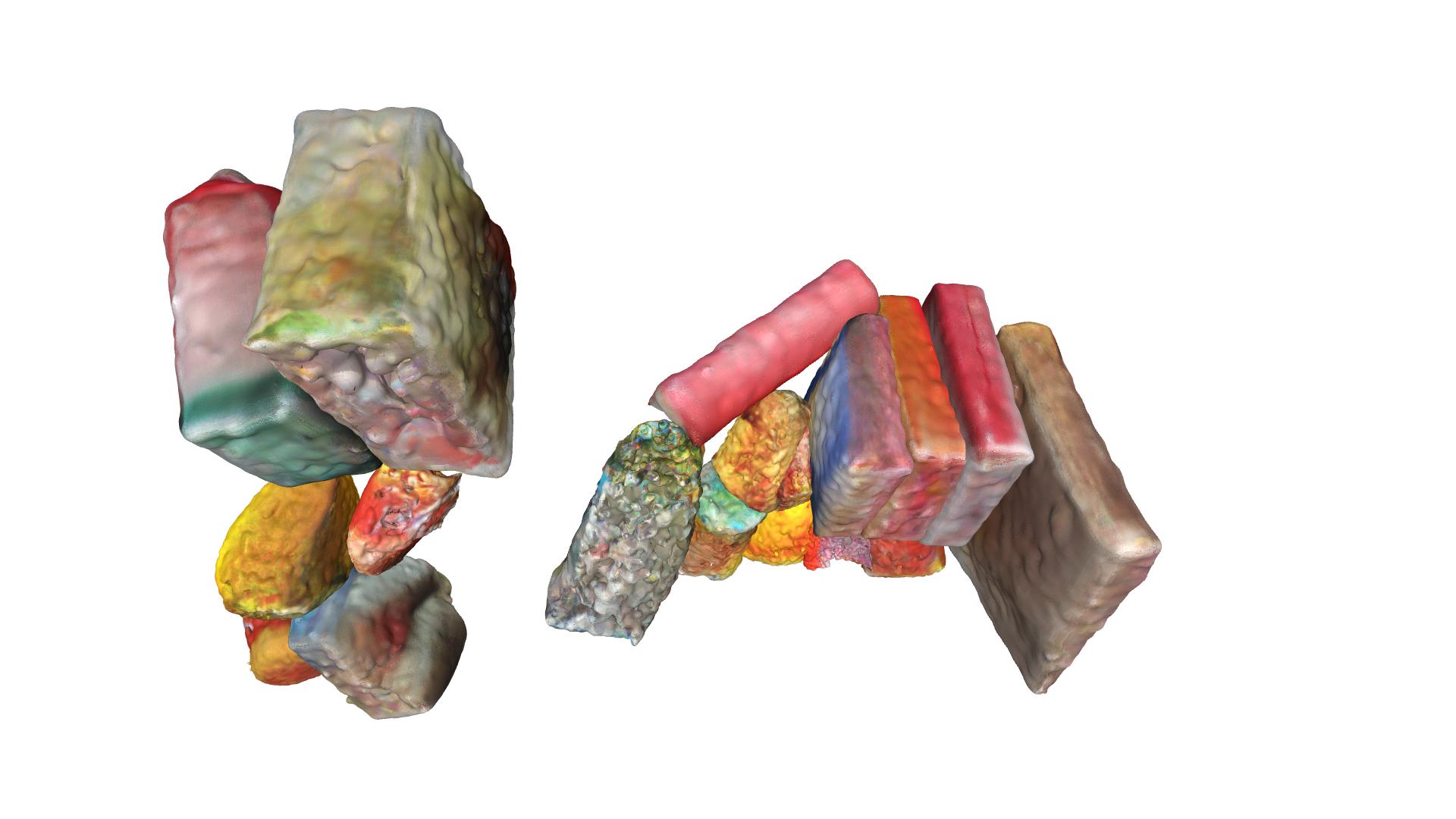}
        \end{tabular}
        & 
        \hspace{\mrg}
        \begin{tabular}{c}
            \includegraphics[width=\wid]{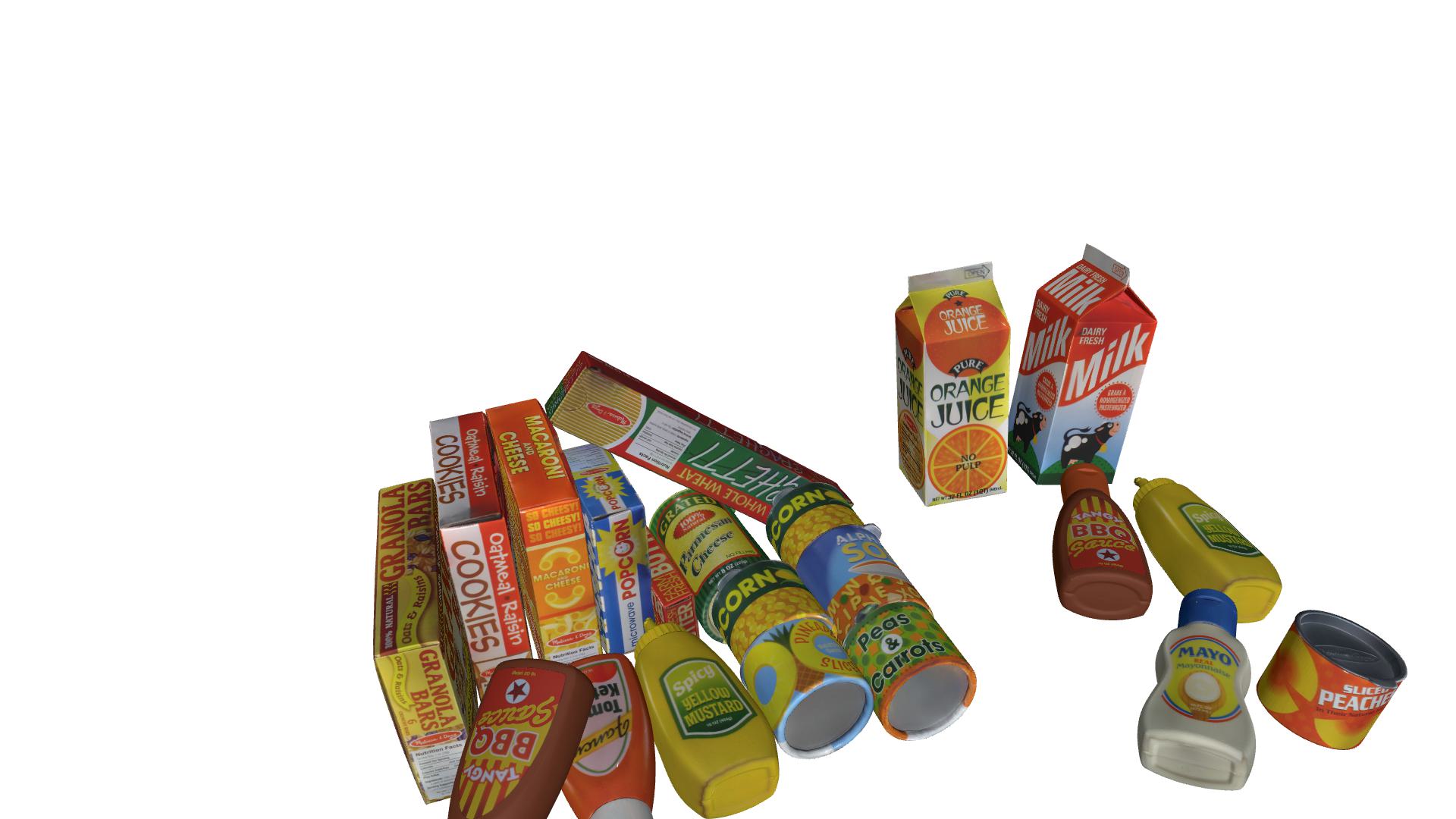}  \\
            \includegraphics[width=\wid]{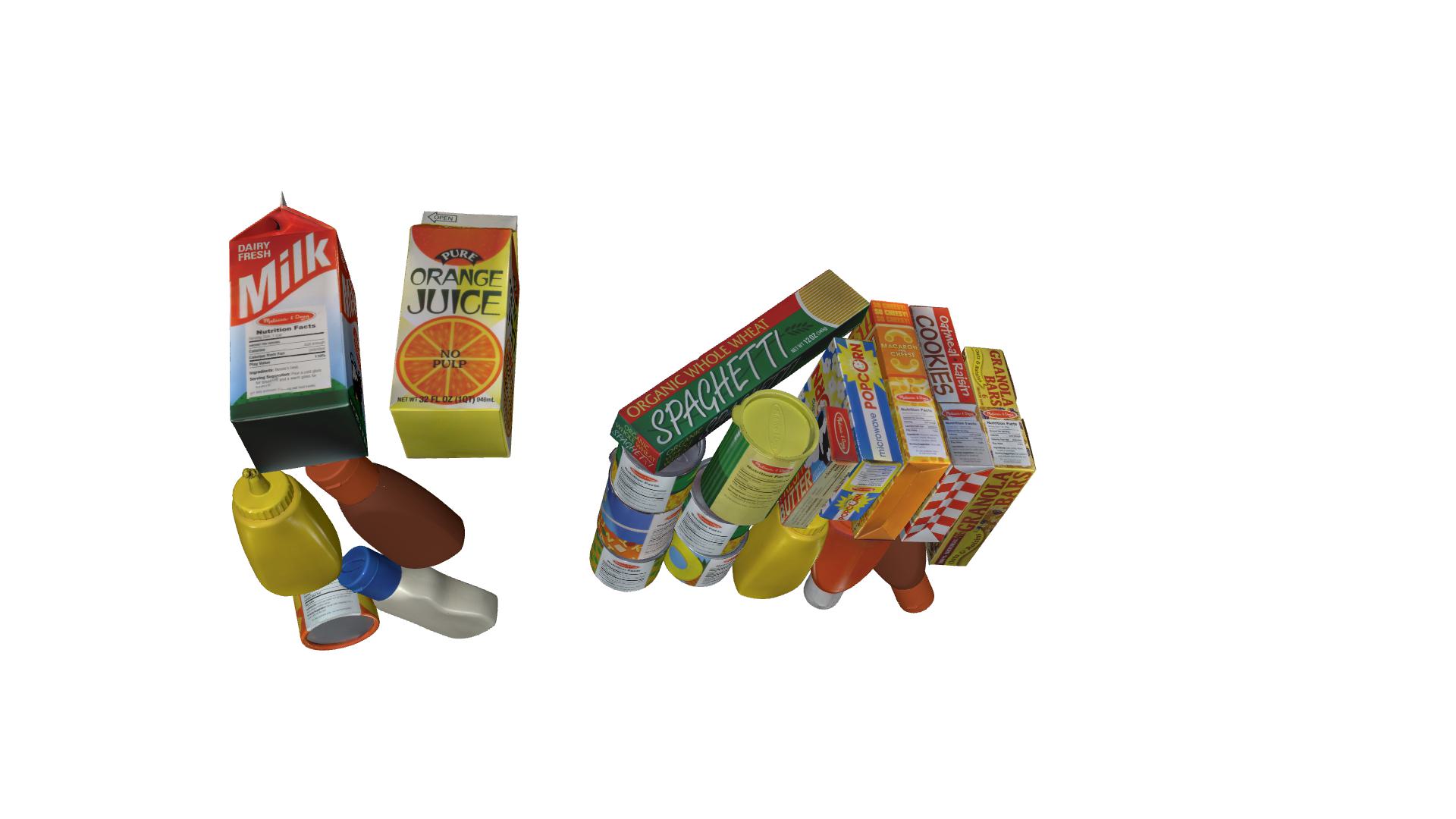}
        \end{tabular} \\
        \vspace{\mrgv}
       Input image & \hspace{\mrg}  DreamGaussian \cite{tang2023dreamgaussian} &  \hspace{\mrg}
        Ours
        & \hspace{\mrg}
        Ground Truth
    \end{tabular}
    \caption{Qualitative results on HOPE-Image \cite{tyree2022hope}. We show each reconstruction from two camera views, including the input one. 
    } 
    \label{fig:hope_qual}
    \vspace{-0.3cm}
\end{figure*}

We include in Figure \ref{fig:hope_qual} the qualitative results of the experiments on HOPE-Image dataset \cite{tyree2022hope}. We compare our compositional approach with DreamGaussian~\cite{tang2023dreamgaussian}, which reconstructs all the objects in the image at once. Our method can better handle complex scenes with many objects, as each instance is individually reconstructed. In contrast, both the appearance and the geometry of DreamGaussian's reconstructions degrades when applied on scenes with multiple objects. This is mainly due to limitations of the Zero-1-to-3 XL method, which fails to generate realistic, multi-view-consistent views. The pitfall is expected, as the domain required to be modeled by the prior increases exponentially with the number of objects. We do not include InstPIFu~\cite{liu2022towards} in the comparison on HOPE-Image dataset because the method fails to detect any of the considered objects in the scenes.

\subsection{Alternative models}
\label{sec:alternative}
\begin{table*}
    \centering
    \setlength{\tabcolsep}{0.5em}
    \begin{tabular}{ccccc}
        \\
        Depth estimation & stuff vs thing & Object reconstruction & Chamfer $\downarrow$ & F-Score $\uparrow$
        \\
        \hline
        Marigold & OneFormer & DreamGaussian & \textbf{0.099} & \textbf{75.33} \\
        \hdashline
        Depth Anything & OneFormer & DreamGaussian & 0.135 & 68.43 \\
        Marigold & CLIPSeg & DreamGaussian & 0.166 & 65.56 \\
        Marigold & OneFormer & One-2-3-45 & 0.110 & 69.40 \\
        \hline
    \end{tabular}
    \caption{Quantitative evaluation of our method using alternative models on 3D-FRONT~\cite{front20213d} dataset.}
    \label{tab:modules}
    \vspace{-0.3cm}
\end{table*}

The proposed method is designed as a modular framework, allowing the straightforward replacement of the integrated models summarized in \ref{sec:integrated}. 
We showcase this property of our method by exchanging the Marigold~\cite{ke2023repurposing} depth estimation with Depth Anything~\cite{depthanything}, the stuff-thing detection based on OneFormer~\cite{jain2023oneformer} with one based on CLIPSeg~\cite{lueddecke22clipseg}, and the single-view object reconstruction model, DreamGaussian~\cite{tang2023dreamgaussian}, with One-2-3-45~\cite{liu2023one}. We evaluate in Table \ref{tab:modules} the performance of these modifications on full scene reconstruction (background and foreground objects) using the 3D-FRONT~\cite{front20213d} dataset. For CLIPSeg, we define empirically a set of prompts for foreground (object, furniture) and background (background, floor, wall, curtain, window, ceiling), and consider foreground pixels the ones that have a lower than $0.5$ score for all the background prompts or higher than $0.1$ for any of the foreground ones. The performance of our method decreases in this case because DreamGaussian poorly reconstructs the background regions that are misclassified as things. We observe this happens more frequently on indoor rooms when using CLIPSeg, and on tabletop scenes, such as the one in our method figure from the main paper, when using OneFormer. 

The results show that our default configuration (as reported in the main paper) yields the best metrics on this dataset. Still, the alternative configurations are competitive, which proves that our pipeline is flexible and robust to different implementations of the modules.

\subsection{Ablations}
\label{sec:qual_ablations}

\begin{figure*}
    \centering    
    \setlength{\wid}{0.24\textwidth}
    \setlength{\mrg}{-0.5cm}
    \setlength{\mrgv}{-0.8cm}
    \begin{tabular}{ccccc}
        \vspace{\mrgv} &
        \vcentered{Ours\textsubscript{DG} \\ w/o reprojection} &
        \hspace{\mrg}
        \includegraphics[width=\wid, align=c]{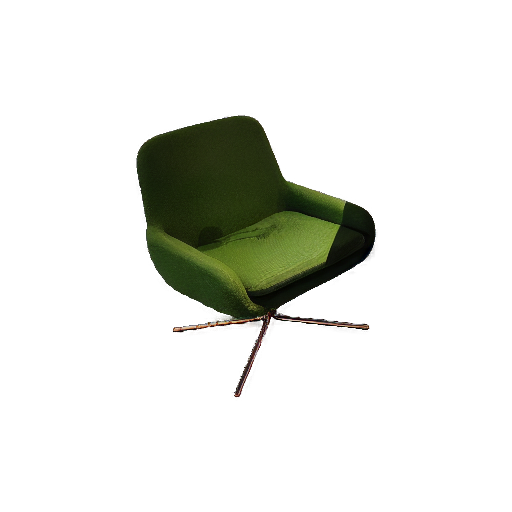} &
        \hspace{\mrg}
        \includegraphics[width=\wid, align=c]{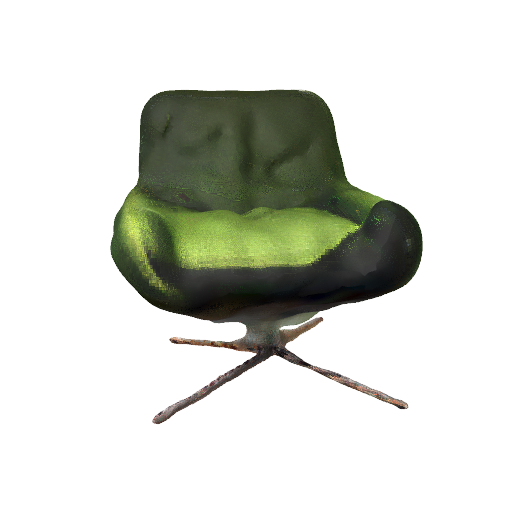} & 
        \hspace{\mrg}
        \includegraphics[width=\wid, align=c]{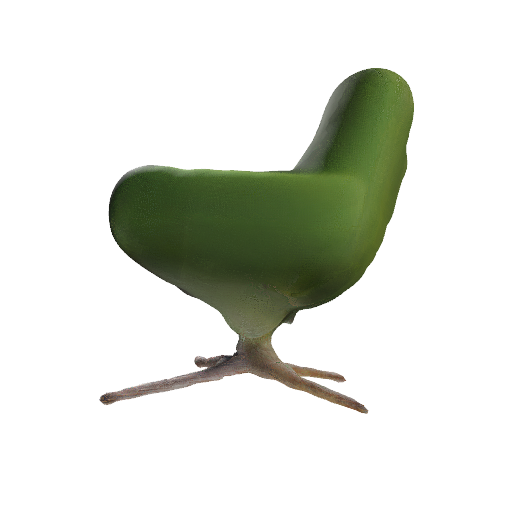} 
        \\ 
        \vspace{\mrgv} &
        \vcentered{Ours\textsubscript{DG} \\ w/o amodal \\ completion} &
        \hspace{\mrg}
        \includegraphics[width=\wid, align=c]{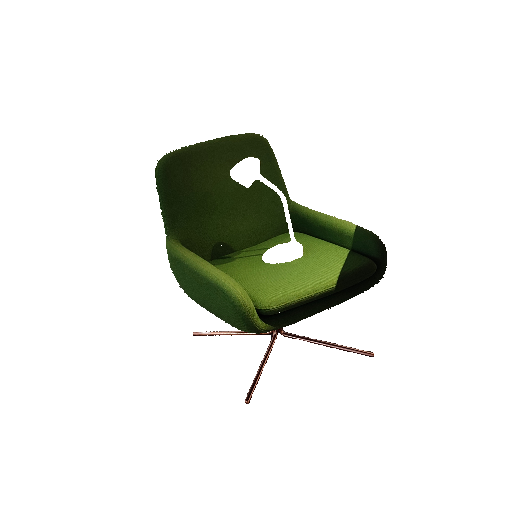} &
        \hspace{\mrg}
        \includegraphics[width=\wid, align=c]{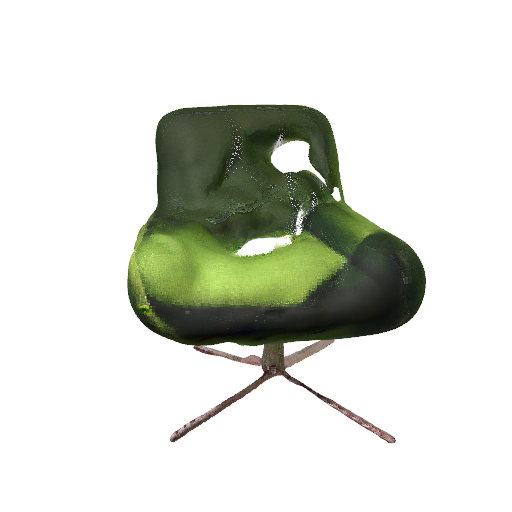} & 
        \hspace{\mrg}
        \includegraphics[width=\wid, align=c]{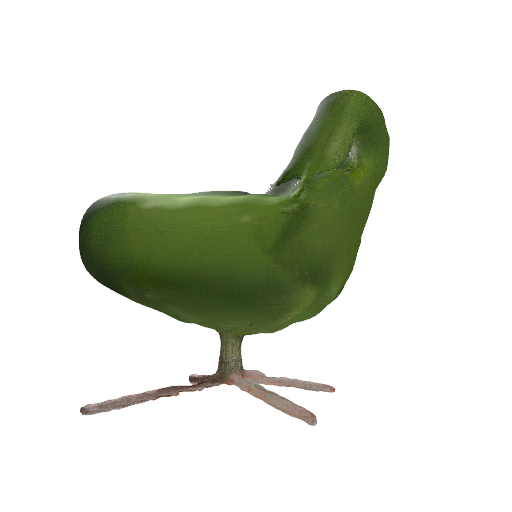} 
        \\
        \vspace{\mrgv} &
        \vcentered{Ours\textsubscript{DG} \\ (full)} &
        \hspace{\mrg}
        \includegraphics[width=\wid, align=c]{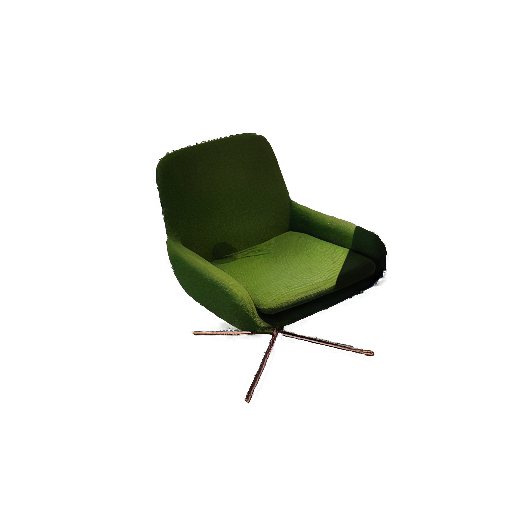} &
        \hspace{\mrg}
        \includegraphics[width=\wid, align=c]{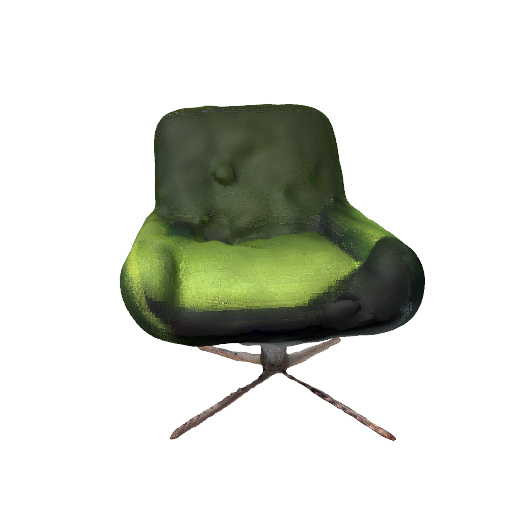} & 
        \hspace{\mrg}
        \includegraphics[width=\wid, align=c]{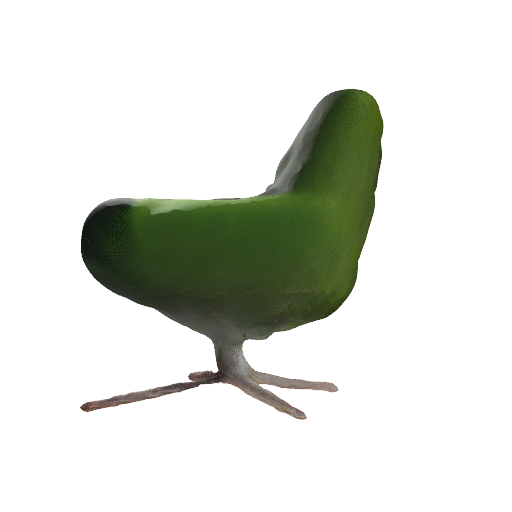} 
        \\
        \vspace{\mrgv} &
        \vcentered{Ours\textsubscript{DG} \\ w/o reprojection} &
        \hspace{\mrg}
        \includegraphics[width=\wid, align=c]{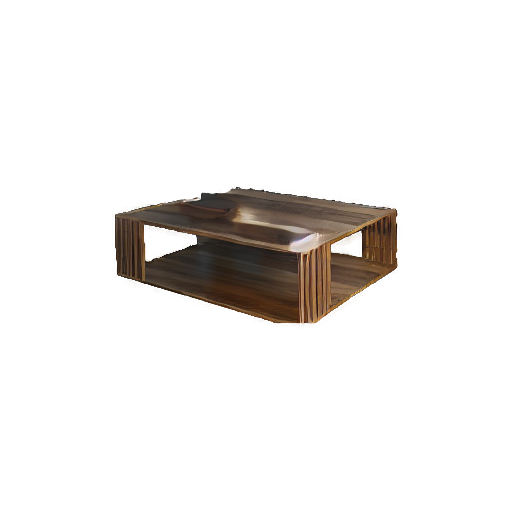} &
        \hspace{\mrg}
        \includegraphics[width=\wid, align=c]{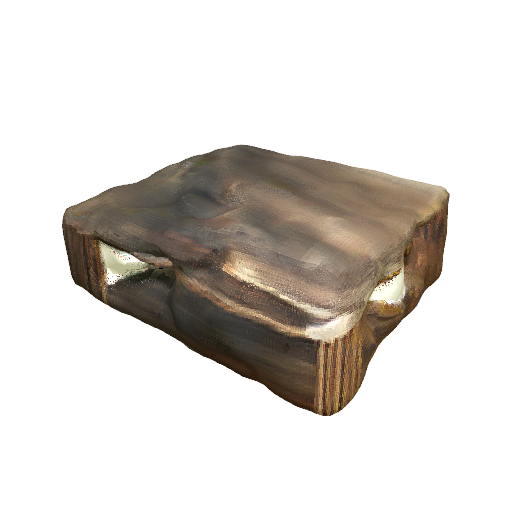} & 
        \hspace{\mrg}
        \includegraphics[width=\wid, align=c]{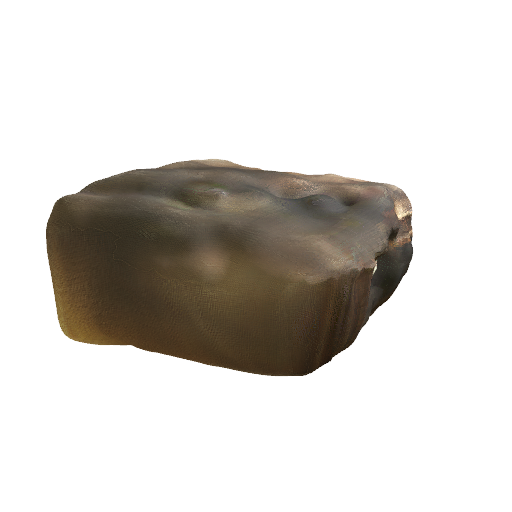} 
        \\ 
        \vspace{\mrgv} &
        \vcentered{Ours\textsubscript{DG} \\ w/o amodal \\ completion} &
        \hspace{\mrg}
        \includegraphics[width=\wid, align=c]{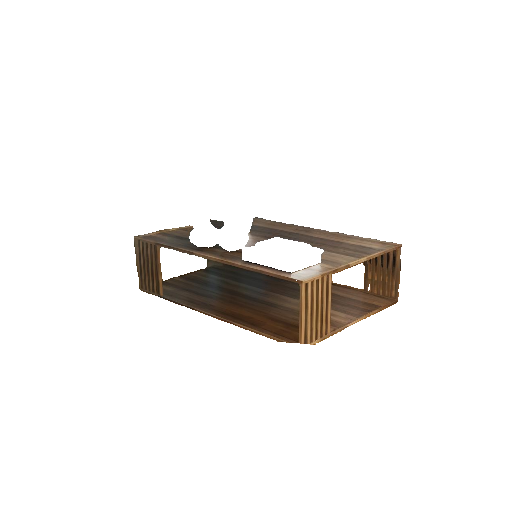} &
        \hspace{\mrg}
        \includegraphics[width=\wid, align=c]{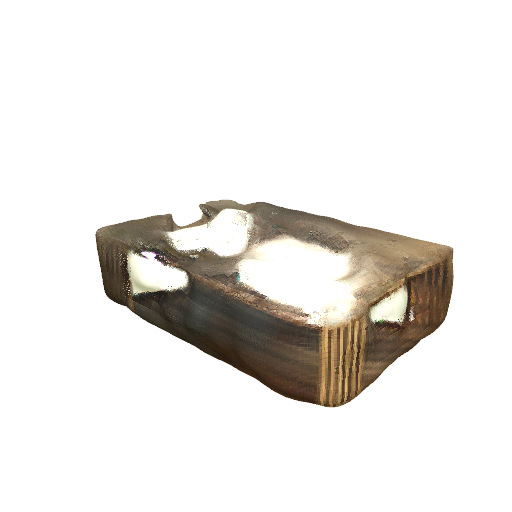} & 
        \hspace{\mrg}
        \includegraphics[width=\wid, align=c]{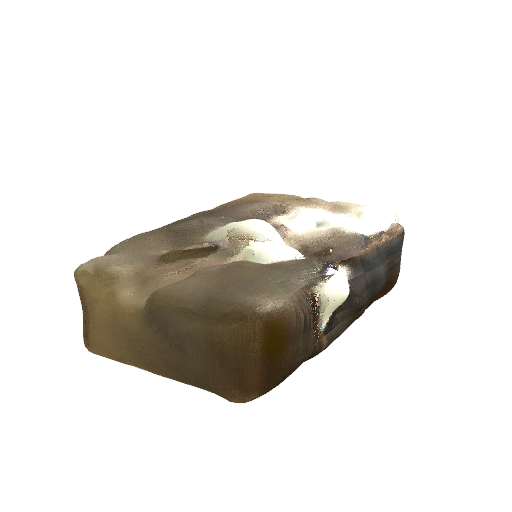} 
        \\
         &
        \vcentered{Ours\textsubscript{DG} \\ (full)} &
        \hspace{\mrg}
        \includegraphics[width=\wid, align=c]{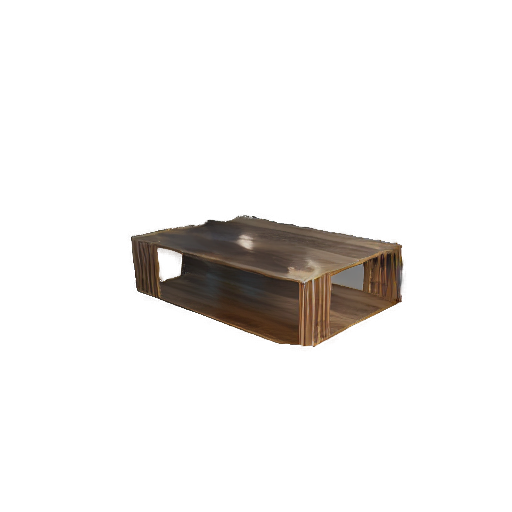} &
        \hspace{\mrg}
        \includegraphics[width=\wid, align=c]{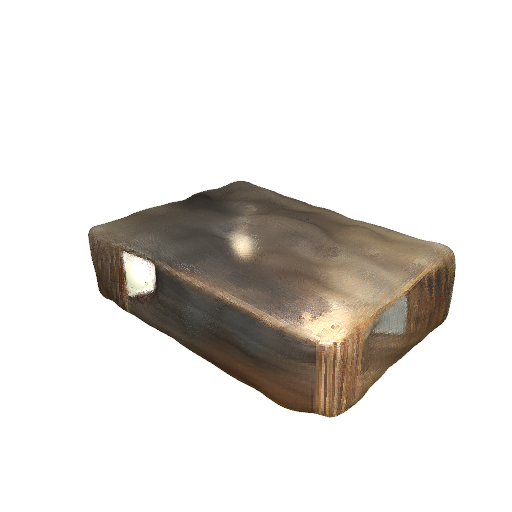} & 
        \hspace{\mrg}
        \includegraphics[width=\wid, align=c]{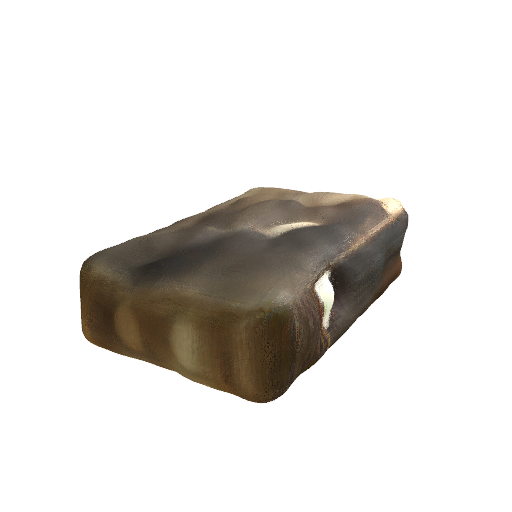} 
        \\
         &
        Method & Input crop & \multicolumn{2}{c}{Reconstructed object} 
        \\
    \end{tabular}
    \caption{Ablations visual comparison. In Ours\textsubscript{DG} we use DreamGaussian~\cite{tang2023dreamgaussian} for reconstructing individual objects with the proposed reprojection of the input pixels and amodal completion of the missing parts. 
    } 
    \label{fig:ablation}
\end{figure*}

\begin{figure*}
    \centering    
    \setlength{\wid}{0.24\textwidth}
    \setlength{\mrg}{-0.5cm}
    \setlength{\mrgv}{-0.8cm}
    \begin{tabular}{ccccc}
        \vspace{\mrgv} &
        \vcentered{Ours\textsubscript{LaRa} \\ w/o reprojection} &
        \hspace{\mrg}
        \includegraphics[width=\wid, align=c]{figures_sup/ablation/0_worep.png} &
        \hspace{\mrg}
        \includegraphics[width=\wid, align=c]{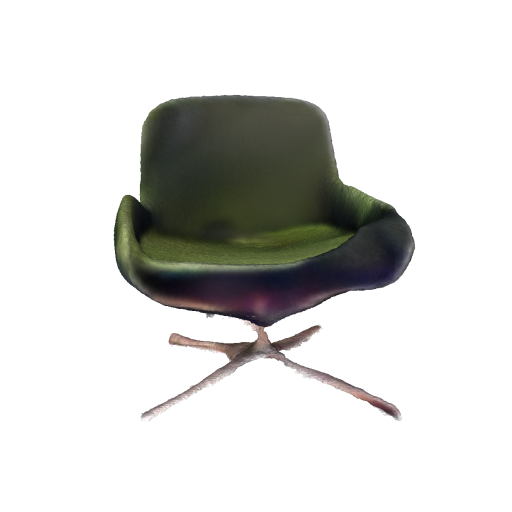} & 
        \hspace{\mrg}
        \includegraphics[width=\wid, align=c]{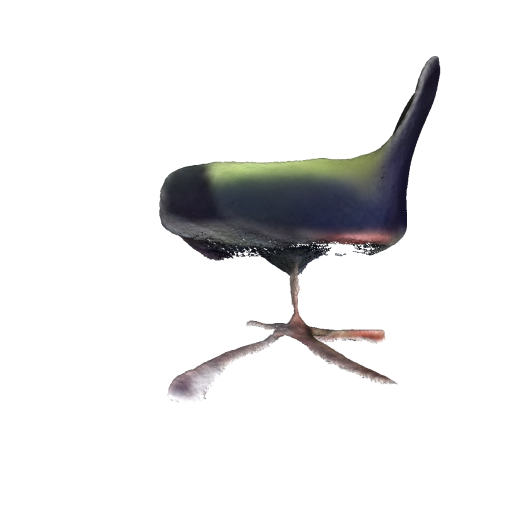} 
        \\ 
        \vspace{\mrgv} &
        \vcentered{Ours\textsubscript{LaRa} \\ w/o amodal \\ completion} &
        \hspace{\mrg}
        \includegraphics[width=\wid, align=c]{figures_sup/ablation/0_wocomp.png} &
        \hspace{\mrg}
        \includegraphics[width=\wid, align=c]{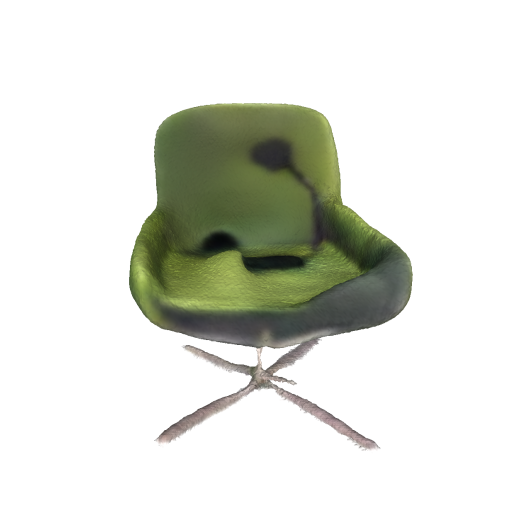} & 
        \hspace{\mrg}
        \includegraphics[width=\wid, align=c]{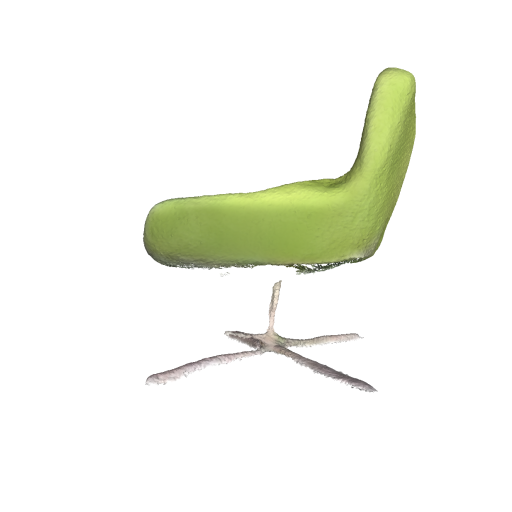} 
        \\
        \vspace{\mrgv} &
        \vcentered{Ours\textsubscript{LaRa} \\ (full)} &
        \hspace{\mrg}
        \includegraphics[width=\wid, align=c]{figures_sup/ablation/0_full.png} &
        \hspace{\mrg}
        \includegraphics[width=\wid, align=c]{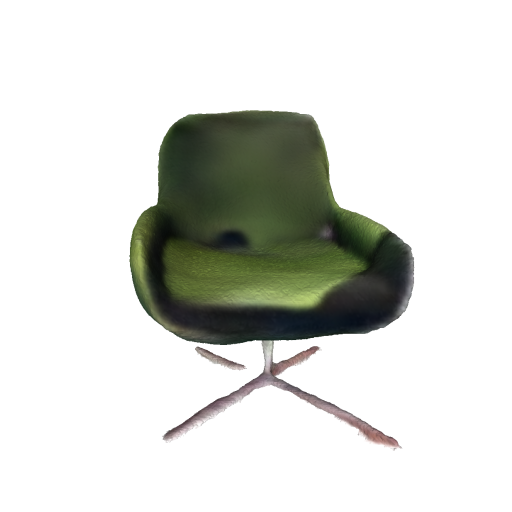} & 
        \hspace{\mrg}
        \includegraphics[width=\wid, align=c]{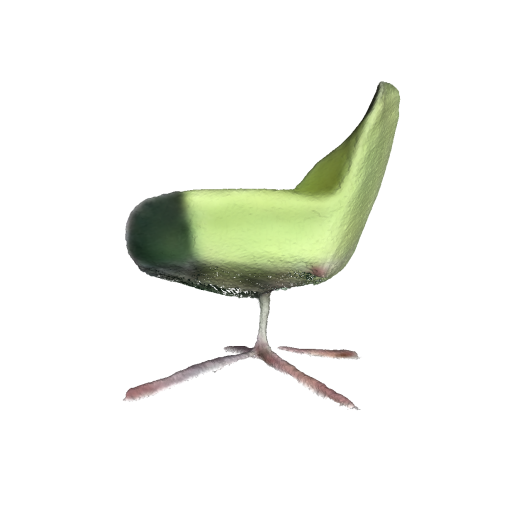} 
        \\
        \vspace{\mrgv} &
        \vcentered{Ours\textsubscript{LaRa} \\ w/o reprojection} &
        \hspace{\mrg}
        \includegraphics[width=\wid, align=c]{figures_sup/ablation/1_worep.png} &
        \hspace{\mrg}
        \includegraphics[width=\wid, align=c]{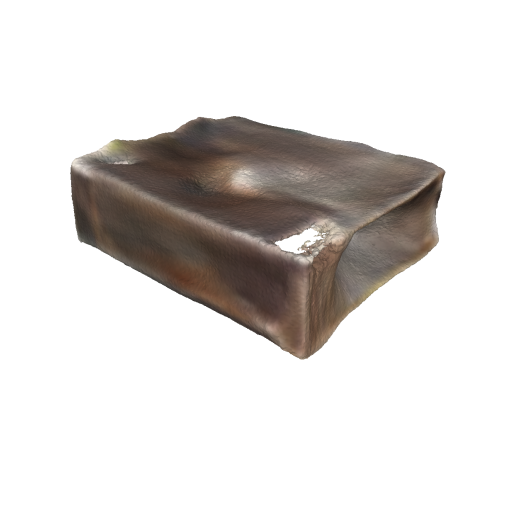} & 
        \hspace{\mrg}
        \includegraphics[width=\wid, align=c]{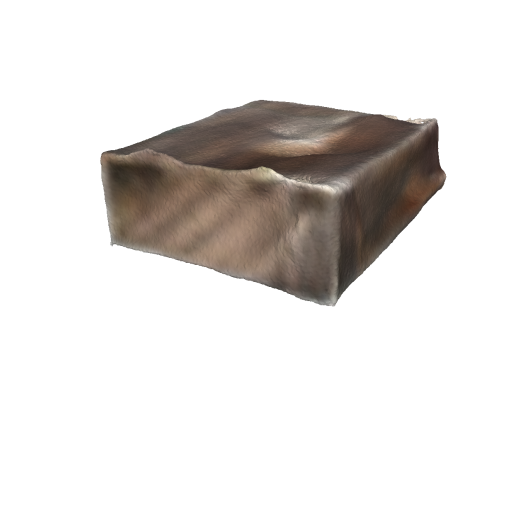} 
        \\ 
        \vspace{\mrgv} &
        \vcentered{Ours\textsubscript{LaRa} \\ w/o amodal \\ completion} &
        \hspace{\mrg}
        \includegraphics[width=\wid, align=c]{figures_sup/ablation/1_wocomp.png} &
        \hspace{\mrg}
        \includegraphics[width=\wid, align=c]{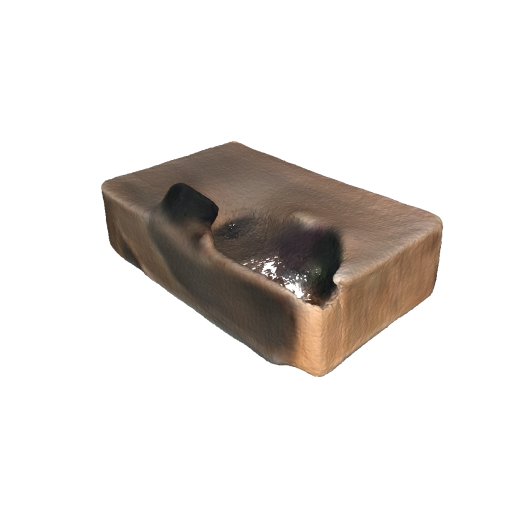} & 
        \hspace{\mrg}
        \includegraphics[width=\wid, align=c]{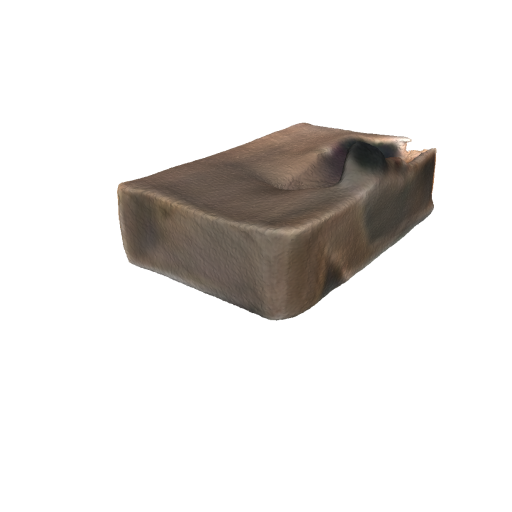} 
        \\
         &
        \vcentered{Ours\textsubscript{LaRa} \\ (full)} &
        \hspace{\mrg}
        \includegraphics[width=\wid, align=c]{figures_sup/ablation/1_full.png} &
        \hspace{\mrg}
        \includegraphics[width=\wid, align=c]{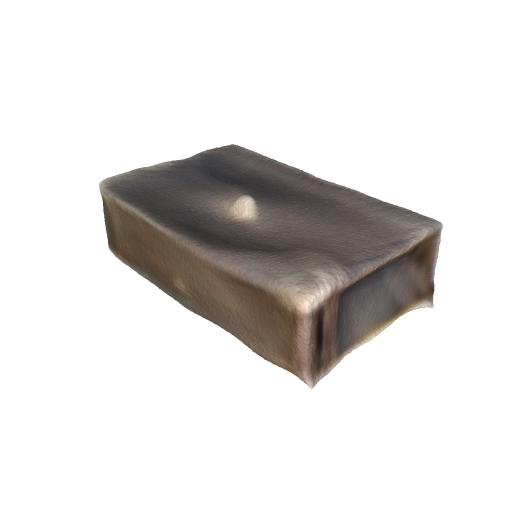} & 
        \hspace{\mrg}
        \includegraphics[width=\wid, align=c]{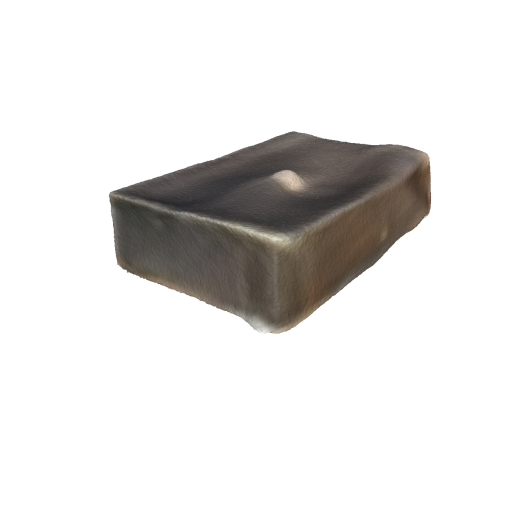} 
        \\
         &
        Method & Input crop & \multicolumn{2}{c}{Reconstructed object} 
        \\
    \end{tabular}
    \caption{Ablations visual comparison. In Ours\textsubscript{LaRa} we use LaRa~\cite{LaRa} for reconstructing individual objects with the proposed reprojection of the input pixels and amodal completion of the missing parts. 
    } 
    \label{fig:ablation_lara}
\end{figure*}

We provide in Figure \ref{fig:ablation} qualitative results corresponding to the two ablations considered in the main paper (see section 4.3). 
When ablating the amodal completion step, the reconstructed objects have holes or incorrect textures, corresponding to the regions that were occluded by other objects in the scenes. When skipping the reprojection step of the pipeline and directly using crops from the input image, the objects are reconstructed with deformed shapes to match the input view. By including both components, our full method reconstructs complete objects with a physically-correct geometry. 
We also consider these ablations together with LaRa~\cite{LaRa}, a very recent method for single view object reconstruction that operates in a feed-forward fashion. Compared to DreamGaussian which relies on Zero-1-to-3 XL to generate novel views conditioned on the input crop, LaRa uses a newer model, Zero123++, that can generate six fixed views that are more 3D consistent. As can be seen in Figure \ref{fig:ablation_lara}, the proposed amodal completion and reprojection are also beneficial for LaRa, boosting the reconstruction performance.

\subsection{Outdoor scenes}
\label{sec:outdoor}

\begin{figure*}
    \centering    
    \setlength{\wid}{0.24\textwidth}
    \setlength{\mrg}{-0.1cm}
    \setlength{\mrgv}{0cm}
    \begin{tabular}{cccc}
        \includegraphics[width=\wid]{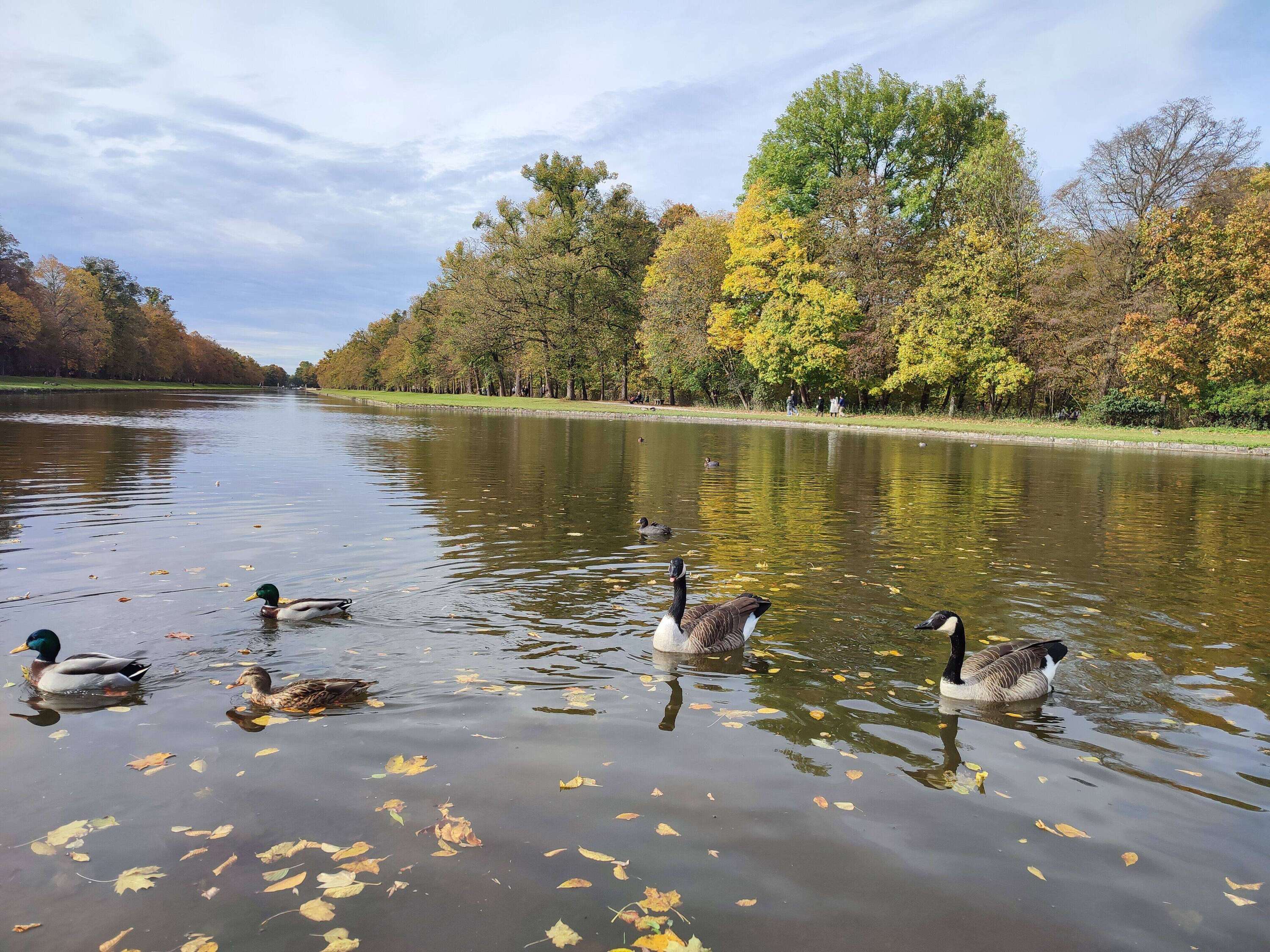} &
        \hspace{\mrg}
        \includegraphics[width=\wid]{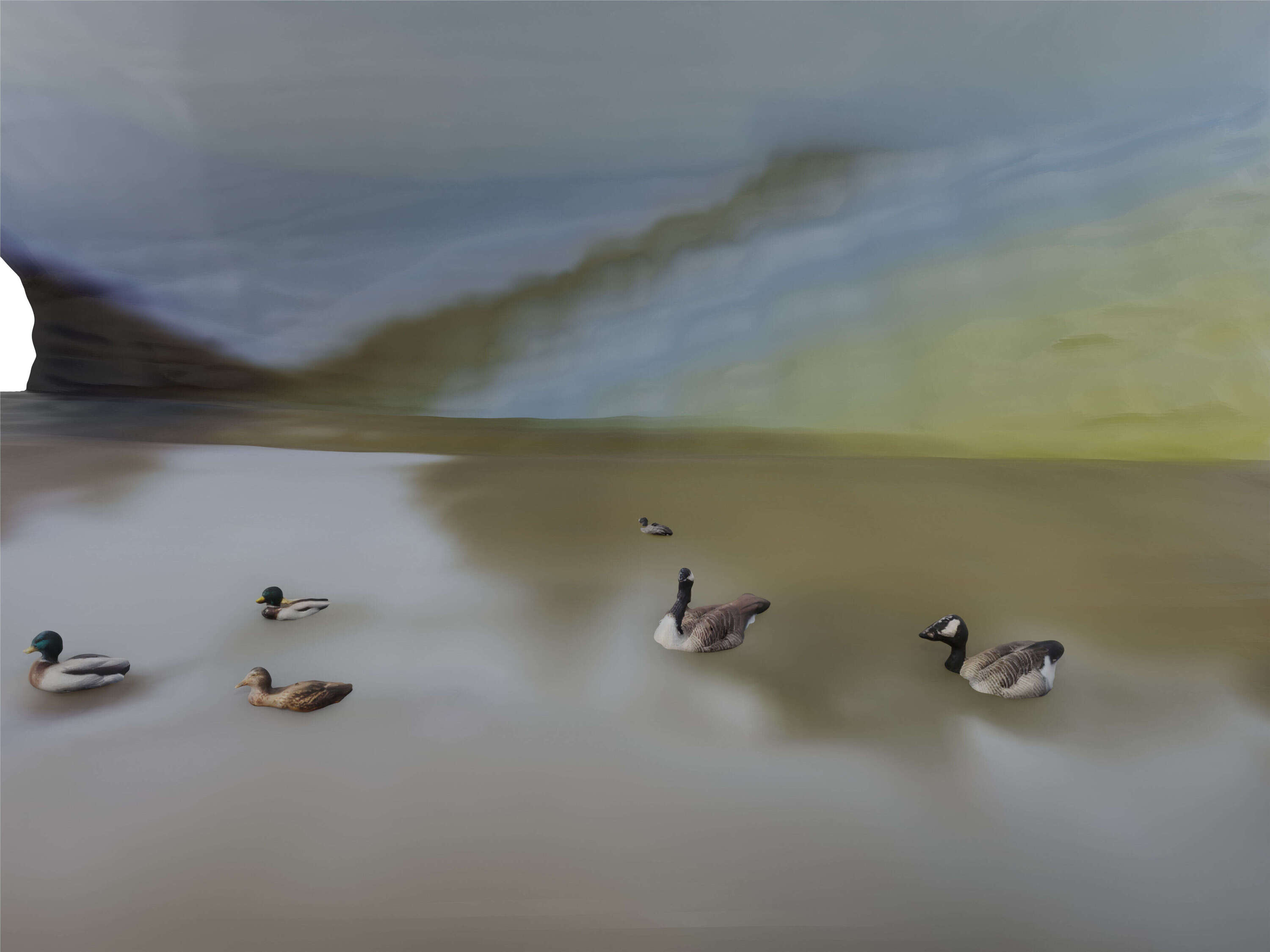} &
        \hspace{\mrg}
        \includegraphics[width=\wid]{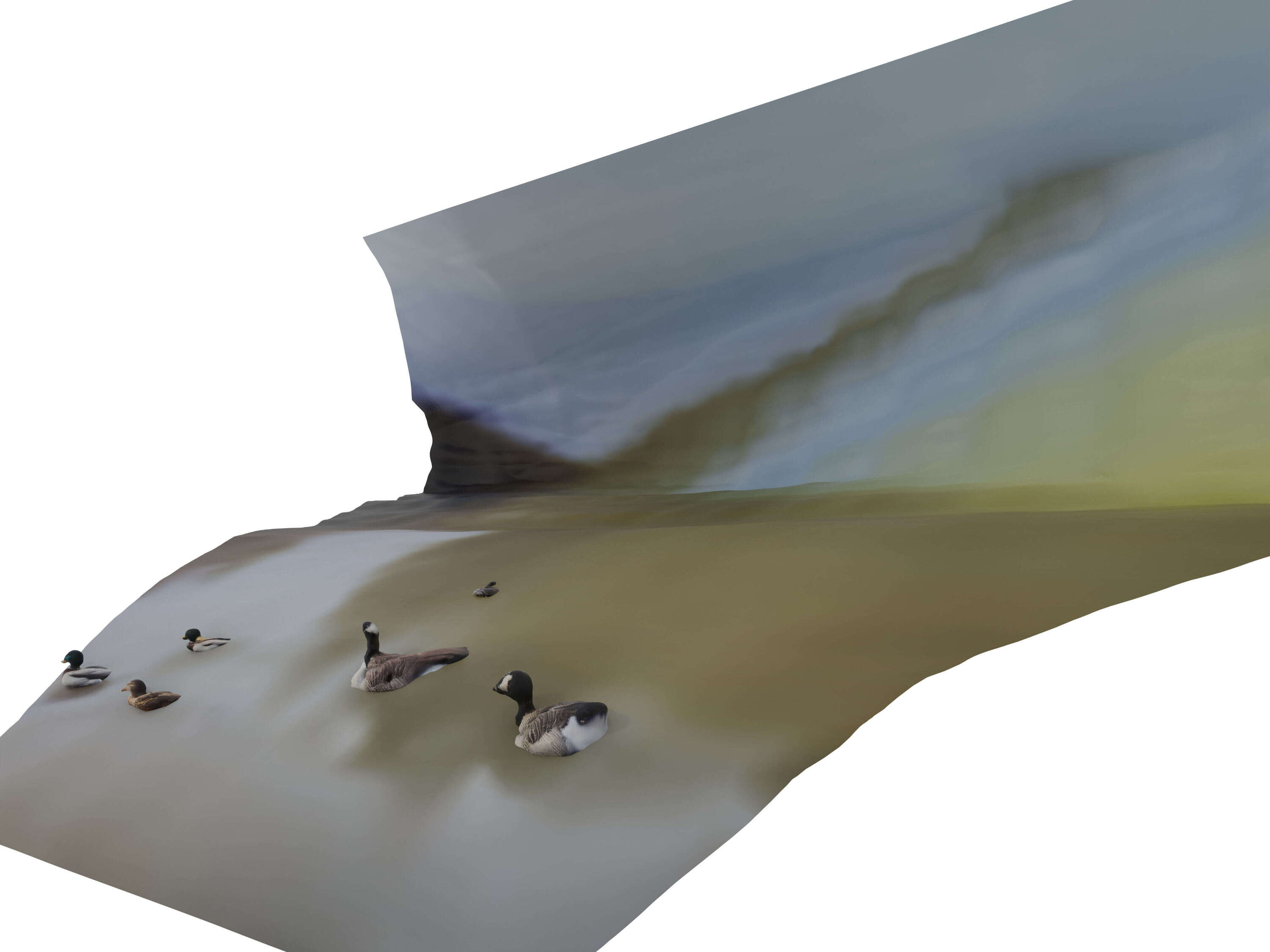} & 
        \hspace{\mrg}
        \includegraphics[width=\wid]{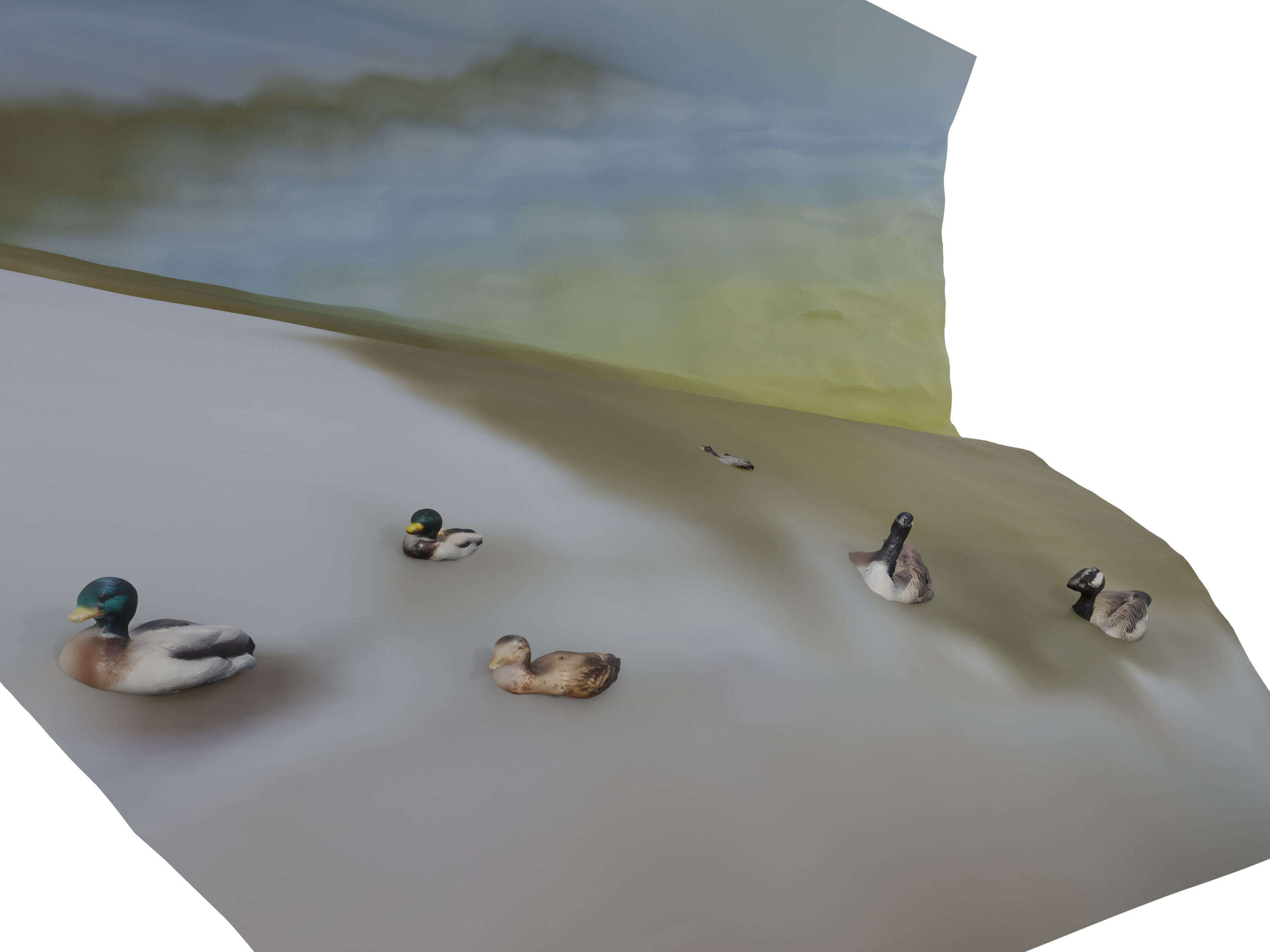} 
        \\ 
        \includegraphics[width=\wid]{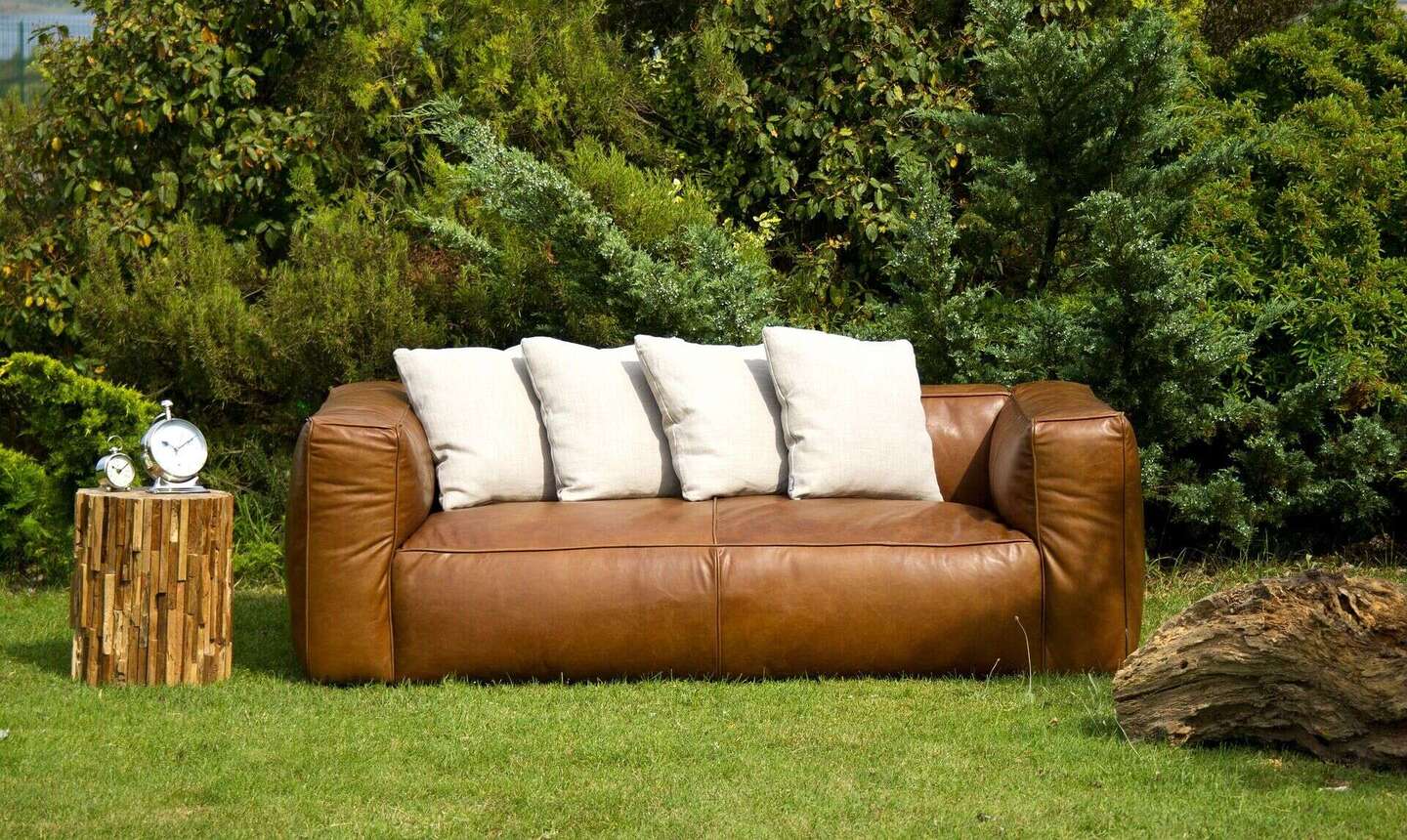} &
        \hspace{\mrg}
        \includegraphics[width=\wid]{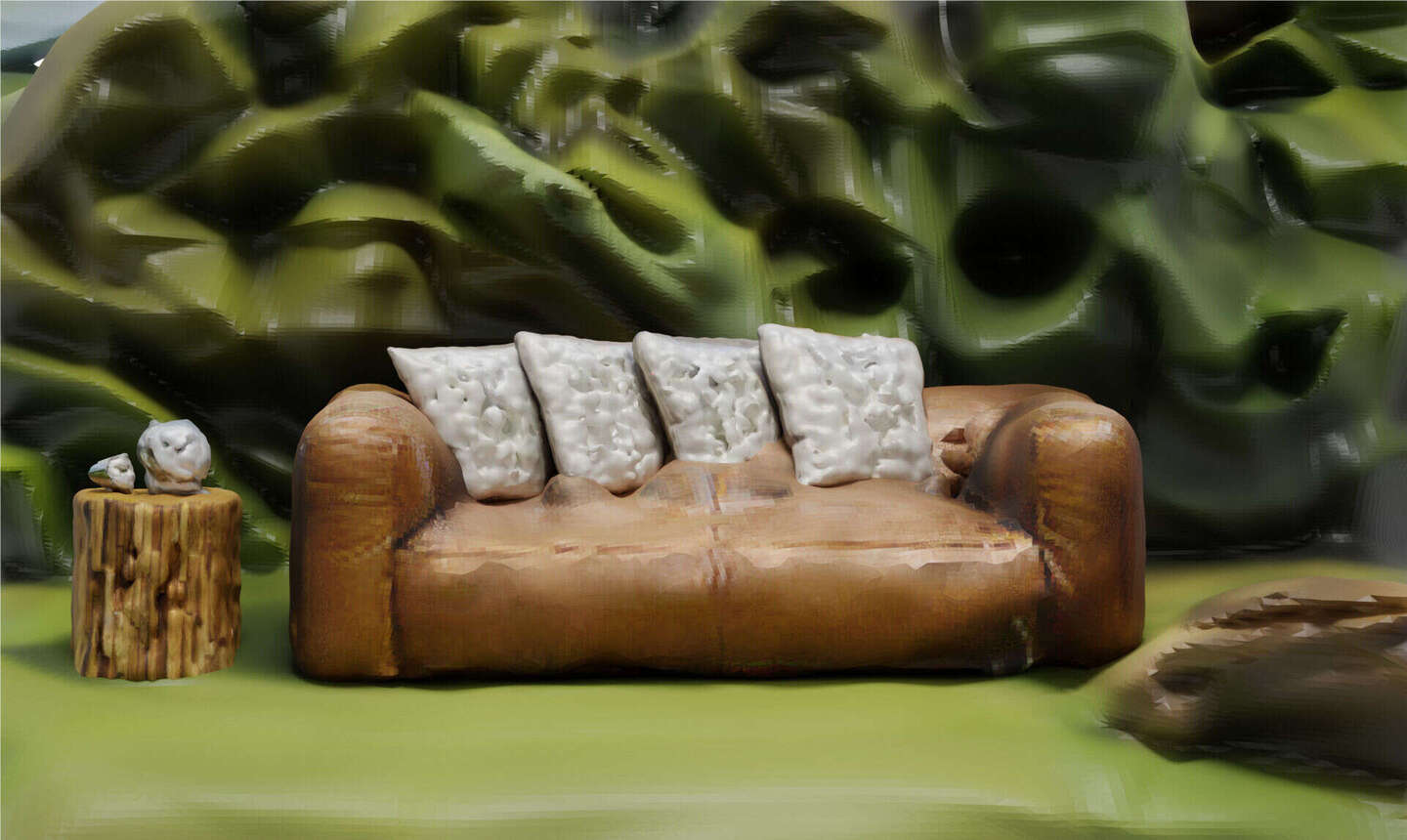} &
        \hspace{\mrg}
        \includegraphics[width=\wid]{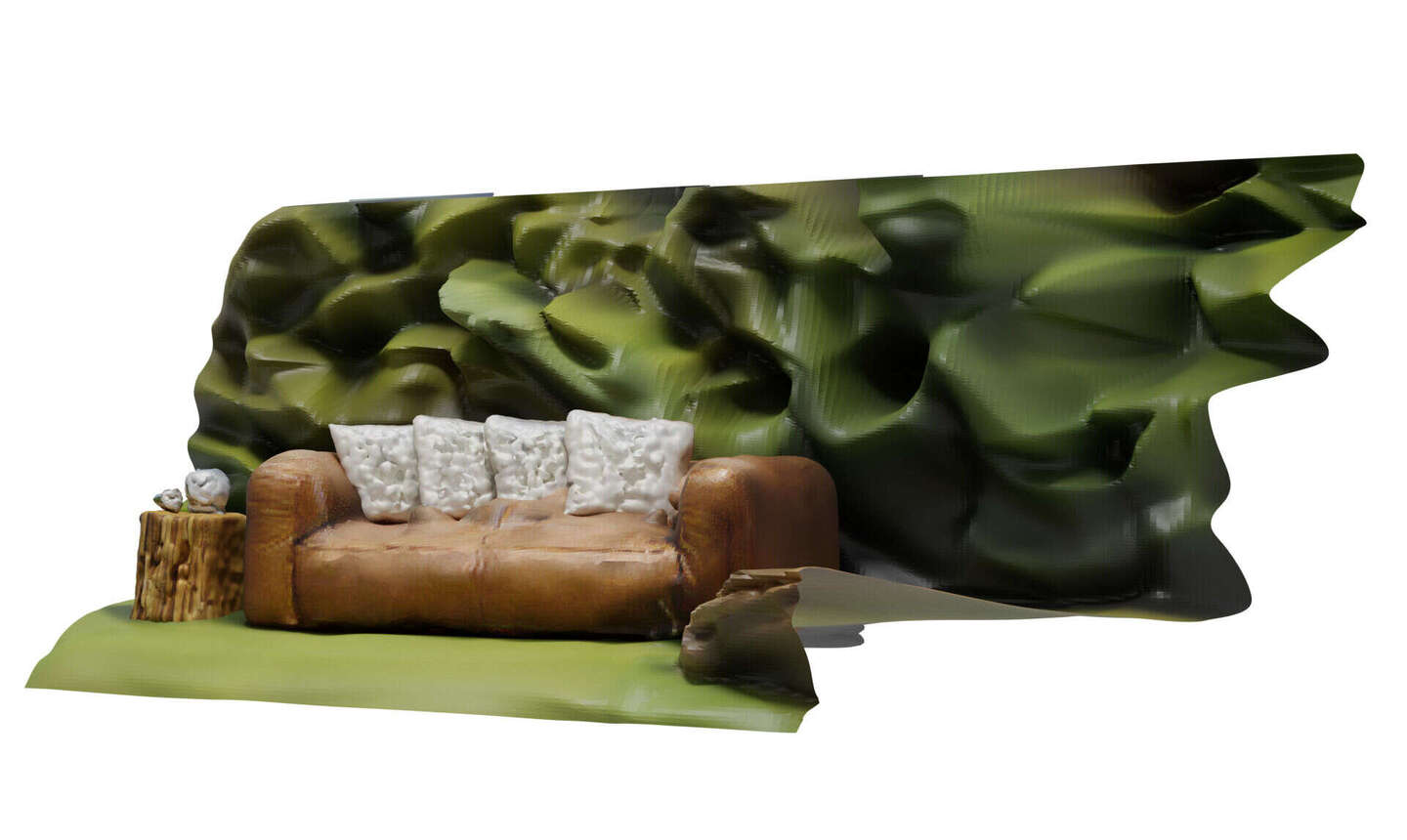} & 
        \hspace{\mrg}
        \includegraphics[width=\wid]{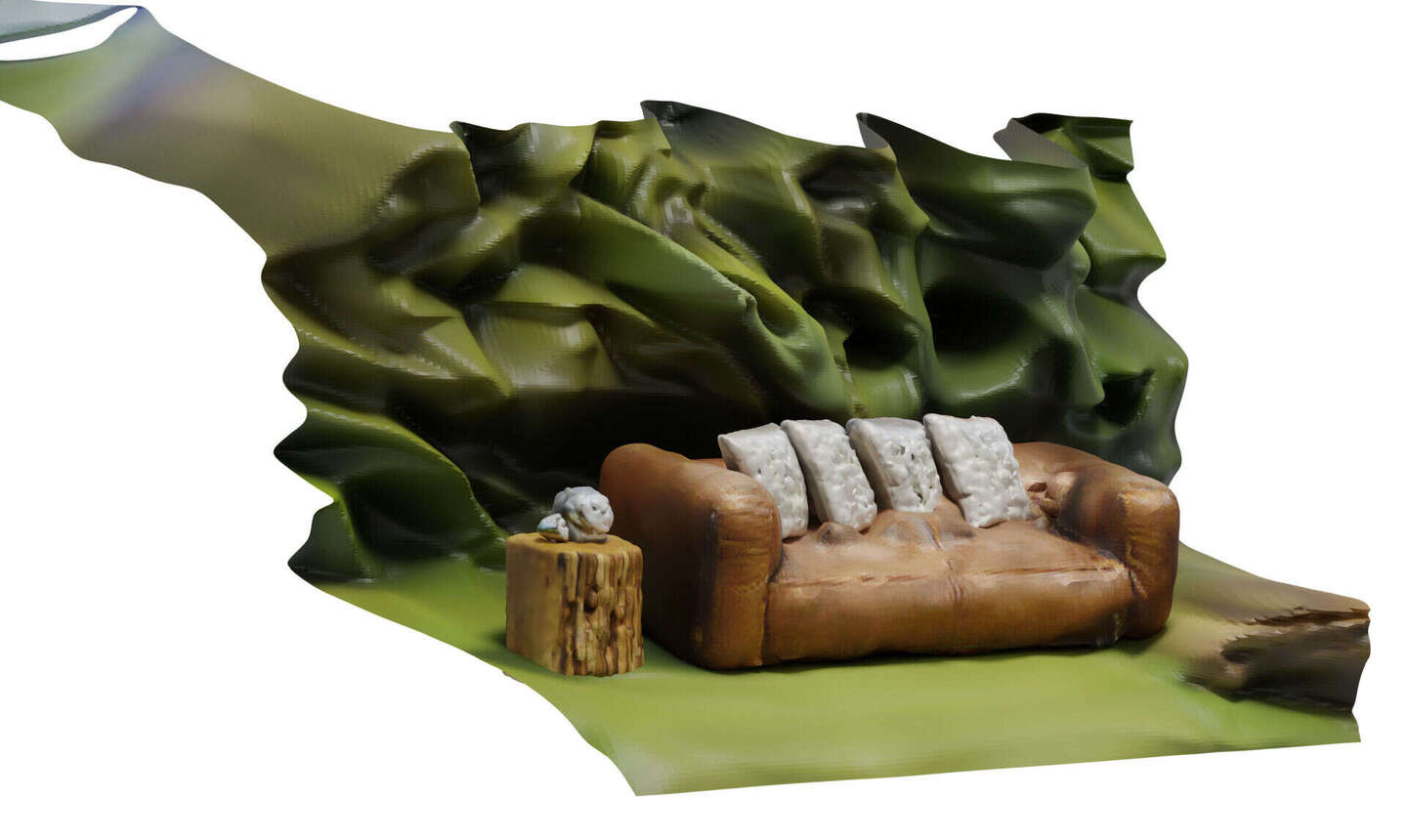} 
        \\ 
        \includegraphics[width=\wid]{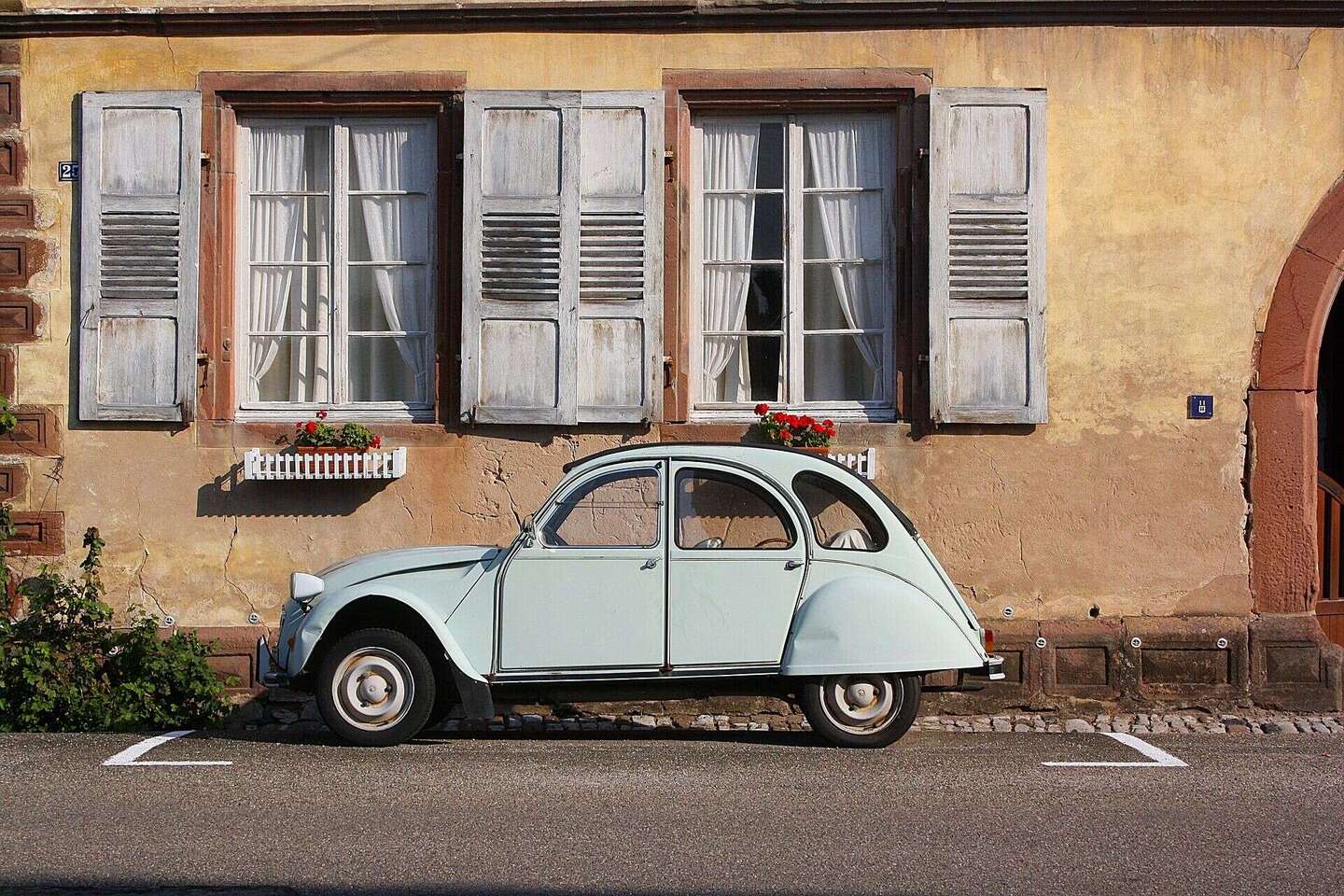} &
        \hspace{\mrg}
        \includegraphics[width=\wid]{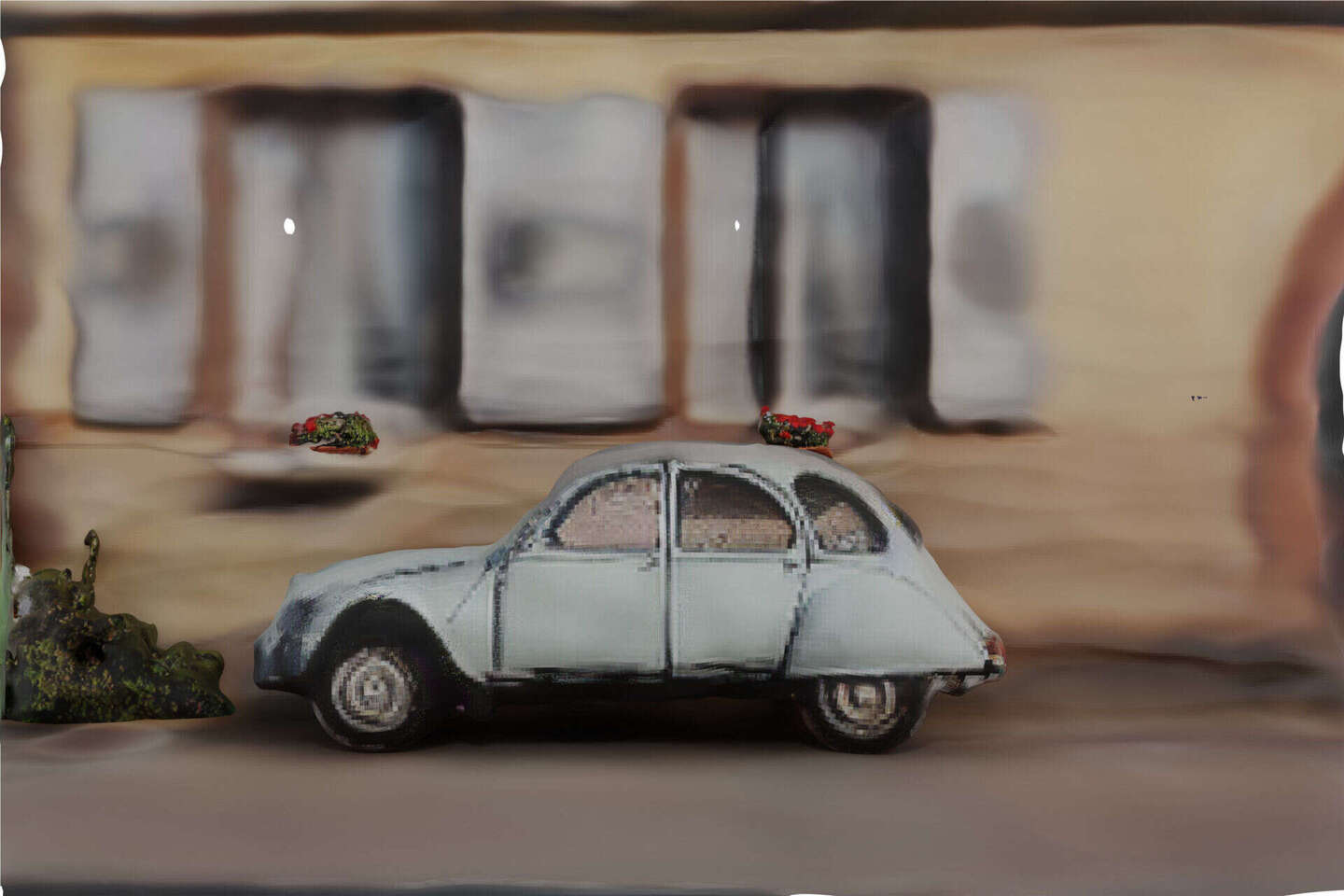} &
        \hspace{\mrg}
        \includegraphics[width=\wid]{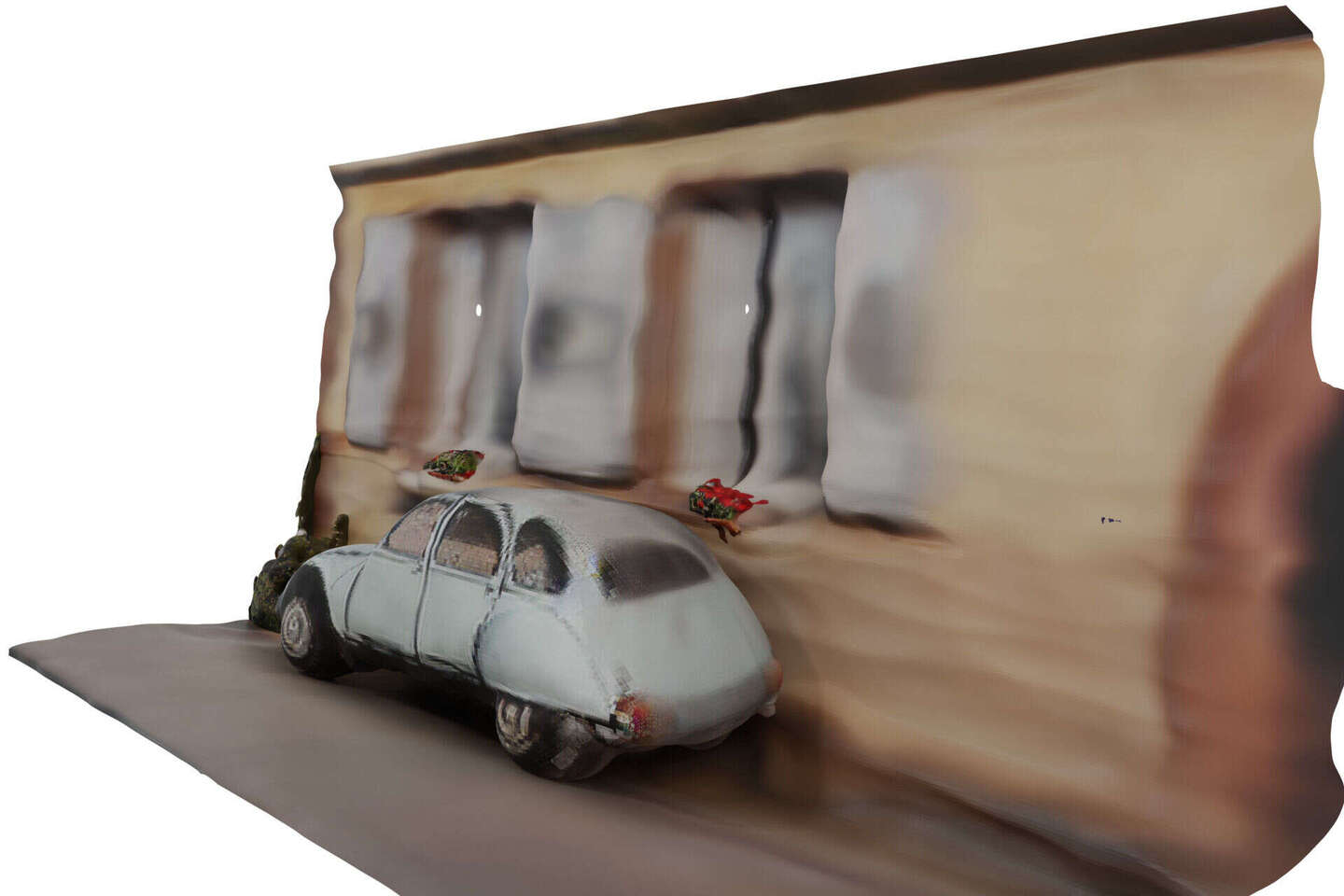} & 
        \hspace{\mrg}
        \includegraphics[width=\wid]{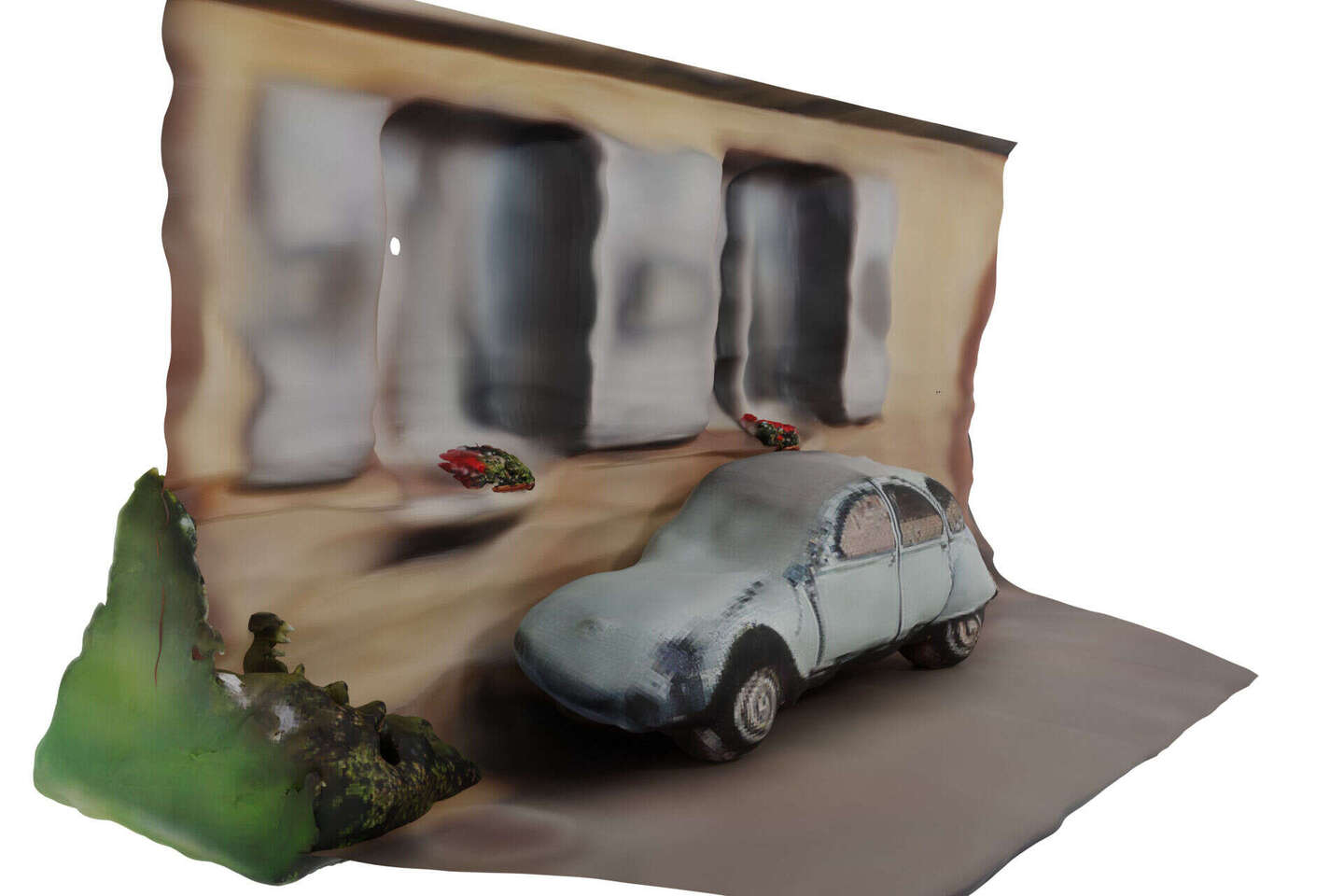} 
        \\ 
        \includegraphics[width=\wid]{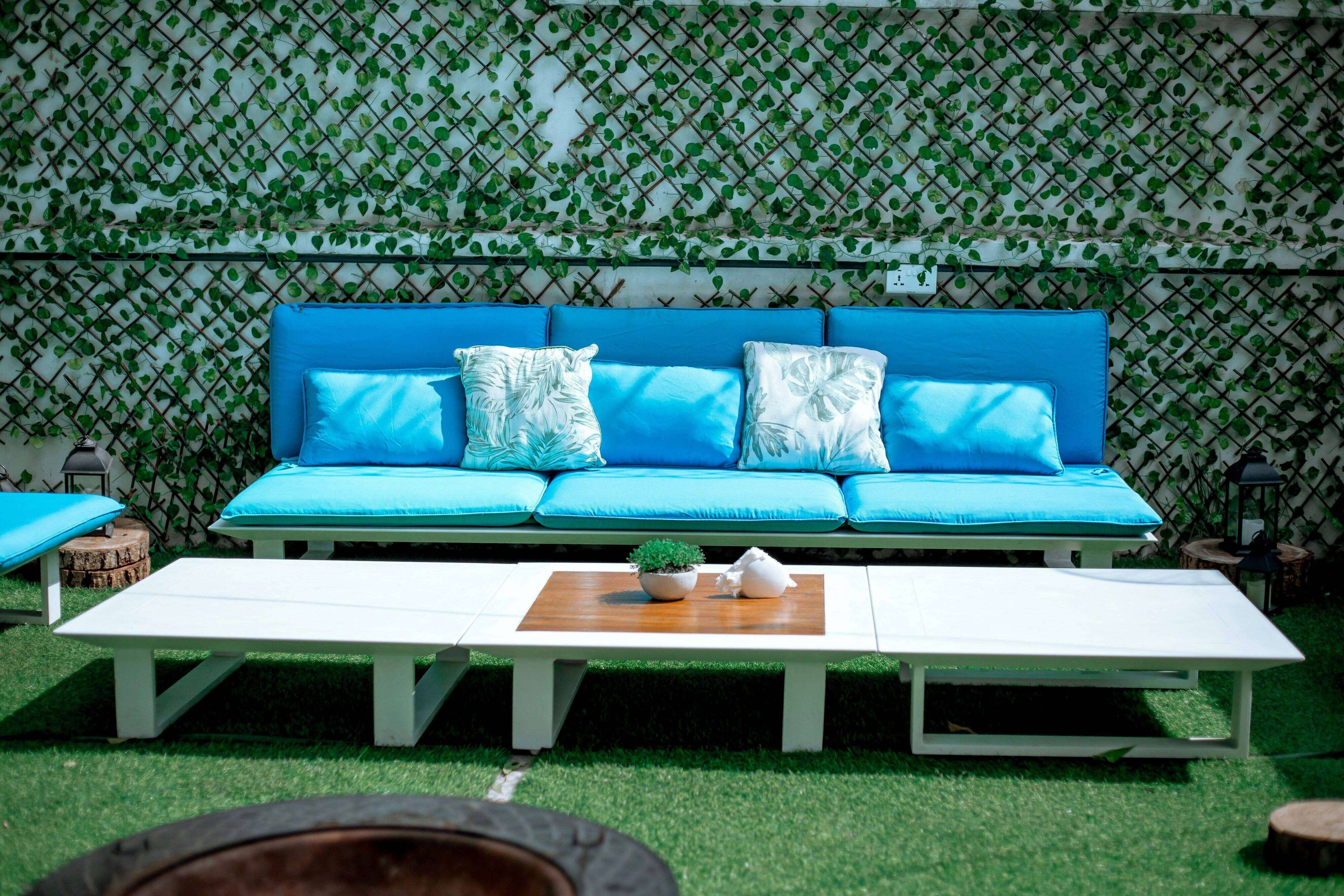} &
        \hspace{\mrg}
        \includegraphics[width=\wid]{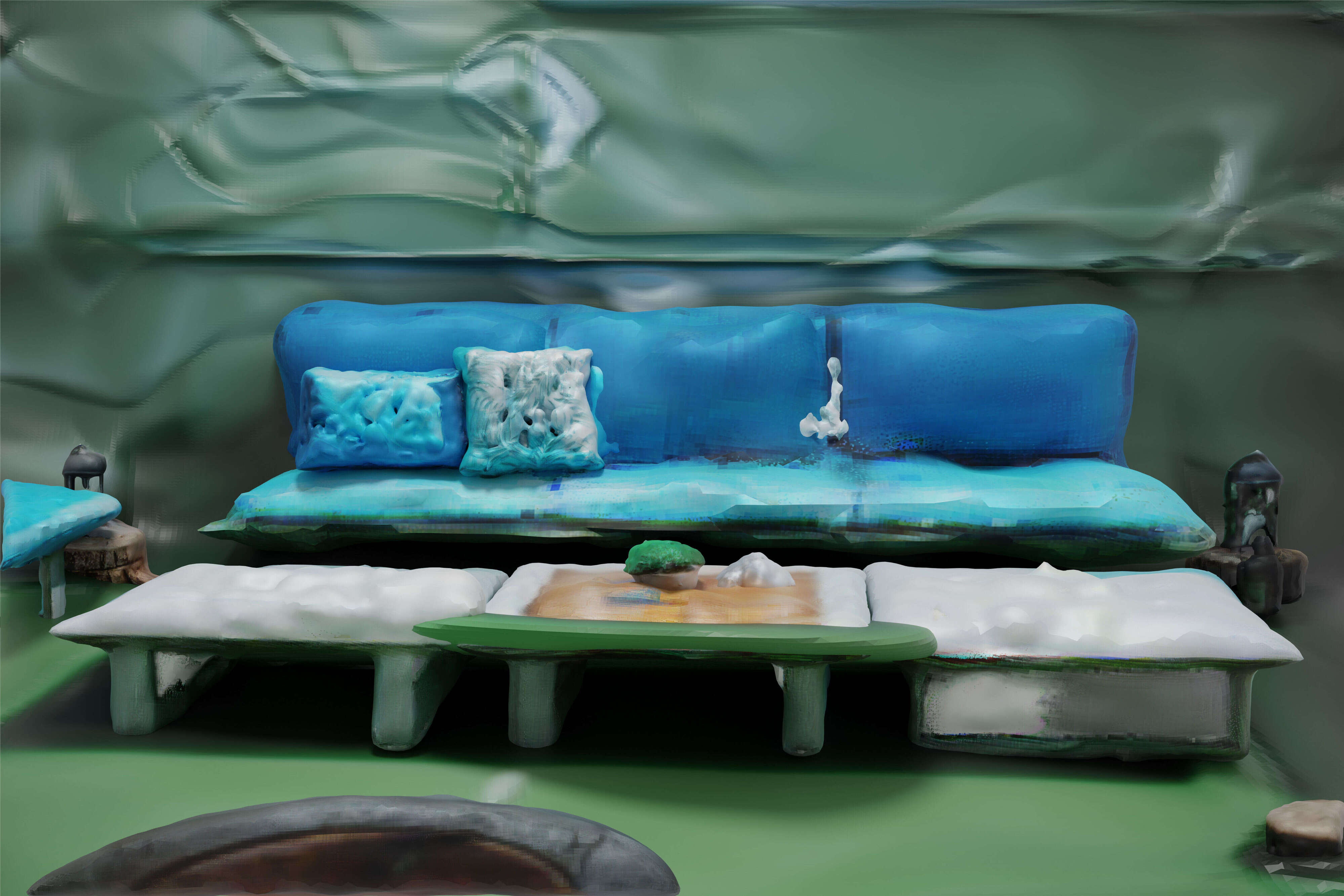} &
        \hspace{\mrg}
        \includegraphics[width=\wid]{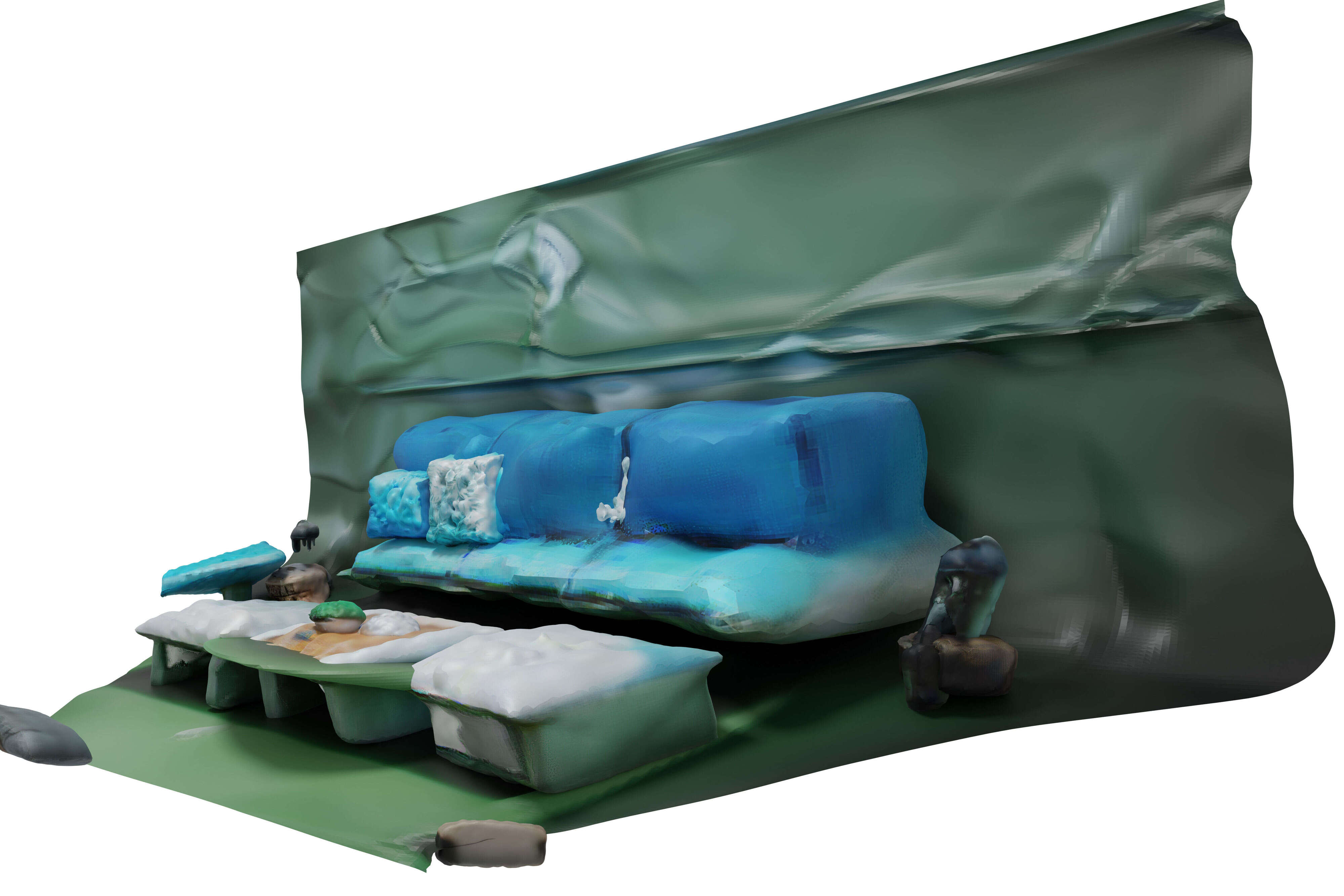} & 
        \hspace{\mrg}
        \includegraphics[width=\wid]{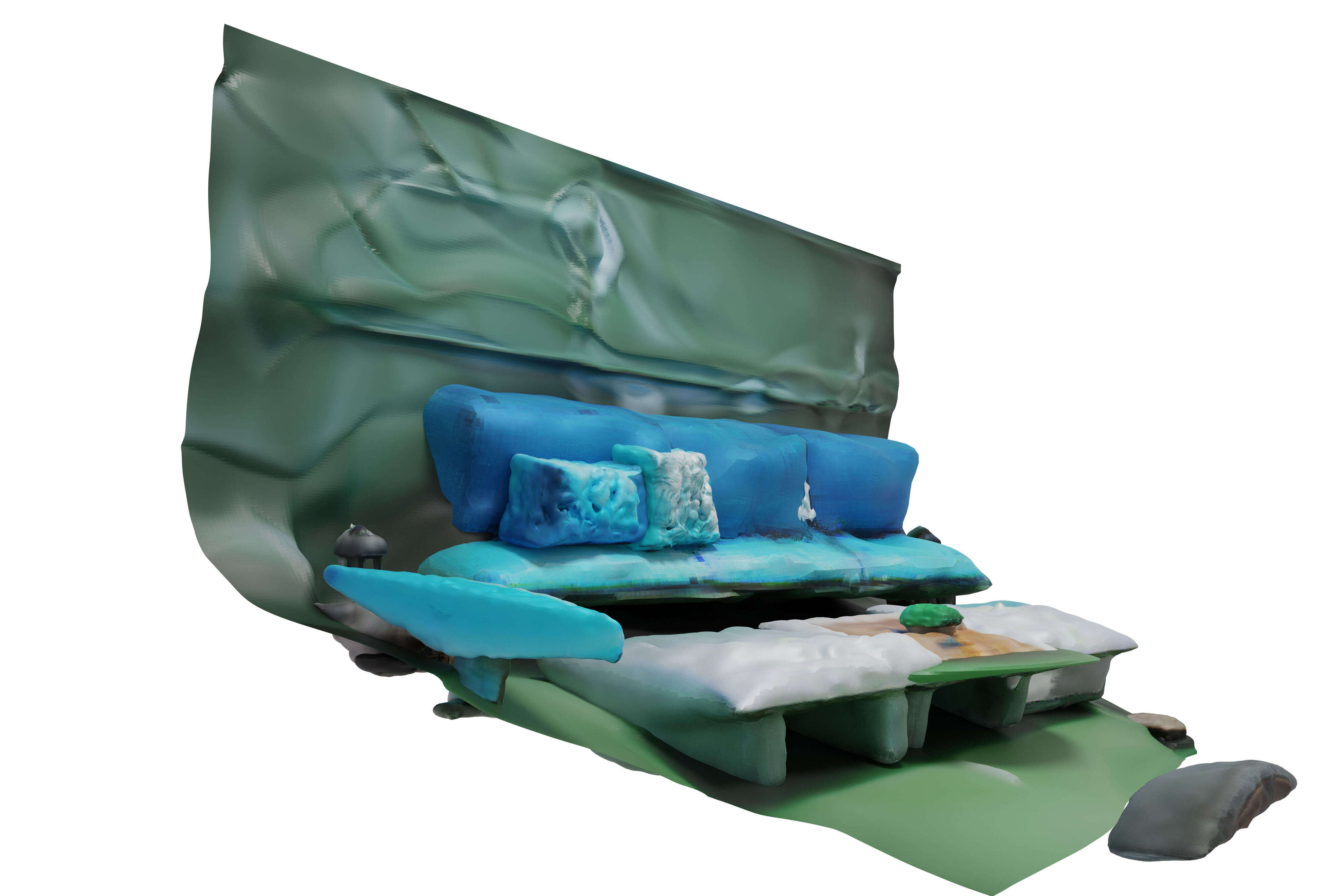} 
        \\ 
        \includegraphics[width=\wid]{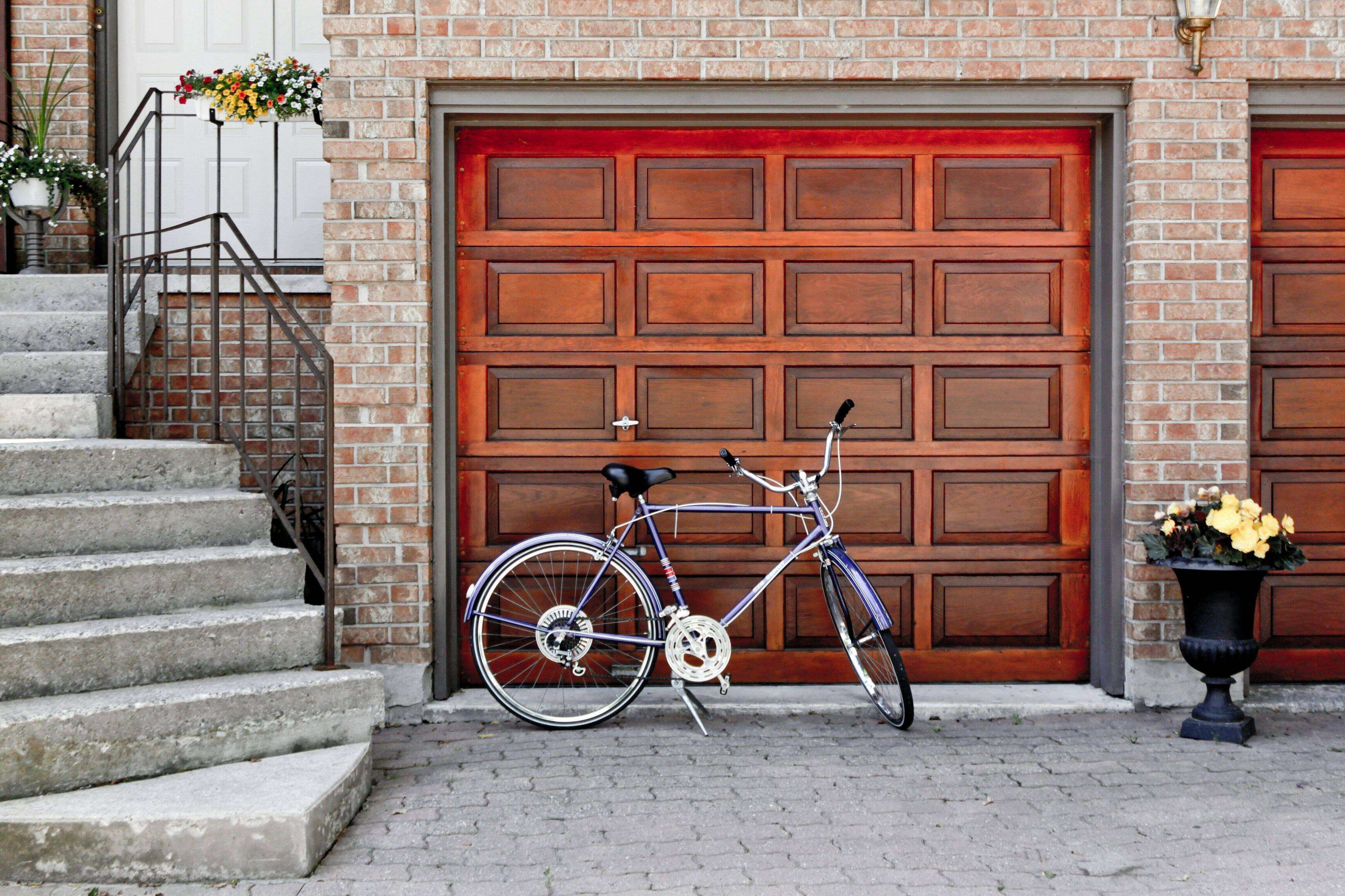} &
        \hspace{\mrg}
        \includegraphics[width=\wid]{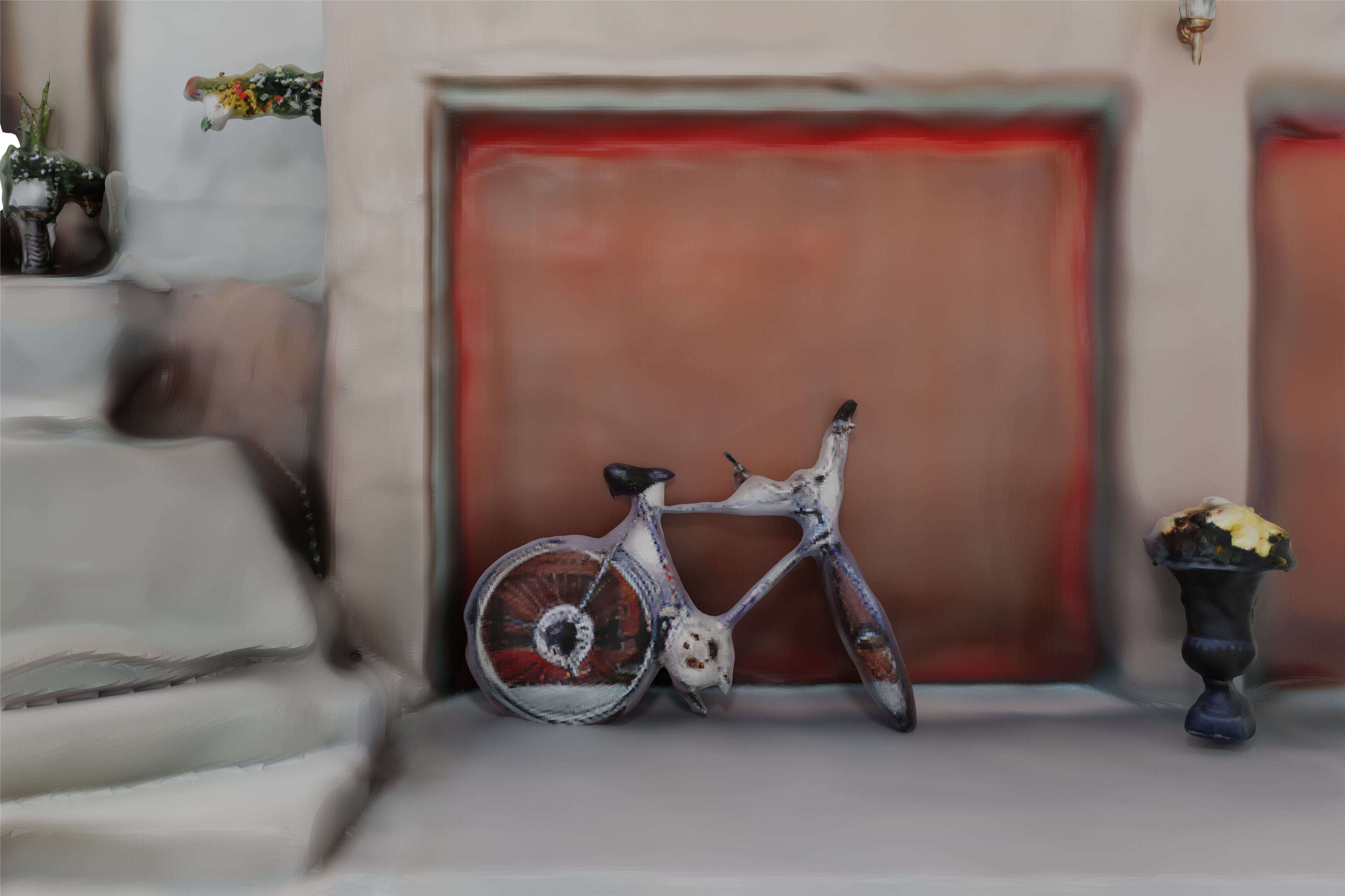} &
        \hspace{\mrg}
        \includegraphics[width=\wid]{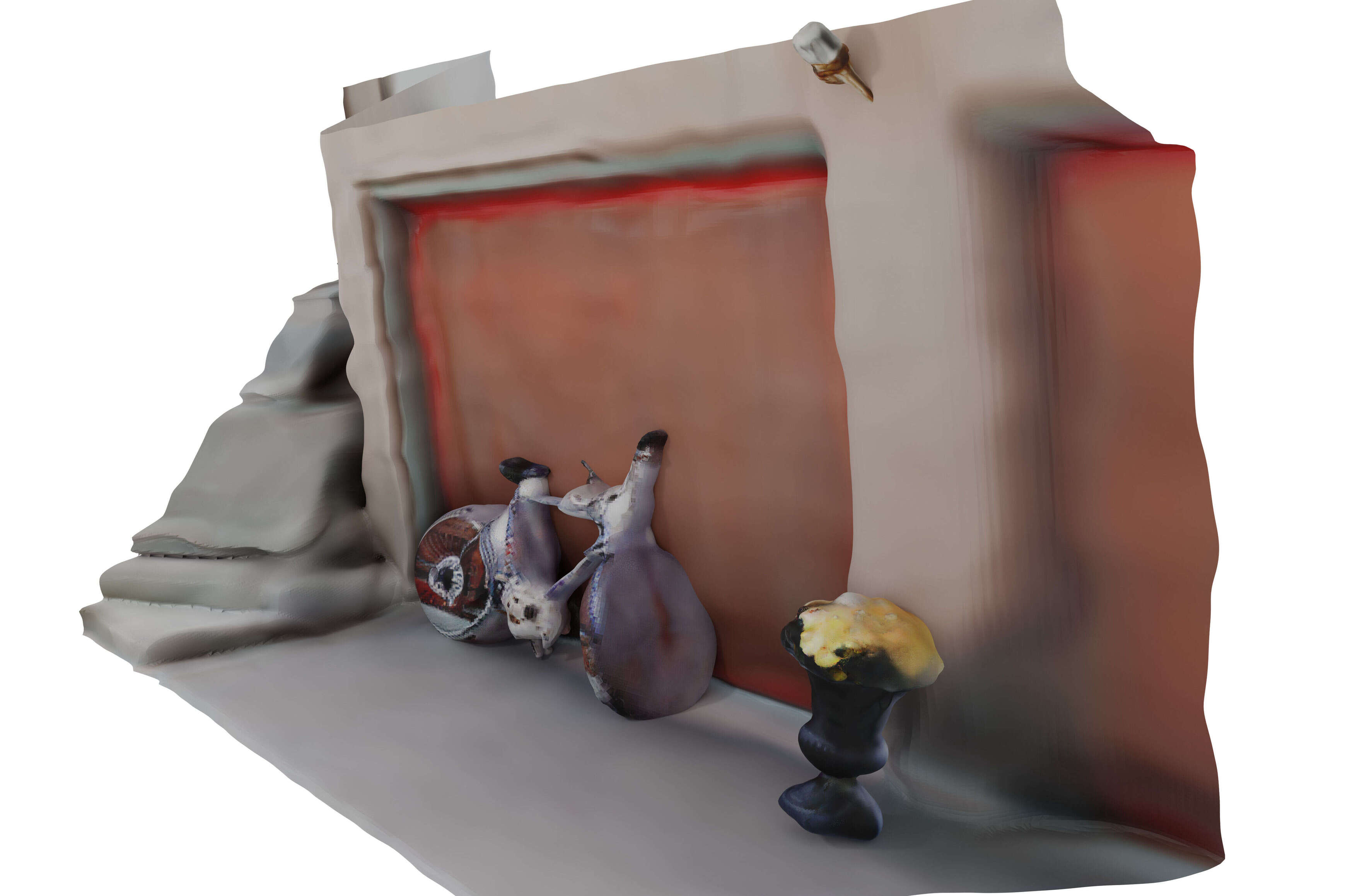} & 
        \hspace{\mrg}
        \includegraphics[width=\wid]{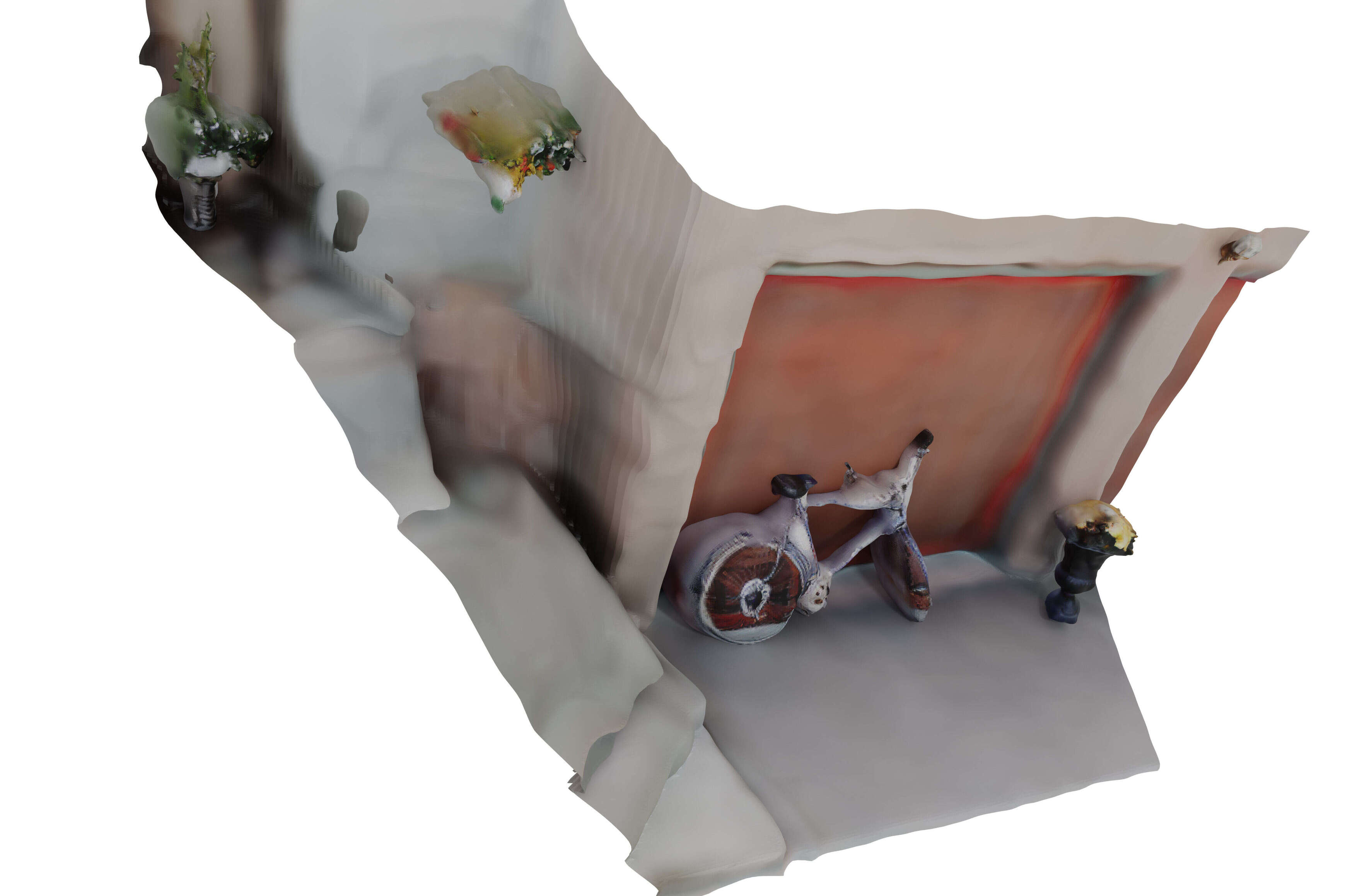} 
        \\ 
        \vspace{\mrgv}
        Input image & Ours & & 
    \end{tabular}
    \caption{Qualitative results of our method on real-world outdoor scene reconstruction.
    } 
    \label{fig:out_qual}
    \vspace{-0.3cm}
\end{figure*}

Our method can reconstruct well a wide range of scenes, from tabletop setups to large rooms with many objects of varying sizes. Though we exemplify the performance of the proposed pipeline on indoor environments, by design the method is also applicable to outdoor scene reconstruction. In Figure \ref{fig:out_qual}, we provide qualitative results for such scenes. In these experiments, we use Marigold~\cite{ke2023repurposing} as a depth estimator and compute the scale and shift based on Depth Anything~\cite{depthanything} prediction of the model fine-tuned on KITTI~\cite{geiger2013vision} (suited for outdoor scenes). 

We choose to focus more on indoor scene reconstruction because most outdoor scenes are predominantly composed of entities categorized as stuff, which we represent simply with a mesh approximating the estimated depth map. We believe a dedicated approach for outdoor background reconstruction would be better suited for this type of scenes. Nonetheless, outdoor scenes with several objects are reconstructed reasonably well by our method, as can be observed in the qualitative results. 

\section{Limitations}
\label{sup:limit}

We describe in this section some of the limitations our method inherits from the integrated modules.

\begin{figure*}
    \centering    
    \setlength{\wid}{0.30\textwidth}
    \setlength{\mrg}{-0.0cm}
    \setlength{\mrgv}{0cm}
    \begin{tabular}{ccc}
        \includegraphics[height=0.85\wid]{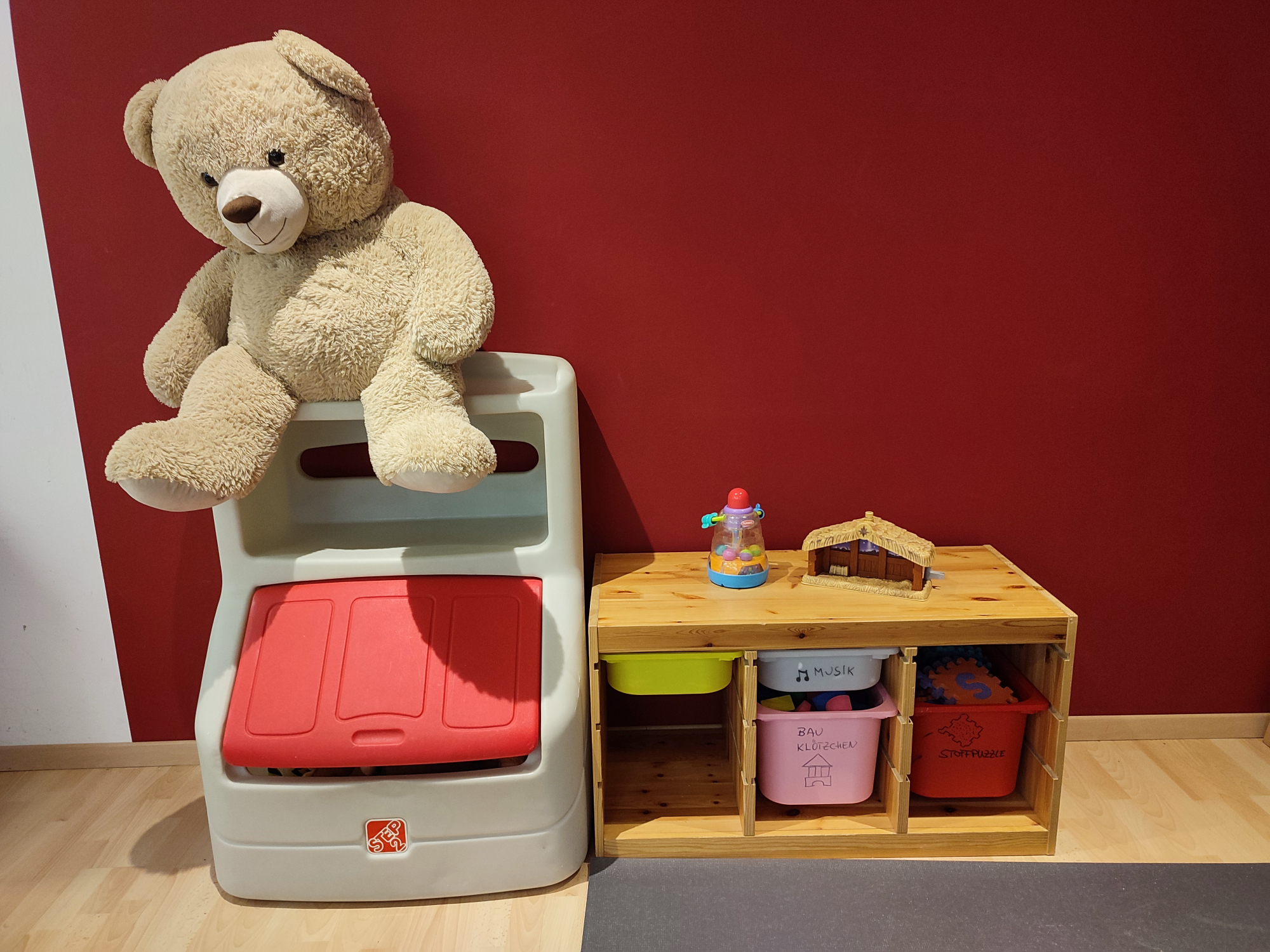} & 
        \includegraphics[height=\wid]{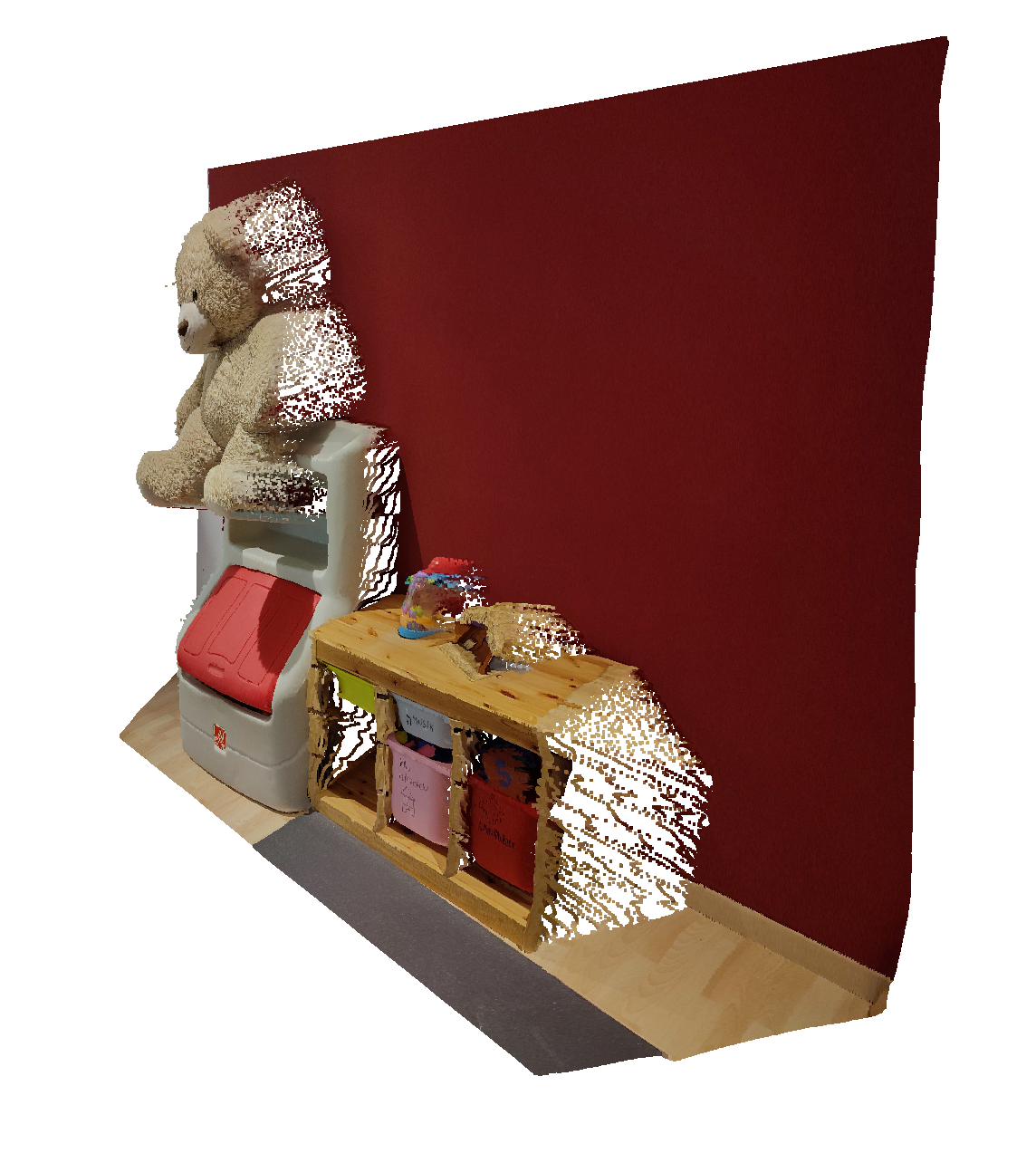} &
        \includegraphics[height=\wid]{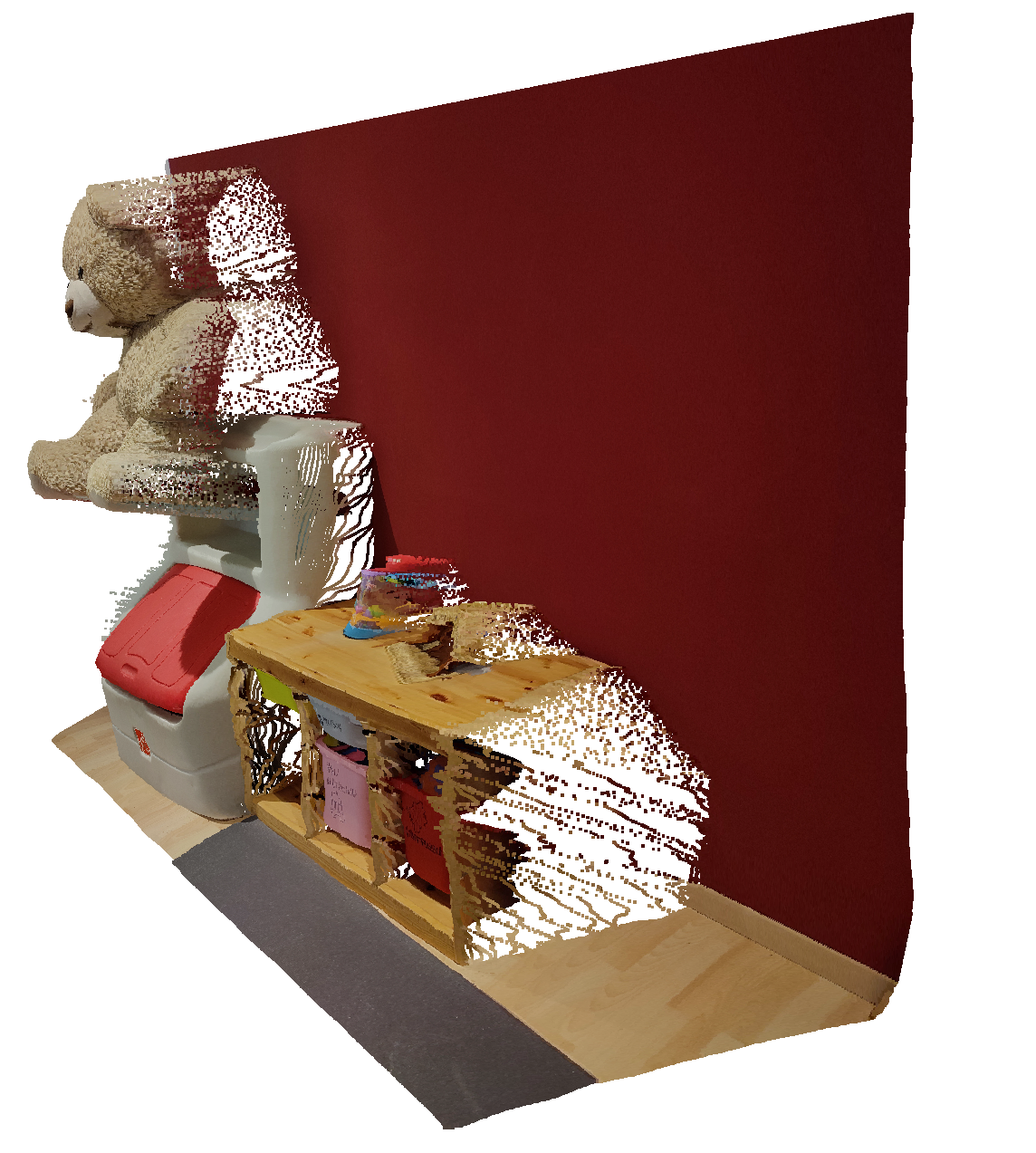}
        \\
       Input image & Predicted FOV &  Adjusted FOV
    \end{tabular}
    \caption{Comparison of unprojected depth under two camera settings. The field of view prediction of PerspectiveFields~\cite{jin2023perspective} is incorrect for this image, resulting in a flattened scene. The unprojected depth using a camera with a manually adjusted the field of view has more realistic proportions.} 
    \label{fig:fov}
\end{figure*}

As we rely on the unprojected depth for the layout of the scene, when the camera calibration or the scale and offset estimation fail to match the correct units, the reconstructed scene will have an erroneous structure. We provide an example of incorrect camera calibration in Figure \ref{fig:fov}. Furthermore, since entities are reconstructed independently, without relative constraints between the objects, the method does not address potential physical impossibilities such as intersecting objects.

\begin{figure*}
    \centering
    \begin{subfigure}{\textwidth}
        \centering
        \includegraphics[width=\textwidth]{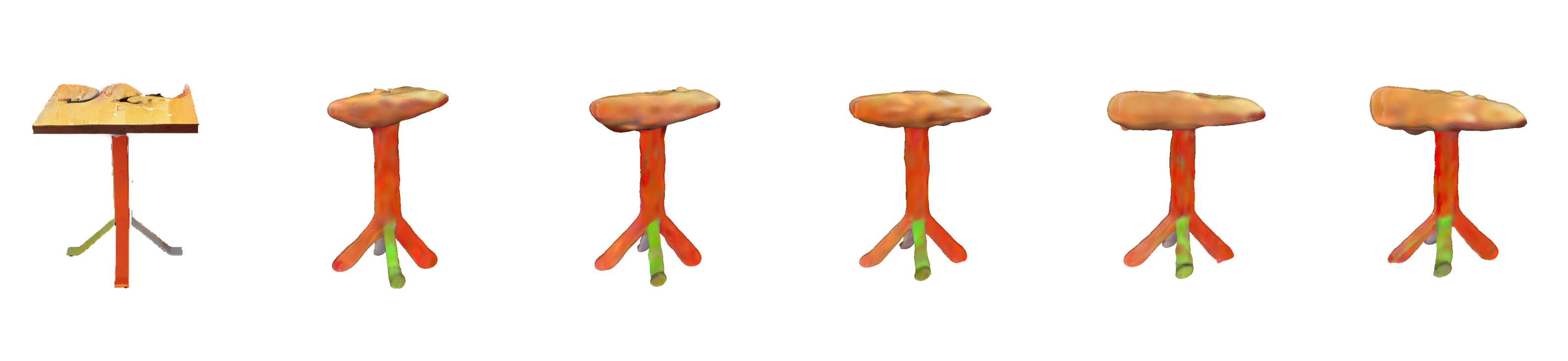}
    \end{subfigure}

    \vspace{-0.2cm}
    \begin{tabularx}{\textwidth}{XXXXXX}
        \centering Input view & \centering \textcolor{purple}{$-35 ^{\circ}$} & \centering $-30 ^{\circ}$ & \centering \textcolor{teal}{$-25 ^{\circ}$} & \centering $-20 ^{\circ}$ & \centering $-15 ^{\circ}$ 
    \end{tabularx}
    \caption{DreamGaussian~\cite{tang2023dreamgaussian} object reconstructions of the input view under different camera elevation angles. We illustrate the 3D reconstructions using a camera view from the side to highlight their differences. The elevation angle estimated by the method proposed in \cite{liu2023one} is $-35 ^{\circ}$. However, the best reconstruction is achieved using an elevation angle of around $-25 ^{\circ}$.}
    \vspace{-1.8em}
    \label{fig:elevation}
\end{figure*}

In our experiments, we notice that DreamGaussian sometimes overestimates the size of the object along the $z$-dimension, when it cannot be observed in the input view. 
Such cases are prominent in Figure~\ref{fig:hope_qual}, \eg, orange juice box in the last example. 
Additionally, DreamGaussian requires the estimation of the perceived camera elevation of the input crop with respect to the normalized object pose. 
Since this cannot be robustly inferred, it significantly affects the reconstruction performance of the model, as can be analysed in Figure~\ref{fig:elevation}. 

Though our approach for modeling the background is superior to previous works, using a small MLP to model the `stuff' of the scene (entities that are not objects) may sacrifice details in the texture. Also, the implicit interpolation for the occluded areas is sometimes insufficient to cover large missing regions, resulting in holes in the background surface. 

As the proposed method follows a modular approach, we can improve on this limitations by upgrading the particular modules. For example, conditioning the depth estimation on the inferred camera calibration can help disambiguate depth scale \cite{saxena2023zero, hu2024metric3dv2}. Also, the 3D object reconstruction module can benefit from having the estimated instance depth as an additional input \cite{huang2023zeroshape, Wu_2023_MCC}. Lastly, relying more on the overall scene context during the amodal completion stage can improve the recovery of missing object parts due to occlusion, as it has been recently shown in \cite{ozguroglu2024pix2gestalt, xu2023amodal}.

\end{document}